\pdfoutput=1
\documentclass[11pt]{article}

\usepackage[]{acl}

\usepackage{times}
\usepackage{latexsym}

\usepackage[T1]{fontenc}
\usepackage[utf8]{inputenc}

\usepackage{microtype}

\usepackage{inconsolata}

\usepackage{hyperref}
\usepackage{url}
\usepackage{booktabs}       % professional-quality tables
\usepackage{amsfonts}       % blackboard math symbols
\usepackage{nicefrac}       % compact symbols for 1/2, etc.
\usepackage{microtype}      % microtypography
\usepackage{xcolor}         % colors
\usepackage{subfig}
\usepackage{amsmath}
\usepackage{multirow}
\usepackage{multicol}
\usepackage{wrapfig}
\usepackage[ruled,linesnumbered,boxed]{algorithm2e}
\usepackage{graphicx}
\usepackage{lipsum}
\usepackage{adjustbox}
\usepackage{subfig,graphicx}
\usepackage{tabularx}
\usepackage{array}
\usepackage{color}
\usepackage{colortbl}
\usepackage{tcolorbox}

\title{Can LLMs Learn New Concepts Incrementally without Forgetting?}

\author{Junhao Zheng, Shengjie Qiu, Qianli Ma* \\
  School of Computer Science and Engineering, \\
  South China University of Technology, Guangzhou, China\\
  \texttt{junhaozheng47@outlook.com}, 
  \texttt{shengjieqiu6@gmail.com},
  \texttt{qianlima@scut.edu.cn}\thanks{*Corresponding author}}

\begin{document}
\maketitle
\addtocontents{toc}{\protect\setcounter{tocdepth}{0}}
\begin{abstract}

Large Language Models (LLMs) have achieved remarkable success across various tasks, yet their ability to learn incrementally without forgetting remains underexplored. Incremental learning (IL) is crucial as it enables models to acquire new knowledge while retaining previously learned information, akin to human learning. Existing benchmarks for IL are insufficient due to data leakage issues and the overqualification of LLMs. To address these challenges, we introduce Concept-1K, a novel dataset comprising 1,023 recently emerged concepts across diverse domains.
The concepts in Concept-1K are discrete, interpretable units of knowledge that allow for fine-grained analysis of learning and forgetting processes. Using Concept-1K as a testbed, we aim to answer the question: ``Can LLMs learn new concepts incrementally without forgetting like humans?'' Our investigation reveals that LLMs still suffer from catastrophic forgetting and that LoRA, despite fine-tuning fewer parameters, may lead to more forgetting on training data. Additionally, we explore the roles of in-context learning, model scale, buffer size, and pretraining in IL performance. These findings highlight the strengths and limitations of LLMs in IL scenarios and provide a robust benchmark for future research.
% The data, code and scripts are in the supplementary material and will be publicly available \footnote{Anonymous URL: \href{https://anonymous.4open.science/r/code-for-concept-1k-B366}{https://anonymous.4open.science/r/code-for-concept-1k-B366}}. 
The data, code and scripts are publicly available \footnote{https://github.com/zzz47zzz/codebase-for-incremental-learning-with-llm}.
\end{abstract}
\section{Introduction}
\label{sec:introduction}

Large Language Models (LLMs) have recently achieved remarkable success, exhibiting human-level performance on various professional and academic benchmarks \cite{OpenAI2023GPT4TR}. Numerous studies have investigated various abilities of LLMs, such as reasoning \cite{wei2022chain}, programming \cite{chen2021evaluating}, and planning \cite{yao2022react}. However, a crucial human ability, incremental learning (IL) (also known as continual learning), remains less explored in LLMs.

Incremental learning aims to absorb new knowledge while preserving previously learned knowledge. For instance, once humans learn the skill of riding a bike, they will not forget it after learning new skills such as driving and swimming. Naturally, one might wonder, ``Since LLMs are so powerful, do they still suffer from forgetting when learning incrementally?''

To answer this question, we first need to find a proper benchmark for evaluating the IL ability of LLMs. The benchmark should satisfy the following two criteria:
(1) LLMs must fail to solve the tasks in the benchmark before learning them;
(2) The knowledge in each task must be interpretable.
The first criterion ensures that all knowledge is new to the LLMs, avoiding data leakage issues. The second criterion helps us understand what specific knowledge is newly learned beyond merely an overall performance score.

\begin{figure}[!t]
    \centering
    \includegraphics[width=0.99\linewidth]{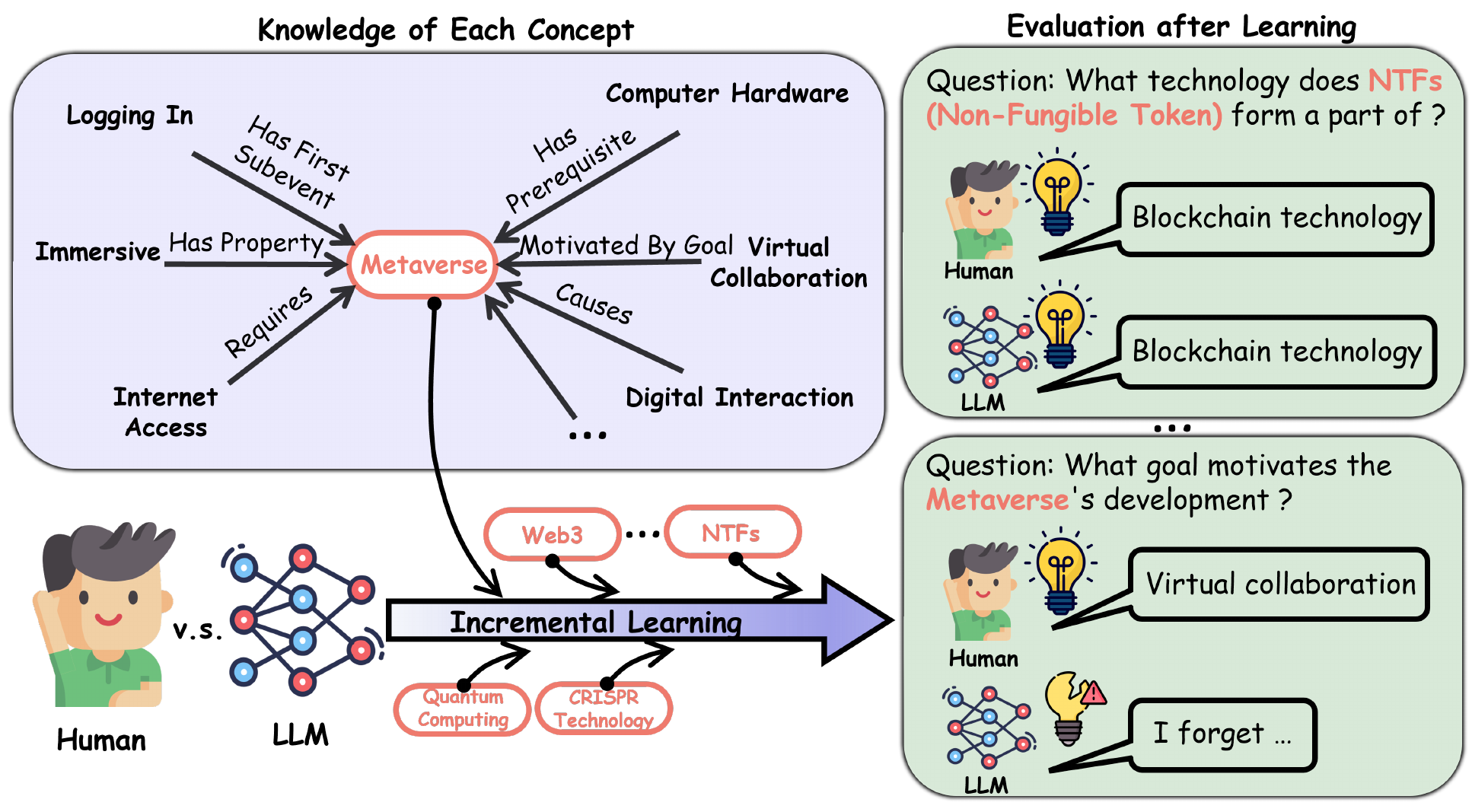}
    \caption{The illustration of the proposed Concept-1K. LLMs suffer from catastrophic forgetting when learning new concepts while humans do not.}
    \label{fig:illustraion_iil}
\end{figure}

\begin{table}[!t]
  \centering
  \caption{The data leakage issue in popular datasets for IL. The linear probing performance \cite{zheng2023learn} on Topic3Datasets, CLINC150, FewRel, OntoNotes5, and I2B2 before IL training is reported, as well as the test accuracy of Concept-1K before IL training. ``/'' represents not applicable.}
  \resizebox{\linewidth}{!}{
    \begin{tabular}{lcccccc}
    \toprule
          & Topic3Datasets & CLINC150 & FewRel & OntoNotes5 & I2B2  & Concept-1K \\
    \midrule
    Pythia-410m & 87.45\tiny{±0.36} & 91.05\tiny{±0.65} & 74.16\tiny{±0.11} & /     & /     & 0.68\tiny{±0.17} \\
    bert-base-cased & 88.02\tiny{±0.56} & 80.39\tiny{±0.27} & 52.18\tiny{±0.05} & 52.93\tiny{±0.48} & 58.29\tiny{±0.73} & / \\
    \bottomrule
    \end{tabular}%
    }
  \label{tab:data_leakage_classification}%
\end{table}%

However, none of the existing benchmarks satisfy these two criteria simultaneously. Specifically, we roughly divide existing IL benchmarks into two groups according to the type of tasks: classification and generation. \emph{Classification benchmarks} are widely used in IL studies from the pre-LLM era, including text classification \cite{zhang2015character}, named entity recognition \cite{ding-etal-2021-nerd}, and relation extraction \cite{han-etal-2018-fewrel}. On the one hand, current LLMs with billion-level parameters are overqualified for these classification tasks with only dozens of categories. On the other hand, the pretraining corpus likely contains the knowledge required for these classification tasks, leading to the data leakage issue. As shown empirically by \citet{zheng2023learn}, sequentially training frozen LLMs with expanding classifiers yields comparable or even superior performance to state-of-the-art (SOTA) IL methods. 
\emph{Generation benchmarks} \cite{zhang2023citb,wang-etal-2022-super} include various tasks such as question generation, style transfer, and wrong candidate generation. However, the data leakage issue remains. In the experiments of \citet{zhang2023citb}, training T5 \cite{raffel2020exploring} on 19 various tasks jointly achieves 42.1\% average performance (i.e., upper bound performance), while sequential finetuning achieves 35.7\% (i.e., lower bound performance). Further discussion on data leakage issues is provided in Appendix \ref{sec:appendix_data_leakage}.

To address these challenges, we construct a dataset called Concept-1K, which satisfies the two criteria for investigating the IL ability of open-sourced LLMs such as LLaMa \cite{touvron2023llama}. Specifically, Concept-1K minimizes the data leakage issue by selecting recently emerged concepts such as ``Metaverse'' and ``Quantum Computing'' from various vertical domains that require domain-specific knowledge to answer. The comparison between the popular datasets and Concept-1K is summarized in Table \ref{tab:data_leakage_classification}. Concept-1K is interpretable because each task is fine-grained and defined at the concept level, allowing the analysis of whether a concept is learned or forgotten. Additionally, Concept-1K contains 1,023 concepts, supporting an order of magnitude larger incremental learning steps than existing benchmarks, which can push LLMs' IL ability \emph{to their limits}.

Using the constructed Concept-1K as a testbed, we aim to answer the question: ``Can LLMs learn new concepts incrementally without forgetting, like humans?'' The choice of ``concept'' as the fundamental unit in Concept-1K is deliberate. As shown in Figure \ref{fig:illustraion_iil}, concepts are discrete, interpretable units of knowledge that allow for fine-grained analysis of learning and forgetting processes. By focusing on concepts, we can precisely identify what knowledge is acquired, retained, or forgotten, providing clearer insights into the incremental learning abilities of LLMs. Our investigation also delves into how in-context learning, parameter-efficient methods like LoRA \cite{hu2021lora}, and factors such as model scale, buffer size, and pretraining influence IL performance.

Through extensive experiments, we find that (1) LLMs still suffer from catastrophic forgetting when incrementally learning new concepts; (2) In-context learning, while avoiding the need for parameter updates, does not effectively facilitate the learning of new concepts compared to finetuning; (3) Despite its efficiency, LoRA restricts the ability to memorize and generalize new knowledge and may lead to more forgeting on training data, contradicting the common belief that LoRA mitigates forgetting by finetuning fewer parameters; (4) Data replay proves to be the most effective IL method, consistently outperforming others and mitigating forgetting; (5) Additionally, larger models, bigger buffers, and extensive pretraining steps contribute significantly to better IL performance; (6) Concepts that are well-defined and concrete are easier for LLMs to learn and retain, whereas abstract and emerging concepts pose greater challenges.

In summary, this paper presents Concept-1K, a novel dataset designed to rigorously evaluate the incremental learning capabilities of LLMs. Our findings provide valuable insights into the strengths and limitations of current LLMs in IL scenarios and offer a robust benchmark for future research in this area.

\section{Concept-1K}

\begin{table*}[!t]
  \centering
  \caption{Examples of Concept-1K. Each triplet corresponds to a training instance and a test instance. More examples are provided in Table \ref{tab:examples_concept_1k_appendix}.}
  \resizebox{0.97\linewidth}{!}{
    \begin{tabular}{cllll}
    \toprule
    \multicolumn{1}{l}{\textbf{Domain}} & \textbf{Concept} & \textbf{Triplet} & \textbf{Training and Test Input} & \textbf{Target Output} \\
    \midrule
    \multirow{20}[20]{*}{\textbf{Environment}} & \multirow{10}[10]{*}{\textbf{Groundwater Recharge}} & \multicolumn{1}{l}{\multirow{2}[2]{*}{\textbf{(Groundwater Recharge, IsA, HydrologicalProcess)}}} & What is Groundwater Recharge classified as? & \multirow{2}[2]{*}{hydrological process} \\
          &       &       & What kind of process is Groundwater Recharge? &  \\
\cmidrule{3-5}          &       & \multirow{2}[2]{*}{\textbf{(Groundwater Recharge, UsedFor, AquiferSustainability)}} & What is Groundwater Recharge used for? & \multirow{2}[2]{*}{aquifer sustainability} \\
          &       &       & What purpose does Groundwater Recharge serve in relation to aquifers? &  \\
\cmidrule{3-5}          &       & \multirow{2}[2]{*}{\textbf{(Groundwater Recharge, Requires, PermeableSurfaces)}} & What does Groundwater Recharge require? & \multirow{2}[2]{*}{permeable surfaces} \\
          &       &       & What are essential for the process of Groundwater Recharge? &  \\
\cmidrule{3-5}          &       & \multirow{2}[2]{*}{\textbf{(Groundwater Recharge, ResultsIn, WaterTableRise)}} & What is a result of Groundwater Recharge? & \multirow{2}[2]{*}{water table rise} \\
          &       &       & What does Groundwater Recharge lead to regarding water tables? &  \\
\cmidrule{3-5}          &       & \multirow{2}[2]{*}{\textbf{(Groundwater Recharge, MotivatedByGoal, DroughtMitigation)}} & What goal motivates Groundwater Recharge? & \multirow{2}[2]{*}{drought mitigation} \\
          &       &       & Why is Groundwater Recharge important? &  \\
\cmidrule{2-5}          & \multirow{10}[10]{*}{\textbf{Sea Level Rise}} & \multicolumn{1}{l}{\multirow{2}[2]{*}{\textbf{(Sea Level Rise, CausedBy, GlobalWarming)}}} & What causes Sea Level Rise? & \multirow{2}[2]{*}{global warming} \\
          &       &       & What is the primary factor leading to Sea Level Rise? &  \\
\cmidrule{3-5}          &       & \multirow{2}[2]{*}{\textbf{(Sea Level Rise, AnalyzedBy, Climatologists)}} & Who analyzes Sea Level Rise? & \multirow{2}[2]{*}{climatologists} \\
          &       &       & What group of professionals study Sea Level Rise? &  \\
\cmidrule{3-5}          &       & \multirow{2}[2]{*}{\textbf{(Sea Level Rise, ResultsIn, HabitatLoss)}} & What does Sea Level Rise result in? & \multirow{2}[2]{*}{habitat loss} \\
          &       &       & What is a significant impact of Sea Level Rise on natural habitats? &  \\
\cmidrule{3-5}          &       & \multirow{2}[2]{*}{\textbf{(Sea Level Rise, MeasuredBy, TideGauges)}} & How is Sea Level Rise measured? & \multirow{2}[2]{*}{tide gauges} \\
          &       &       & What instrument is used to measure Sea Level Rise? &  \\
\cmidrule{3-5}          &       & \multirow{2}[2]{*}{\textbf{(Sea Level Rise, AddressedBy, EmissionReductions)}} & How is Sea Level Rise addressed? & \multirow{2}[2]{*}{emission reductions} \\
          &       &       & What strategy addresses Sea Level Rise? &  \\
    \bottomrule
    \end{tabular}%
    }
  \label{tab:examples_concept_1k_main}%
\end{table*}

\subsection{Problem Formulation}
We consider an incremental scenario where LLMs explicitly learn the knowledge of each concept. Specifically, we aim to train a model $f_\theta:\mathbf{x} \rightarrow \mathbf{y}$ from a sequence of concepts $\mathcal{C}=\{\mathcal{C}_1, \mathcal{C}_2, \cdots, \mathcal{C}_n, \cdots, \mathcal{C}_N\}$, where $N$ is the number of concepts, and both the input $\mathbf{x}$ and output $\mathbf{y}$ are natural language. The $n$-th concept $\mathcal{C}_n$ contains $M_n$ training-test pairs $\mathcal{D}^{(n)}=\{\mathbf{x}_i^{(n),train}, \mathbf{x}_i^{(n),test}, \mathbf{y}_i^{(n)}\}_{i=1}^{M_n}$, where $\mathbf{x}_i^{(n),train}$ and $\mathbf{x}_i^{(n),test}$ are the training and test inputs, and $\mathbf{y}_i^{(n)}$ is the target output. Each training-test pair corresponds to the same knowledge point about the concept $\mathcal{C}_n$. 

For instance, in Table \ref{tab:examples_concept_1k_main}, the target output for both questions, ``What is Groundwater Recharge classified as?'' and ``What kind of process is Groundwater Recharge?'' is ``hydrological process''. We expect LLMs to learn the knowledge point ``Groundwater Recharge, IsA, HydrologicalProcess'' from the training sample and generalize it to answer the rephrased test question correctly. For practical training and evaluation, we evenly divide $N$ concepts into $T$ ($T \leq N$) tasks. The model is evaluated after learning the concepts in each task.

\subsection{Evaluation Metric}
We adopt four evaluation metrics for Concept-1K: Memorization Accuracy (MA), Memorization Forgetting rate (MF), Generalization Accuracy (GA), and Generalization Forgetting rate (GF). Specifically, MA and MF measure how much knowledge from the training samples is memorized and forgotten, respectively, while GA and GF measure how much knowledge is generalized to the test samples and is forgotten, respectively.

Memorization accuracy is defined as:
\begin{equation}
    MA = \frac{1}{T}\sum_{t=1}^{T}\mathcal{A}_t,\quad
    \mathcal{A}_t = \frac{1}{t}\sum_{i=1}^{t}a_{t,i},
\end{equation}
where $T$ is the number of tasks. $\mathcal{A}_t$ represents the average accuracy on the training instances from all learned concepts. $a_{t,i}$ represents the accuracy evaluated on the $i$-th task after training the model incrementally from concepts belonging to task $1, \cdots, t$. The accuracy is calculated as the exact match between the model output and the target output.

Memorization forgetting is computed as the average accuracy on all training instances of all learned concepts:
\begin{equation}
    MF = \frac{1}{T-1}\sum_{i=1}^{T-1}\left[ \max_{j<T}(\{a_{j,i}\}_j) - a_{T,i}\right],
\end{equation}
where $\max_{j<T}(\{a_{j,i}\}_j)$ represents the highest accuracy of task $i$ since it has been learned, and $a_{T,i}$ represents the accuracy of task $i$ at step $T$. $\left[\max_{j<T}(\{a_{j,i}\}_j) - a_{T,i}\right]$ computes the decrease in the accuracy of task $i$ when learning the $T$-th task.

Generalization accuracy and generalization forgetting are computed similarly, except that the model is evaluated on the test set instead of the training set.

\subsection{Dataset Construction}
To avoid data leakage, we collect novel concepts from six domains: economy, culture, science and technology, environment, education, and health and medical. Introductions to the concepts in each domain are provided in Appendix \ref{sec:appendix_domain_introduction}. Initially, we generate 600 concepts for each domain using GPT-4. We then manually filter out outdated, vague, or imaginary concepts and select the latest, most specific, and most informative concepts, resulting in a total of 1,023 concepts. We follow three criteria in this process: \emph{Length criterion}, \emph{Timeliness criterion}, and \emph{Trend criterion}. Detailed description is provided in Appendix \ref{sec:appendix_concept_selection_criterion}.

The concept list is provided in Table \ref{tab:concept_list_1}. Next, we use triplets to represent ``knowledge'' and prompt GPT-4 to construct 20 triplets for each concept with the relations in ConceptNet \cite{speer2017conceptnet}. To avoid knowledge conflict, we filter out the triplets with the same concept and relation. Additionally, we filter out triplets with relations such as ``RelatedTo'' and ``HasContext'' to ensure specificity. Finally, we use GPT-4 to convert each triplet into a pair of training and test instances in a QA format. Examples are provided in Table \ref{tab:examples_concept_1k_main} and \ref{tab:examples_concept_1k_appendix}, a word cloud diagram in Figure \ref{fig:word_cloud_diagram}, and statistics of Concept-1K in Table \ref{tab:comparison_num_class} and Figure \ref{fig:histogram_relation}.

\begin{table}[!t]
  \centering
  \caption{Comparison between Concept-1K and widely-used datasets for incremental learning with LLMs. Concept-1K supports \textbf{an order of magnitude larger} incremental learning steps than existing ones.}
  \resizebox{0.95\linewidth}{!}{
    \begin{tabular}{lcc}
    \toprule
    \textbf{Dataset} & \textbf{\# Classes / Concepts} & \textbf{Task Type} \\
    \midrule
    AGNews \cite{zhang2015character} & 4     & \multirow{3}[2]{*}{Topic Classification} \\
    DBPedia \cite{zhang2015character} & 14    &  \\
    YaHoo \cite{zhang2015character} & 10    &  \\
    \midrule
    CLINC150 \cite{larson-etal-2019-evaluation} & 150   & \multirow{2}[2]{*}{Intent Classification} \\
    Banking77 \cite{casanueva-etal-2020-efficient} & 77    &  \\
    \midrule
    FewRel \cite{han-etal-2018-fewrel} & 80    & \multirow{2}[2]{*}{Relation Extraction} \\
    TACRED \cite{zhang-etal-2017-position} & 40    &  \\
    \midrule
    Few-NERD \cite{ding-etal-2021-nerd} & 66    & \multirow{3}[2]{*}{Named Entity Recognition} \\
    Ontonotes5 \cite{hovy-etal-2006-ontonotes} & 18    &  \\
    I2B2 \cite{murphy2010serving}  & 16    &  \\
    \midrule
    Concept-1K & 1023  & Question Answering \\
    \bottomrule
    \end{tabular}%
    }
  \label{tab:comparison_num_class}%
\end{table}

\subsection{Comparison with Existing Datasets}

\subsubsection{Concept-1K Minimizes Data Leakage}
Concept-1K is designed to minimize data leakage by focusing on novel concepts that emerged after January 2022. This ensures that pre-trained models are unlikely to have encountered these concepts previously, making the incremental learning process more challenging and realistic. The zero-shot performance of models such as GPT-4, GPT-3.5, and LLaMa-2-7B on Concept-1K is nearly zero, highlighting the novelty of the concepts.

\subsubsection{Concept-1K Defined as Instance-Level Incremental Learning}
Unlike other datasets, which are often designed for task-level incremental learning, Concept-1K is constructed under a new scenario called \emph{Instance-level Incremental Learning} (IIL). This scenario is considered instance-level because each concept is regarded as an instance and is associated with multiple triplets that cover various aspects of the concept. A comparison between IIL and popular IL scenarios is provided in Appendix \ref{sec:appendix_comparison_scenario}.

\subsubsection{Concept-1K Supports More Incremental Tasks}
Compared to existing datasets, Concept-1K supports a significantly larger number of incremental learning steps. As shown in Table \ref{tab:comparison_num_class}, while other datasets typically contain a limited number of classes or concepts (ranging from 4 to 150), Concept-1K includes 1,023 concepts. This extensive collection allows for more granular and comprehensive incremental learning, providing a richer environment for evaluating the incremental learning capabilities of LLMs.

\begin{table}[!t]
  \tiny
  \centering
  \caption{Semantic Diversity in Concept-1K}
   % \resizebox{0.80\linewidth}{!}{
    \begin{tabular}{lcc}
    \toprule
          & Concept Name & Question and Answer \\
    \midrule
    Intra-domain & 0.648 & 0.758 \\
    Inter-domain & 0.607 & 0.704 \\
    \midrule
    All   & 0.613 & 0.713 \\
    \bottomrule
    \end{tabular}%
    % }
  \label{tab:semantic_diversity}%
\end{table}%

Additionally, the concepts, questions, and answers are diverse. We computed the cosine similarity of the average last hidden states of bert-base-uncased \cite{devlin-etal-2019-bert}, as shown in Table \ref{tab:semantic_diversity}. Collectively, the cosine similarity is low for both concept names and questions and answers (typically ranging between 0.5 and 1.0).
\section{Experiments}

We split the 1023 concepts in Concept-1K into 10 tasks for incremental learning. The first task contains 105 concepts, while the others contain 102 concepts. We provide introductions to backbones and implementation details in Appendix \ref{sec:appendix_experimental_settings}.

\begin{figure}[!t]
    \centering
    \subfloat[Memorization Accuracy]{
        \includegraphics[width=0.49\linewidth]{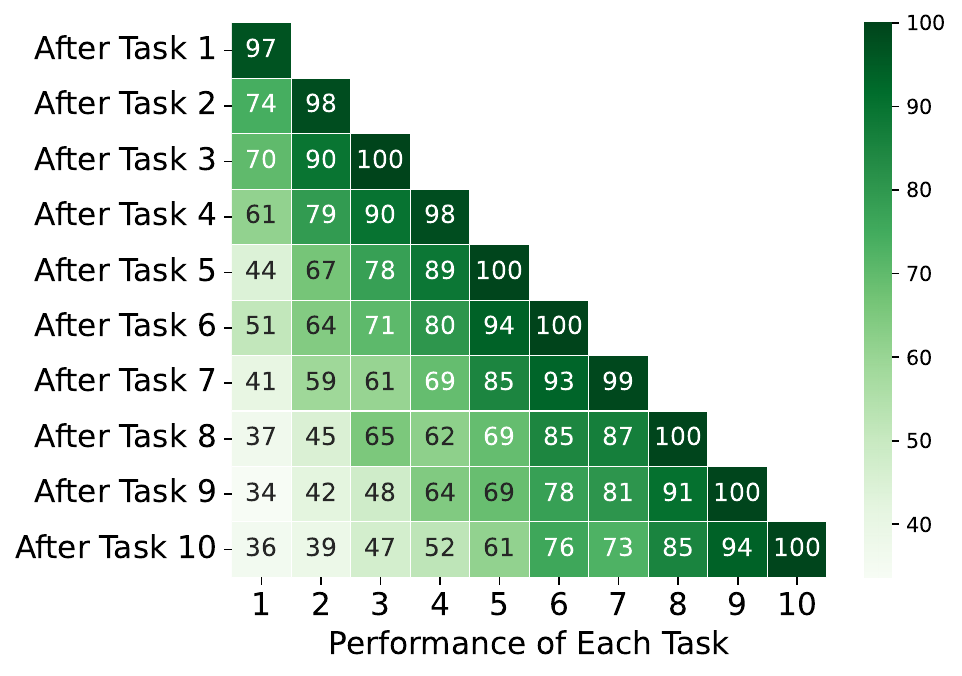}
    }
    \subfloat[Generalization Accuracy]{
        \includegraphics[width=0.49\linewidth]{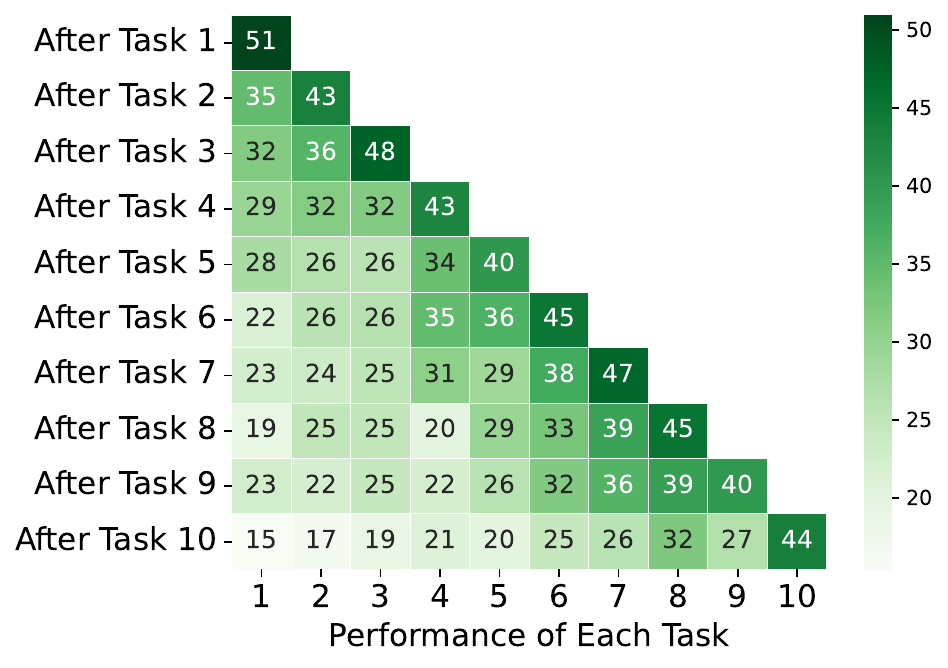}
    }
    \caption{The step-wise performance on Concept-1K. The backbone model is LLaMa-2-7B.}
    \label{fig:step-wise-performance}
\end{figure}

\subsection{RQ1: Can LLMs learn new concepts incrementally without forgetting?}

\begin{tcolorbox}
Main Findings 1: LLMs still suffer from catastrophic forgetting when incrementally learning new concepts.
\end{tcolorbox}

We sequentially fully fine-tuned LLaMa-2-7B on 10 tasks from Concept-1K. Before training, we evaluate the LLM on Concept-1K and find that the accuracy on both the training and test data is nearly zero. This indicates that the LLMs lack the knowledge to answer the questions in Concept-1K, thus avoiding the data leakage issue.

Figure \ref{fig:step-wise-performance} shows a clear tendency for the LLMs to forget old concepts' knowledge when learning new concepts. Specifically, although LLMs achieve 100\% memorization accuracy on each new task, the memorized knowledge is gradually forgotten as more tasks are learned. Similarly, the generalized knowledge also diminishes as new knowledge is acquired. Therefore, despite their power, we conclude that LLMs still suffer from catastrophic forgetting when fully fine-tuning on new data.

\begin{table}[!t]
  \centering
  \caption{The accuracy of in-context learning on the full Concept-1K dataset. ``Rand.'', ``Same Conc.'', and ``Same Know.'' represent that the demonstration samples are selected randomly, or from the instances related to the same concept, or from the instance related to the same knowledge (i.e., the same training-test pair).}
  \resizebox{0.99\linewidth}{!}{
        \begin{tabular}{l|ccc|ccc}
    \toprule
          & \multicolumn{3}{c|}{\textbf{1 Shot}} & \multicolumn{3}{c}{\textbf{5 Shot}} \\
\cmidrule{2-7}          & \textbf{Rand.} & \textbf{Same Conc.} & \textbf{Same Know.} & \textbf{Rand.} & \textbf{Same Conc.} & \textbf{Same Know.} \\
    \midrule
    \textbf{Pythia-70M} & 0.02\tiny{±0.00} & 0.10\tiny{±0.04} & 91.78\tiny{±0.00} & 0.05\tiny{±0.02} & 0.17\tiny{±0.01} & 15.02\tiny{±0.00} \\
    \textbf{Pythia-160M} & 0.10\tiny{±0.02} & 0.15\tiny{±0.02} & 41.36\tiny{±0.00} & 0.37\tiny{±0.02} & 0.55\tiny{±0.15} & 13.49\tiny{±0.00} \\
    \textbf{Pythia-410M} & 0.49\tiny{±0.17} & 0.75\tiny{±0.23} & 40.66\tiny{±0.00} & 2.32\tiny{±0.80} & 2.04\tiny{±0.63} & 18.45\tiny{±0.00} \\
    \textbf{Pythia-1B} & 0.88\tiny{±0.36} & 1.21\tiny{±0.13} & 43.73\tiny{±0.00} & 3.03\tiny{±0.46} & 2.69\tiny{±0.56} & 35.75\tiny{±0.00} \\
    \textbf{Pythia-1.4B} & 1.92\tiny{±0.28} & 2.52\tiny{±0.16} & 54.37\tiny{±0.00} & 4.56\tiny{±0.58} & 3.80\tiny{±0.01} & 45.59\tiny{±0.00} \\
    \textbf{Pythia-2.8B} & 1.81\tiny{±0.45} & 2.50\tiny{±0.27} & 53.56\tiny{±0.00} & 5.69\tiny{±0.35} & 4.07\tiny{±0.93} & 60.20\tiny{±0.00} \\
    \textbf{LLaMa 7B} & 4.32\tiny{±0.11} & 4.93\tiny{±0.72} & 83.57\tiny{±0.00} & 8.79\tiny{±0.16} & 6.24\tiny{±0.23} & 66.25\tiny{±0.00} \\
    \textbf{Vicuna 7B} & 6.67\tiny{±0.74} & 7.00\tiny{±0.88} & 55.67\tiny{±0.00} & 9.63\tiny{±0.78} & 7.75\tiny{±0.15} & 36.04\tiny{±0.00} \\
    \midrule
    \textbf{GPT 3.5} & 6.60  & 8.20  & 51.60  & 8.60  & 13.00  & 74.80  \\
    \textbf{GPT 4} & 10.20  & 10.40  & 76.60  & 7.40  & 21.80  & 86.20  \\
    \bottomrule
    \end{tabular}%
    }
  \label{tab:icl_exp}%
\end{table}%

\subsection{RQ2: Can LLMs learn new concepts through in-context learning instead of finetuning?}
\label{sec:exp_icl}
\begin{tcolorbox}
Main Findings 2: LLMs hardly learn new knowledge through in-context learning compared to finetuning.
\end{tcolorbox}

Given the finding that LLMs tend to forget when learning new concepts, we explore in-context learning as a straightforward method that requires no finetuning and does not cause forgetting. For example, \citet{zheng2023can} show that knowledge can be edited through in-context learning without the need for finetuning. Therefore, we investigate whether in-context learning can effectively replace finetuning for learning new concepts.

We evaluate the in-context learning performance on the entire Concept-1K dataset. Detailed settings and input prompt are provided in Appendix \ref{sec:appendix_experimental_settings}.
Table \ref{tab:icl_exp} shows that GPT-4 achieves 86.20\% under the ``5-shot'' and ``Same Knowledge'' settings, indicating that the training and test instances of Concept-1K share the same knowledge points. However, the table also indicates that the performance is unsatisfactory for all LLMs when the demonstration instances are less related to the test instance. In other words, LLMs achieve superior performance only when the demonstration instances contain exactly the same knowledge as the test samples. Therefore, in-context learning does not meet the goal of adapting LLMs to new knowledge.

Table \ref{tab:icl_exp} also shows that the smallest LLM (Pythia-70M) achieves high accuracy under the ``1-shot'' and ``Same Knowledge'' settings because small LLMs simply copy the output in the demonstration instance as the final output. Under the ``5-shot'' and ``Same Knowledge'' settings, the accuracy of Pythia-70M drops to only 15.02\%.

\begin{table}[!t]
  \centering
  \caption{The performance of full finetuning (FULL) and LoRA on various backbones.}
  \resizebox{0.99\linewidth}{!}{
    \begin{tabular}{l|cc|cc}
    \toprule
          & MA ($\uparrow$) & GA ($\uparrow$) & MF ($\downarrow$) & GF ($\downarrow$) \\
    \midrule
    Pythia-410M (FULL) & 58.28\tiny{±0.64} & 17.68\tiny{±0.31} & 65.19\tiny{±0.31} & 15.39\tiny{±0.16} \\
    Pythia-2.8B (FULL) & 51.91\tiny{±0.51} & 23.18\tiny{±0.14} & 42.65\tiny{±0.64} & 20.59\tiny{±0.20} \\
    Vicuna 7B (FULL) & 77.85\tiny{±0.71} & 33.92\tiny{±0.52} & 36.35\tiny{±0.57} & 22.29\tiny{±0.49} \\
    LLaMa 7B (FULL) & 74.69\tiny{±0.38} & 30.63\tiny{±0.46} & 37.04\tiny{±0.81} & 21.05\tiny{±0.27} \\
    \midrule
    Pythia-410M (LoRA) & 15.72\tiny{±0.72} & 6.58\tiny{±0.15} & 30.97\tiny{±1.15} & 2.96\tiny{±0.14} \\
    Pythia-2.8B (LoRA) & 36.56\tiny{±0.93} & 10.93\tiny{±0.17} & 83.94\tiny{±0.64} & 6.04\tiny{±0.23} \\
    Vicuna 7B (LoRA) & 42.28\tiny{±0.25} & 16.74\tiny{±0.33} & 78.32\tiny{±0.54} & 7.69\tiny{±0.47} \\
    LLaMa 7B (LoRA) & 41.76\tiny{±0.27} & 16.20\tiny{±0.18} & 80.55\tiny{±0.39} & 8.08\tiny{±0.31} \\
    LLaMa 13B (LoRA) & 48.90\tiny{±0.68} & 22.67\tiny{±0.30} & 74.55\tiny{±0.46} & 12.26\tiny{±0.18} \\
    \bottomrule
    \end{tabular}%
    }
  \label{tab:lora_exp}%
\end{table}%

\begin{figure}[!t]
    \centering
    \subfloat[Training Set]{
        \includegraphics[width=0.49\linewidth]{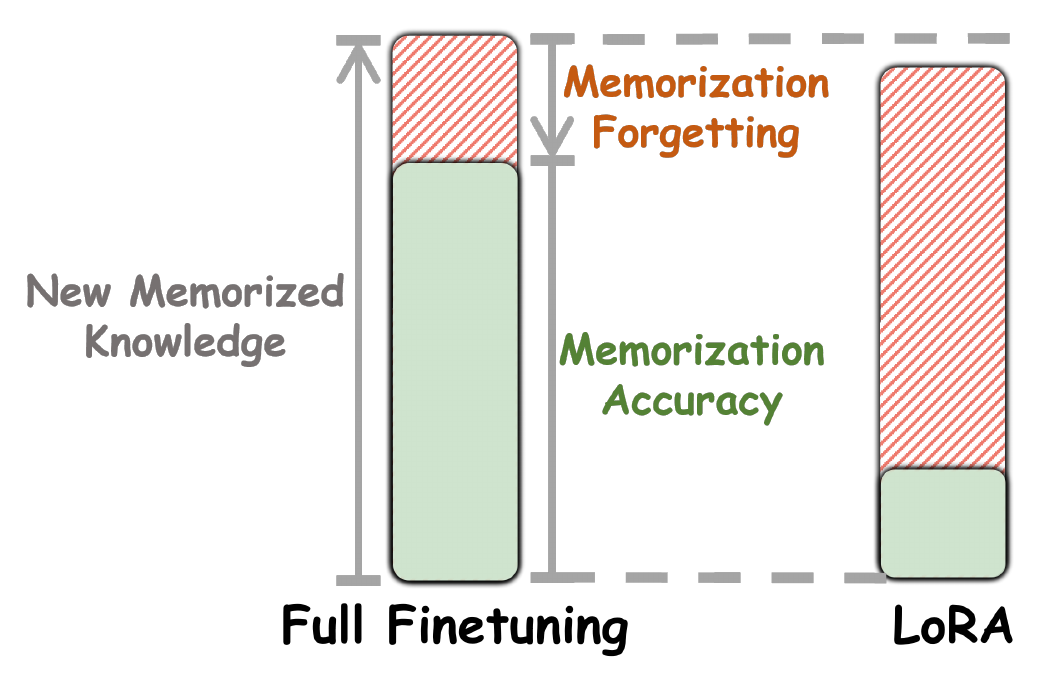}
    }
    \subfloat[Test Set]{
        \includegraphics[width=0.49\linewidth]{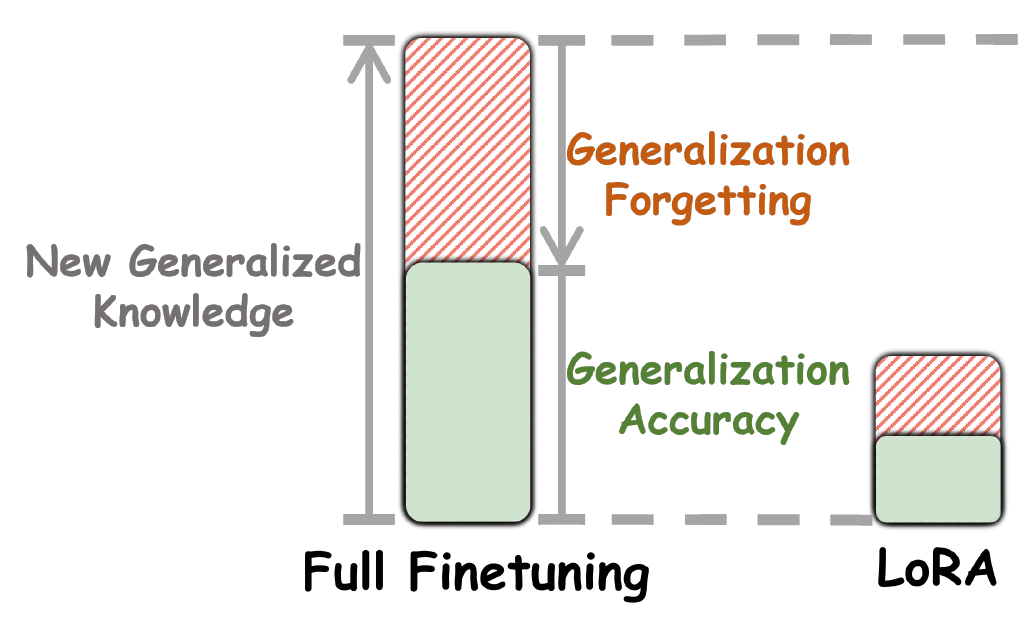}
    }
    
    \caption{Comparison of the performance between full finetuning and LoRA on (a) the training set and (b) the test set. The height represents relative performance.}
    \label{fig:lora_finetuning_comparison}
\end{figure}

\subsection{RQ3: Is LoRA a better choice than full finetuning for IL with LLMs?}
\label{sec:exp_lora}

\begin{tcolorbox}
Main Findings 3: LoRA is worse than full finetuning and may also lead to more forgetting on training data.
\end{tcolorbox}

Given the limitations of in-context learning, we turn our attention to LoRA, a method that fine-tunes only a small proportion of parameters. Recently, LoRA has been widely used for designing IL methods or as an experimental setting \cite{zheng2024lifelong}. Additionally, \citet{biderman2024lora} argue that LoRA learns less and forgets less. 

As shown in Table \ref{tab:lora_exp}, \emph{LoRA significantly limits the ability to learn new memorized or generalized knowledge} compared to full fine-tuning. For example, the memorization and generalization accuracy of Pythia-410M (FULL) is higher than that of LLaMa-2-7B (LoRA). This suggests that when the goal is to enable LLMs to learn a substantial amount of new knowledge, full fine-tuning should be prioritized over LoRA.

Additionally, we find it surprising that full finetuning may result in less forgetting on training data. As illustrated in Figure \ref{fig:lora_finetuning_comparison}, full finetuning learns more memorized and generalized knowledge than LoRA because it modifies a much larger number of parameters. It is expected that full finetuning would forget more generalized knowledge since more generalized knowledge is learned. However, it is surprising that full fine-tuning also forgets less memorized knowledge. This implies that LLMs are more resilient to forgetting when using full finetuning, highlighting the importance of investigating IL in the full finetuning settings instead of LoRA, which is widely adopted in recent IL studies \cite{yang2024moral,ren2024analyzing}.

\begin{table}[!t]
  \centering
  \caption{The performance of SOTA methods on Concept-1K. The detailed results are in Figure \ref{fig:sota_detailed}.}
  \resizebox{0.99\linewidth}{!}{
    \begin{tabular}{l|ll|ll|c}
    \toprule
    Method & MA ($\uparrow$) & GA ($\uparrow$) & MF ($\downarrow$) & GF ($\downarrow$) & Runtime (min) \\
    \midrule
    SEQ   & 58.28\tiny{±0.64} & 17.68\tiny{±0.31} & 65.19\tiny{±0.31} & 15.39\tiny{±0.16} & 27 \\
    EWC \cite{kirkpatrick2017overcoming} & 59.83\tiny{±0.62} & 18.09\tiny{±0.28} & 62.46\tiny{±0.69} & 15.07\tiny{±0.33} & 33 \\
    LAMOL\_g \cite{sun2020lamol} & 58.76\tiny{±0.53} & 15.35\tiny{±0.46} & 64.64\tiny{±0.59} & 13.61\tiny{±0.47} & 48 \\
    LAMOL\_t \cite{sun2020lamol} & 58.29\tiny{±2.17} & 15.24\tiny{±0.37} & 66.66\tiny{±2.42} & 14.05\tiny{±0.38} & 48 \\
    L2KD \cite{chuang-etal-2020-lifelong} & 28.34\tiny{±0.29} & 10.87\tiny{±0.01} & 32.45\tiny{±0.63} & 8.55\tiny{±0.42} & 91 \\
    PCLL \cite{zhao-etal-2022-prompt} & \underline{61.94\tiny{±1.41}} & \textbf{20.06\tiny{±0.42}} & 63.04\tiny{±1.57} & 16.27\tiny{±0.37} & 252 \\
    LFPT5 \cite{qin2022lfpt5} & 0.63\tiny{±0.04} & 0.84\tiny{±0.01} & 0.04\tiny{±0.03} & 0.03\tiny{±0.01} & 44 \\
    LAMOL\_KD \cite{zheng2023learn} & \textbf{72.33\tiny{±0.45}} & \underline{18.20\tiny{±0.37}} & 49.25\tiny{±0.32} & 10.61\tiny{±0.42} & 59 \\
    \midrule
    REPLAY (buffer size=2000) & 77.31\tiny{±0.22} & 22.48\tiny{±0.29} & 46.97\tiny{±1.50} & 10.57\tiny{±0.24} & 44 \\
    REPLAY (buffer size=Alll) & 99.01\tiny{±0.11} & 25.70\tiny{±0.44} & 0.70\tiny{±0.16} & 1.44\tiny{±0.88} & 110 \\
    \bottomrule
    \end{tabular}%
    }
  \label{tab:sota_exp}%
\end{table}%

\subsection{RQ4: What is the most effective and efficient method for IL of LLMs?}

\begin{tcolorbox}
Main Findings 4: Data replay remains the most effective and efficient method for IL of LLMs.
\end{tcolorbox}

Given that full finetuning and LoRA both have their own limitations, we explore what the most effective and efficient method for incremental learning of LLMs might be. Data replay is a straightforward approach to IL that stores a small number of samples from previous tasks and optimizes them jointly with new data when learning new tasks. Although numerous IL methods \cite{zheng2024lifelong} have been designed to function without data replay, we find that none of these methods achieve satisfactory performance in our settings.

We compare data replay (REPLAY) with seven SOTA rehearsal-free methods. The introduction of each method is provided in Appendix \ref{sec:appendix_baselines}. The backbone model used is Pythia-410M. Detailed descriptions of the baseline methods can be found in Appendix \ref{sec:appendix_baselines}. Figure \ref{fig:sota_detailed} (a) and (b) show the step-wise average accuracy on the training and test sets, while Figure \ref{fig:sota_detailed} (c)-(f) present memorization accuracy, generalization accuracy, memorization forgetting, and generalization forgetting, respectively.

Table \ref{tab:sota_exp} summarizes the results, indicating that although existing methods have improved sequential finetuning (SEQ), a significant performance gap remains compared to data replay with only 2000 samples (about 12\% of the total samples). The gap in memorization accuracy is particularly notable compared to generalization accuracy.

Furthermore, as shown in Figure \ref{fig:sota_detailed} (g), the training loss of the prompt-tuning-based method LFPT5 does not decrease to a low value. This indicates that merely using prompt tuning is not practical for learning new knowledge, which aligns with the findings in Section \ref{sec:exp_lora}. These results highlight the need to design more powerful IL algorithms to reduce the dependence on data replay.

\begin{figure*}[!t]
    \centering
    \subfloat[Scale vs Buffer\\ \text{[}Pretraining Step=Final\text{]}]{
        \includegraphics[width=0.23\linewidth]{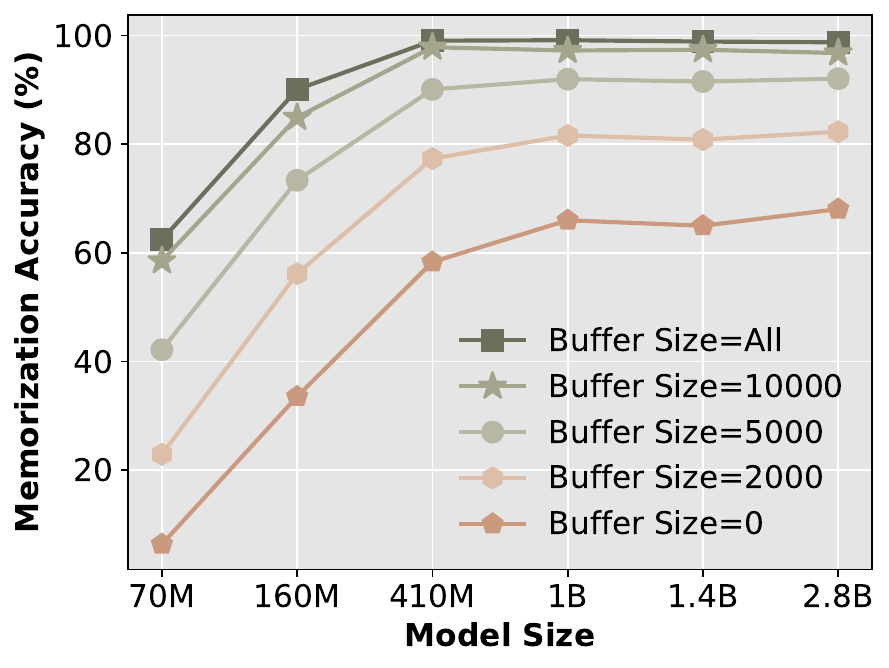}
    }
    \subfloat[Pretraining vs Scale\\ \text{[}Buffer Size=0\text{]}]{
        \includegraphics[width=0.23\linewidth]{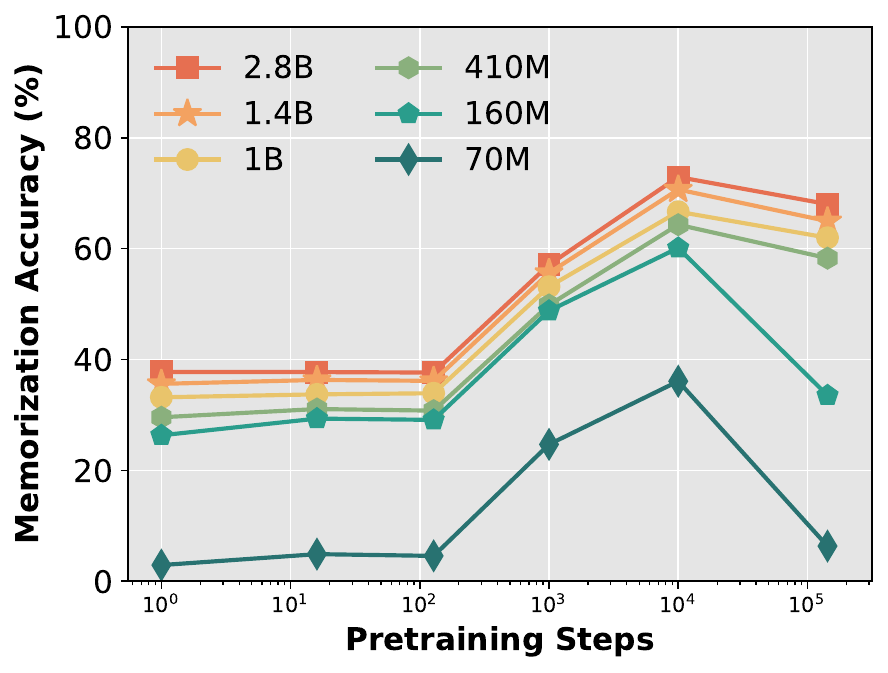}
    }
    \subfloat[Pretraining vs Scale\\ \text{[}Buffer Size=2000\text{]}]{
        \includegraphics[width=0.23\linewidth]{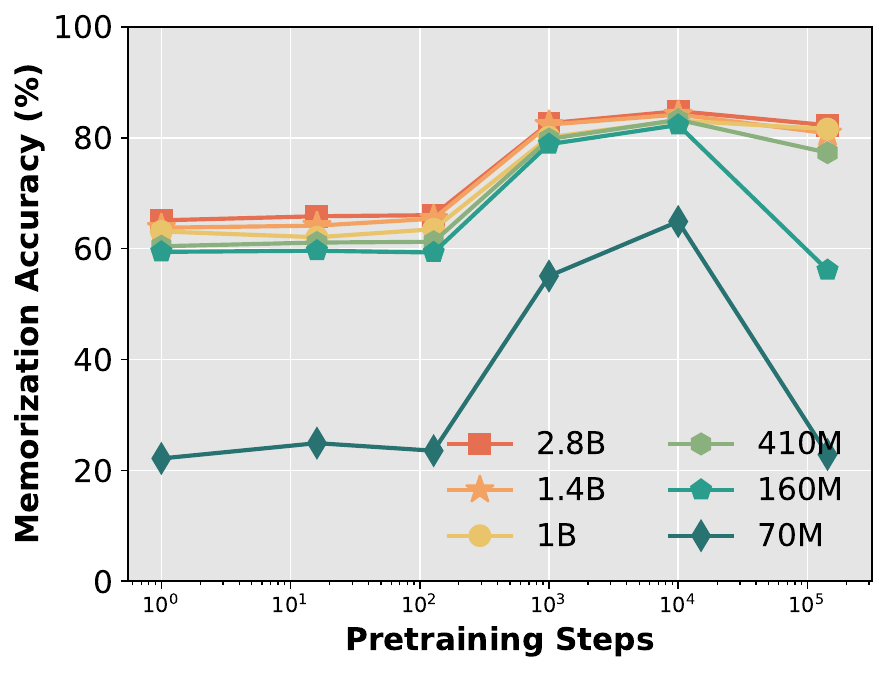}
    }
    \subfloat[Pretraining vs Scale\\ \text{[}Buffer Size=ALL\text{]}]{
        \includegraphics[width=0.23\linewidth]{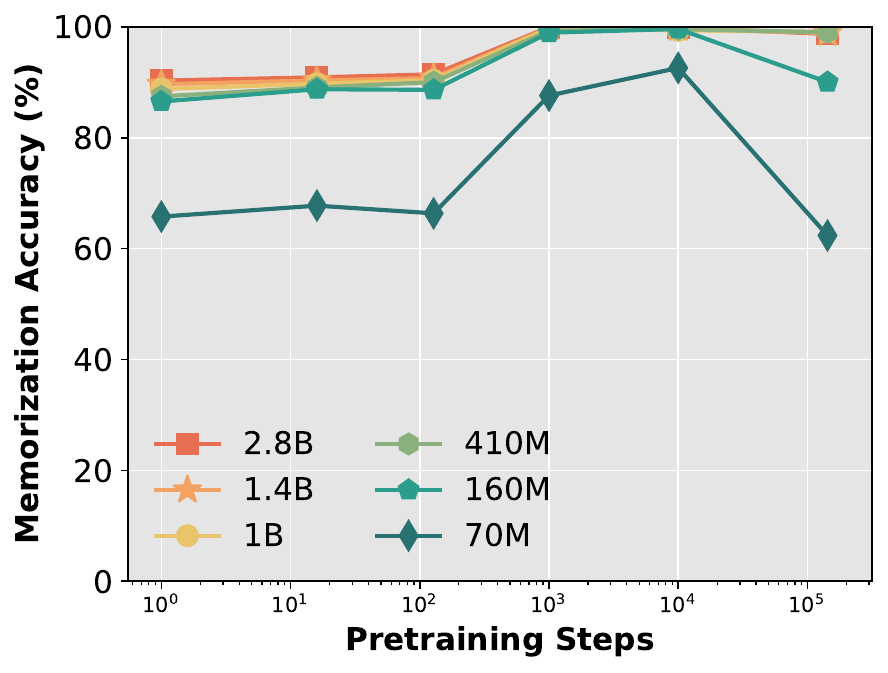}
    }
    
    \subfloat[Scale vs Buffer\\ \text{[}Pretraining Step=Final\text{]}]{
        \includegraphics[width=0.23\linewidth]{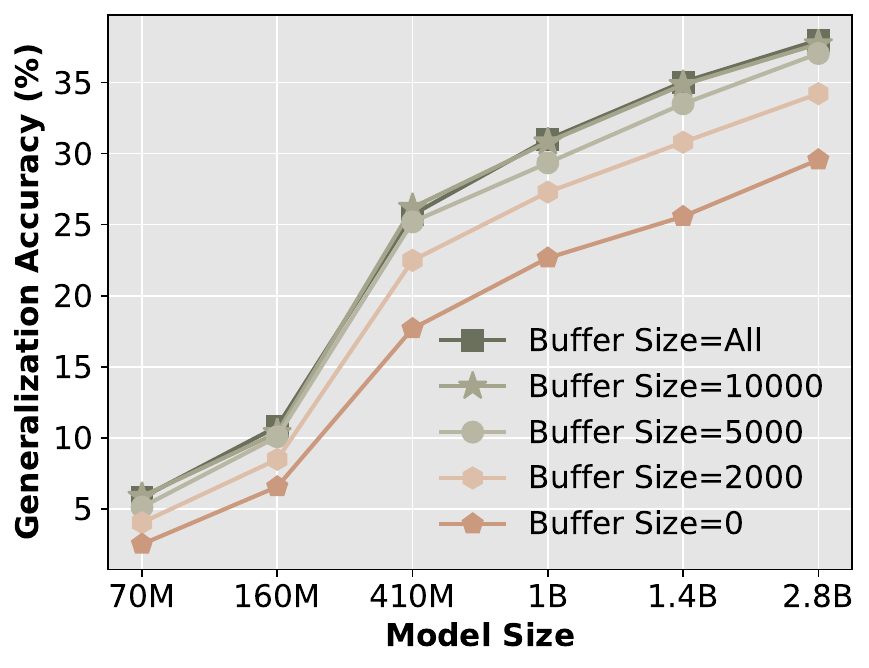}
    }
    \subfloat[Pretraining vs Scale\\ \text{[}Buffer Size=0\text{]}]{
        \includegraphics[width=0.23\linewidth]{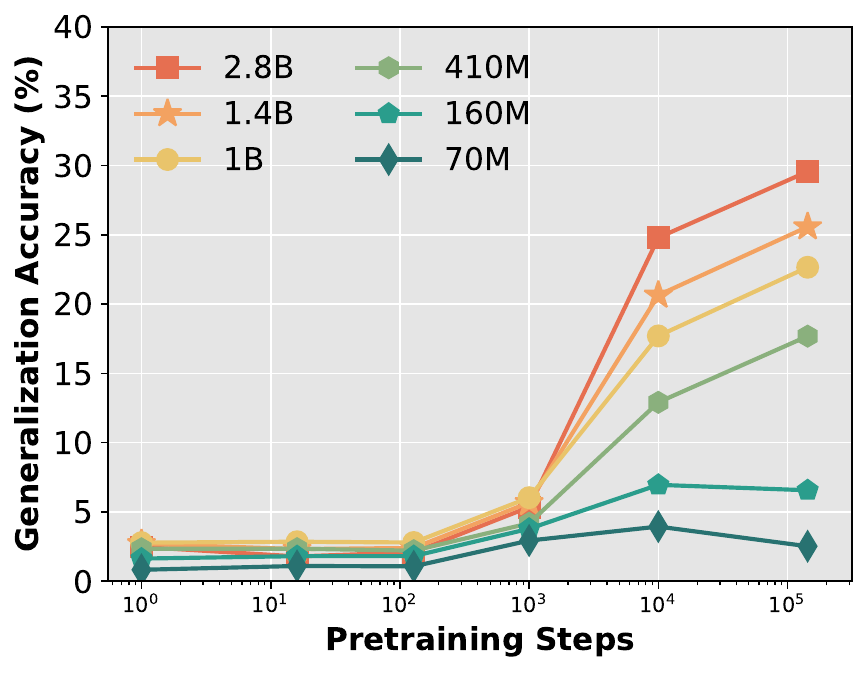}
    }
    \subfloat[Pretraining vs Scale\\ \text{[}Buffer Size=2000\text{]}]{
        \includegraphics[width=0.23\linewidth]{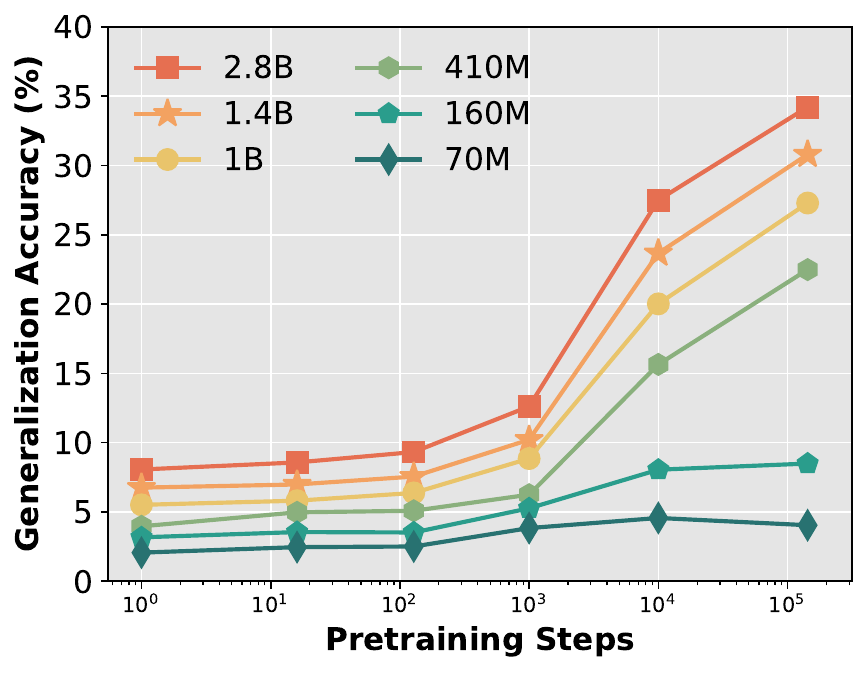}
    }
    \subfloat[Pretraining vs Scale\\ \text{[}Buffer Size=ALL\text{]}]{
        \includegraphics[width=0.23\linewidth]{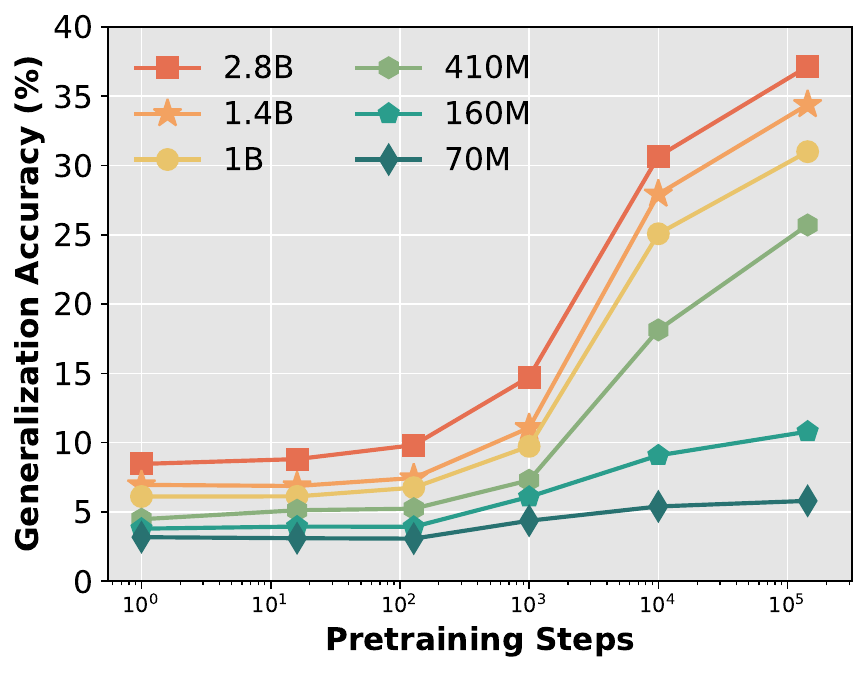}
    }
    \caption{The analysis of memorization (top row) and generalization (bottom row) accuracy on Concept-1K. The backbone model is in \{Pythia-70M, 160M, 410M, 1B, 1.4B, 2.8B\}. The pretraining step is in \{0, 16, 128, 1000, 10000, 143000 (final version)\}. Each point represents the result of IL. The detailed results with standard deveriation are provided in Table \ref{tab:scale_buffer_train}, \ref{tab:scale_buffer_test}, \ref{tab:pretrain_scale_buf0_train}, \ref{tab:pretrain_scale_buf0_test}, \ref{tab:pretrain_scale_buf2000_train}, \ref{tab:pretrain_scale_buf2000_test}, \ref{tab:pretrain_scale_buf20000_train}, and \ref{tab:pretrain_scale_buf20000_test}. }
    \label{fig:buffer_scale_pretraining_exp}
\end{figure*}

\begin{table}[!t]
  \tiny
  \centering
  \caption{The forgetting of LLMs with different scales. The pretraining step is final and buffer size is 0.}
  \resizebox{0.99\linewidth}{!}{
    \begin{tabular}{llllll}
    \toprule
          & 160M  & 410M  & 1B    & 1.4B  & 2.8B \\
    \midrule
    MF ($\downarrow$) & 76.07\tiny{±1.08} & 65.19\tiny{±0.31} & 55.36\tiny{±0.21} & 56.70\tiny{±0.24} & 51.12\tiny{±0.42} \\
    GF ($\downarrow$) & 5.35\tiny{±0.59} & 15.39\tiny{±0.16} & 18.22\tiny{±0.07} & 21.70\tiny{±0.07} & 23.97\tiny{±0.18} \\
    \bottomrule
    \end{tabular}%
    }
  \label{tab:scale_and_forgetting}%
\end{table}%

\subsection{RQ5: What is the role of model scale, buffer size, and pretraining on the IL ability of LLMs?}

\begin{tcolorbox}
Main Findings 5: Larger model scale, buffer size, and more pretraining steps all lead to better IL ability.
\end{tcolorbox}

Given that data replay is effective, we next explore how model scale, buffer size, and pretraining influence the incremental learning ability of LLMs.
The results are summarized in Figure \ref{fig:buffer_scale_pretraining_exp} and Table \ref{tab:scale_and_forgetting}. Detailed results corresponding to Figure \ref{fig:buffer_scale_pretraining_exp} can be found in Tables \ref{tab:scale_buffer_train}, \ref{tab:scale_buffer_test}, \ref{tab:pretrain_scale_buf0_train}, \ref{tab:pretrain_scale_buf0_test}, \ref{tab:pretrain_scale_buf2000_train}, \ref{tab:pretrain_scale_buf2000_test}, \ref{tab:pretrain_scale_buf20000_train}, and \ref{tab:pretrain_scale_buf20000_test}.

\noindent\textbf{Model Scale.}\quad 
The model scale determines the upper limit of generalization performance. Table \ref{tab:scale_and_forgetting} shows that as LLMs become larger, the memorization forgetting decreases while the generalization forgetting increases. This indicates that larger LLMs forget fewer training samples but more generalized knowledge, as they generalize more knowledge.

\noindent\textbf{Buffer Size.}\quad 
Figures \ref{fig:buffer_scale_pretraining_exp} (a) and (e) show that a larger buffer size or a larger LLM improves the accuracy of both memorization and generalization. However, the memorization accuracy of the 2.8B model remains unsatisfactory without a buffer. This suggests that even billion-parameter LLMs suffer from catastrophic forgetting, and data replay is an effective technique for mitigating it. Furthermore, Figures \ref{fig:buffer_scale_pretraining_exp} (b)-(d), (f)-(h) indicate that a larger buffer size improves both memorization and generalization abilities across all pretraining steps, with the improvement in memorization ability being more significant than that in generalization ability.

\noindent\textbf{Pretraining.}\quad 
Figure \ref{fig:buffer_scale_pretraining_exp} (b) demonstrates that memorization performance increases during the early stages of pretraining (step 0 - step 10000), indicating that pretraining enhances the memorization ability of LLMs for novel concepts. However, with more pretraining steps, memorization performance degrades. In contrast, Figure \ref{fig:buffer_scale_pretraining_exp} (f) shows that generalization performance increases monotonically for LLMs larger than 160M. This may be because LLMs gradually learn to extract underlying knowledge from the text during pretraining, rather than merely remembering specific texts. Additionally, especially for larger models, pretraining is more beneficial to generalization ability than memorization ability.

\begin{figure}[!t]
    \centering
    \subfloat[Sorted by MA]{
        \includegraphics[width=0.40\linewidth,height=2.5cm]{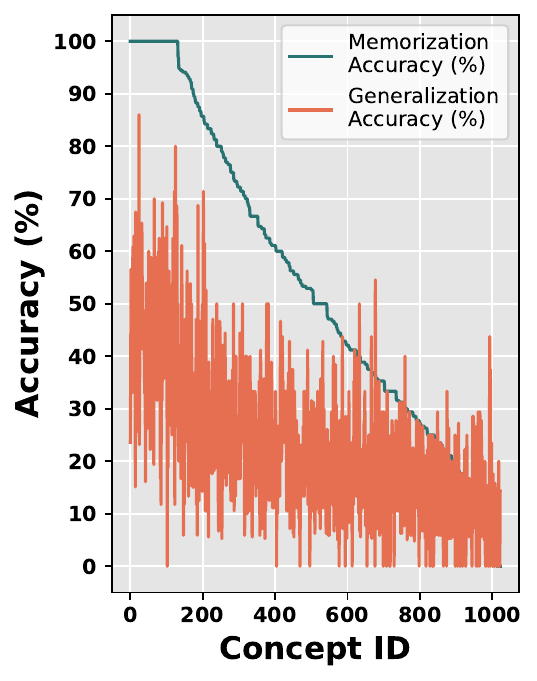}
    }
    \subfloat[Sorted by GA]{
        \includegraphics[width=0.40\linewidth,height=2.5cm]{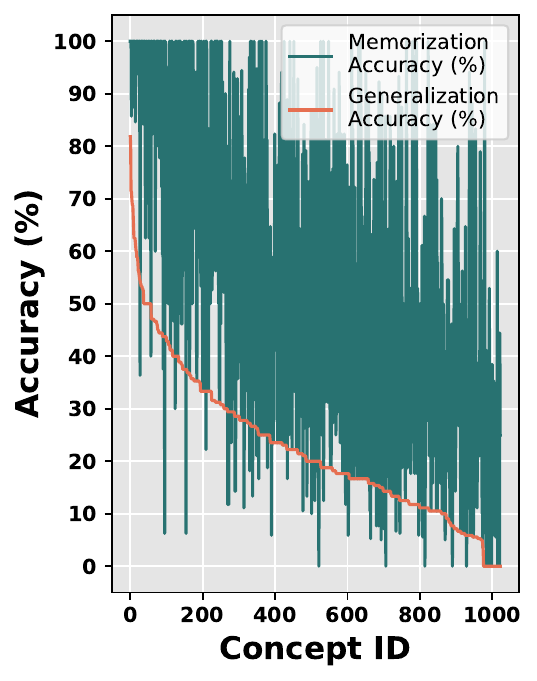}
    }
    \caption{The memorization accuracy and generalization accuracy of different concepts in Concept-1K. The concepts are sorted according to (a) memorization accuracy and (b) generalization accuracy respectively. }
    \label{fig:histogram_concept_acc}
\end{figure}

\begin{table}[!t]
  \centering
  \caption{The concepts with highest and lowest generalization accuracy.}
  \resizebox{0.99\linewidth}{!}{
    \begin{tabular}{c|c|c||c|c|c}
    \toprule
    \multicolumn{3}{c||}{Top 5} & \multicolumn{3}{c}{Bottom 5} \\
    \midrule
    Concept & GA (\%) & MA (\%) & Concept & GA (\%) & MA (\%) \\
    \midrule
    Peer-to-Peer Lending & 82.47  & 100.00  & Smart City Technologies & 0.00  & 25.54  \\
    Letters of Credit & 77.61  & 100.00  & Orbital Mechanics & 0.00  & 44.13  \\
    Streaming Services & 74.43  & 100.00  & Virtual Fitting Rooms & 0.00  & 32.86  \\
    Carbon Neutral & 71.04  & 85.71  & Mars Missions & 0.00  & 5.87  \\
    Interest Rate Hikes & 70.82  & 100.00  & Flexible Displays & 0.00  & 26.38  \\
    \bottomrule
    \end{tabular}%
    }
  \label{tab:concepts_highest_lowest_accuracy}%
\end{table}%

\subsection{RQ6: Are concepts learned equally?}

\begin{tcolorbox}
Main Findings 6: Concepts that are well-defined and concrete are easier to memorized and generalized.
\end{tcolorbox}

Finally, we explore whether LLMs learn all concepts equally. Are some concepts easier to learn? We analyze the memorization and generalization accuracy of various concepts in the Concept-1K dataset, using the LLaMa-2-7B model as the backbone. To mitigate the impact of task order, we aggregate the performance of the concepts in the fifth task after training on all tasks from 10 different task orders.

Figure \ref{fig:histogram_concept_acc} reveals a positive correlation between memorization accuracy and generalization accuracy, indicating that concepts easier to memorize are also easier to generalize, and vice versa. Our findings align with those of \citet{toneva2018empirical}, which suggest that certain examples are unforgettable and their knowledge can be better generalized across datasets.

Table \ref{tab:concepts_highest_lowest_accuracy} highlights that concepts with the highest memorization accuracy tend to be \emph{well-defined and concrete}, often related to established financial or technological terms. In contrast, concepts with the lowest memorization accuracy are often more \emph{complex, abstract, or emerging fields}, which may explain the challenges in both memorization and generalization. This disparity underscores the importance of the nature of the concepts being learned and the inherent difficulty associated with them. Our findings are consistent with those of \citet{toneva2018empirical}, which reveal that unforgettable images are easily recognizable, while the most forgotten examples exhibit more ambiguous characteristics.

\begin{figure}[!t]
    \centering
    \subfloat[After Task 1 (Train)]{
        \includegraphics[width=0.45\linewidth]{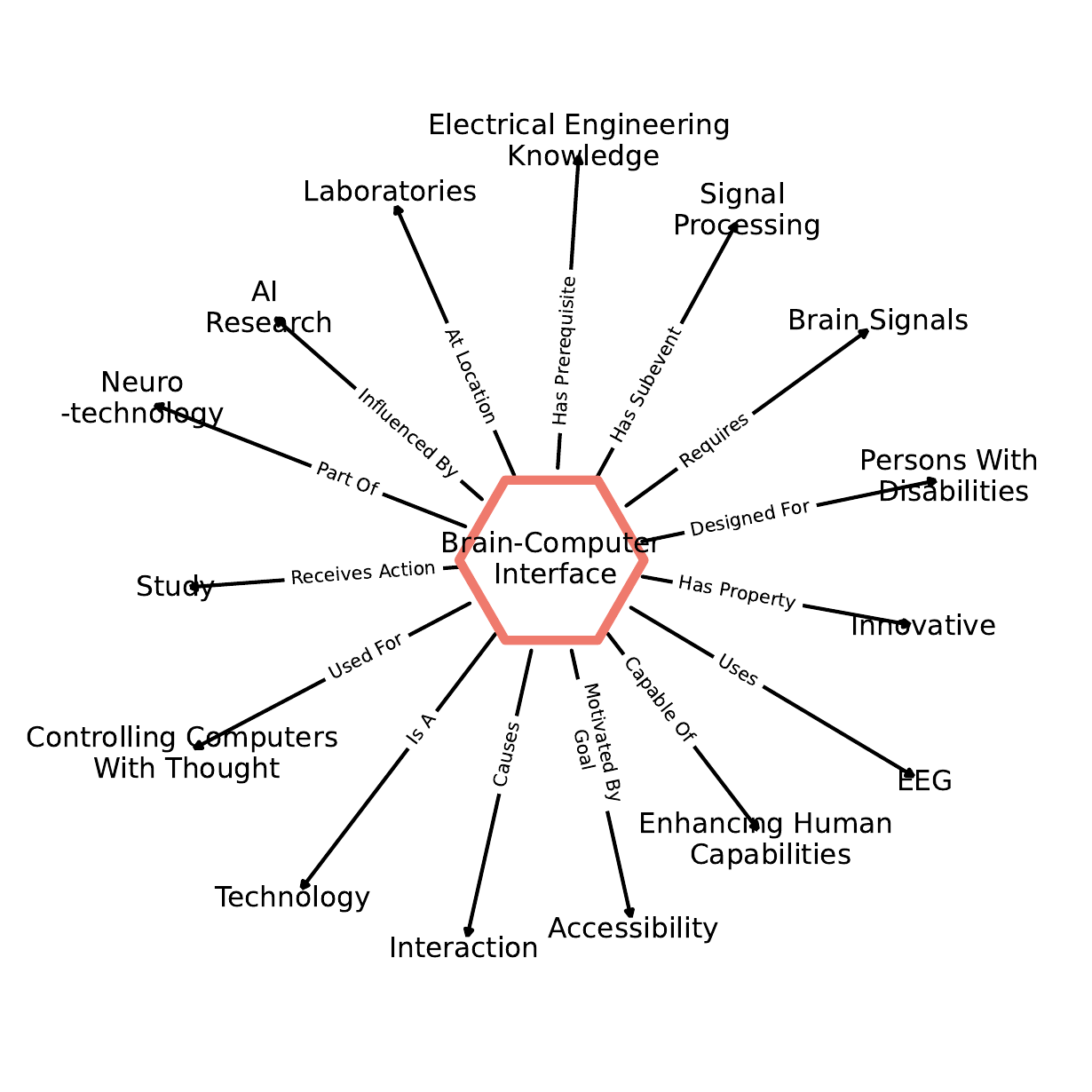}
    }
    \subfloat[After Task 10 (Train)]{
        \includegraphics[width=0.45\linewidth]{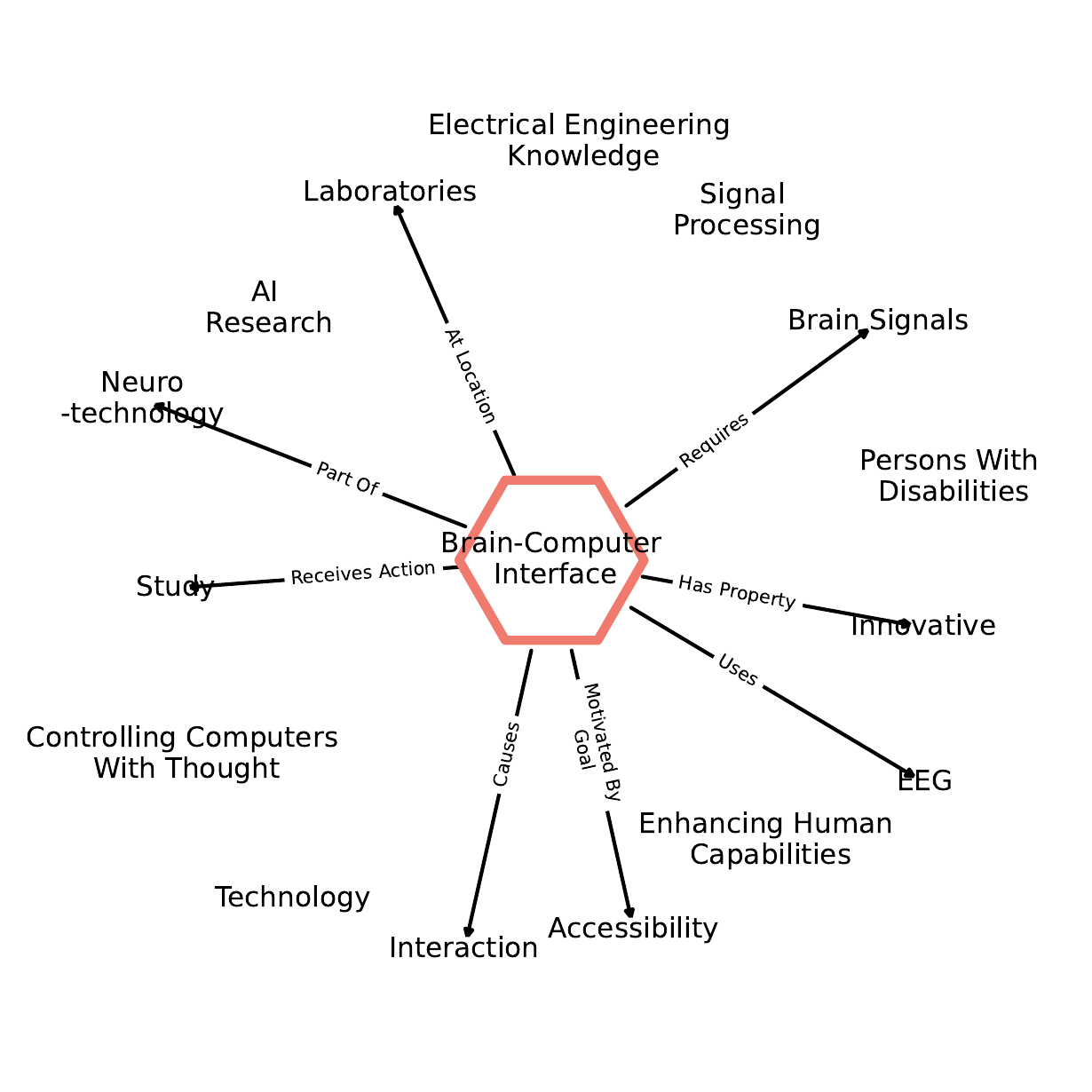}
    }

    \subfloat[After Task 1 (Test)]{
        \includegraphics[width=0.45\linewidth]{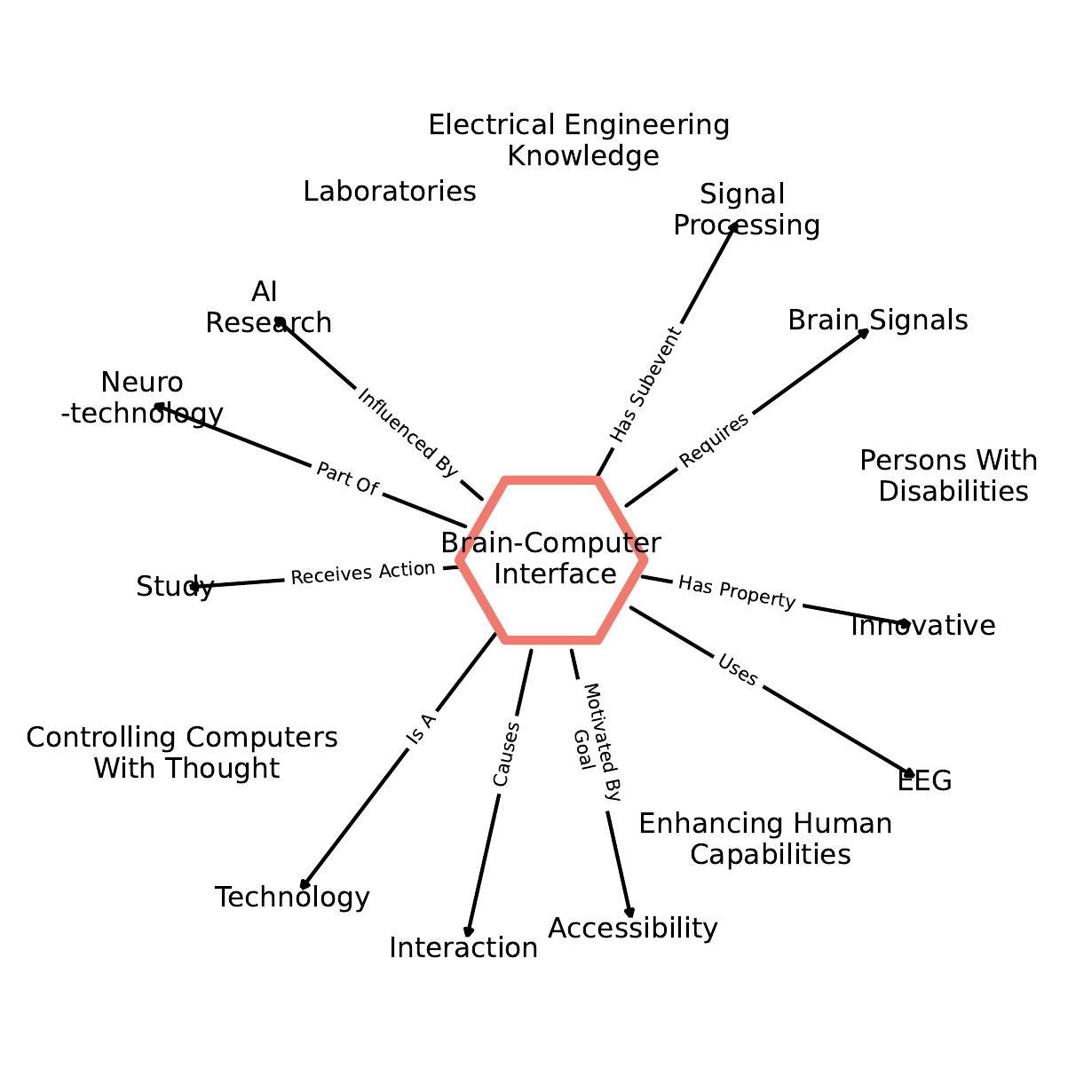}
    }
    \subfloat[After Task 10 (Test)]{
        \includegraphics[width=0.45\linewidth]{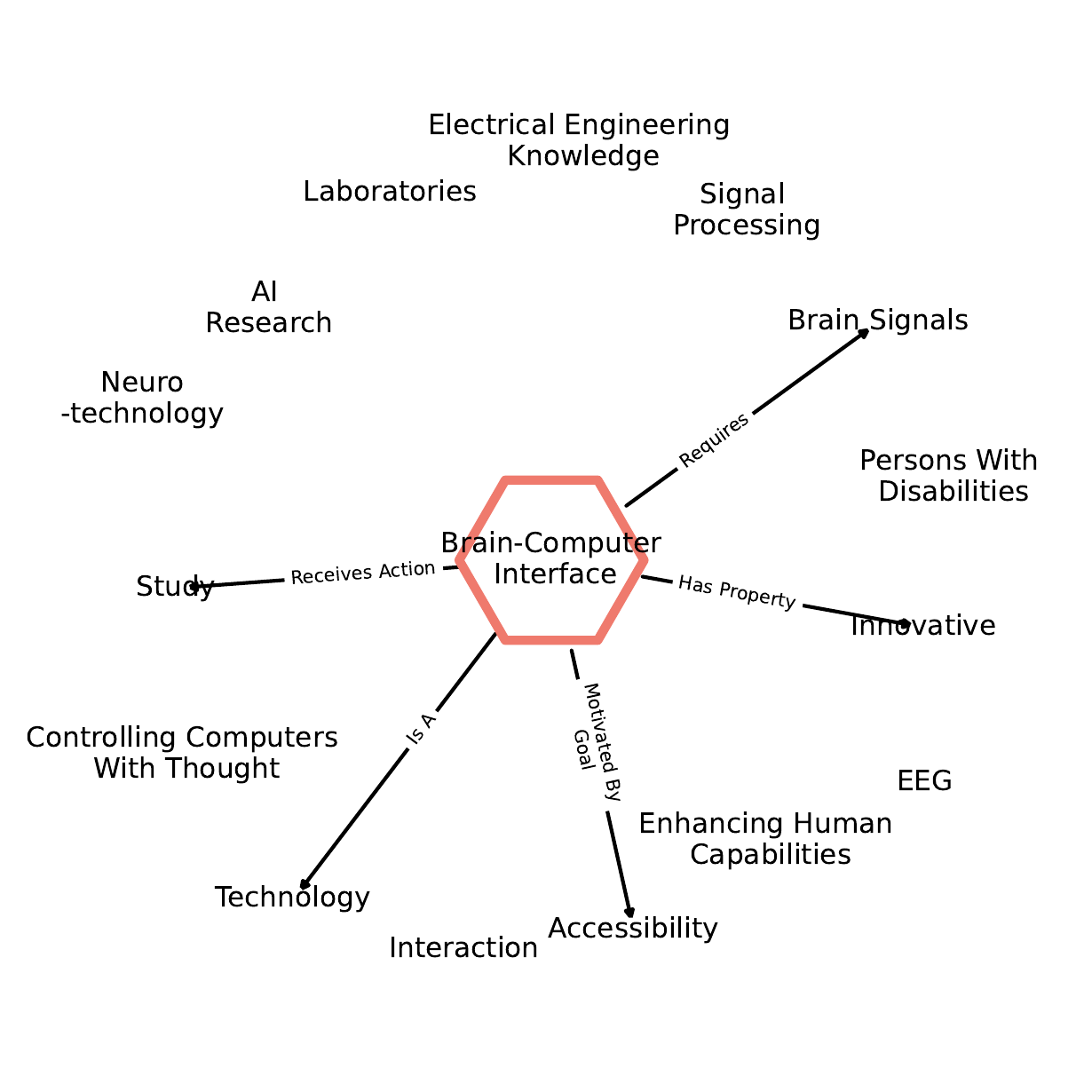}
    }
    
    \caption{The visualization of the \emph{memorized} and \emph{generalized} knowledge related to the ``Brain-Computer Interface'' in IL. The center node represents a concept, while the linked and unlinked edges indicate whether the corresponding \emph{training} and \emph{test} samples are answered correctly. The full results are in Figure \ref{fig:concept_memorize_full} and \ref{fig:concept_generalize_full}.}
    \label{fig:concept_memorize_main}
\end{figure}

Figure \ref{fig:concept_memorize_main} visualize the \emph{memorized} and \emph{generalized} knowledge of one individual concept ``Brain-Computer Interface'' and illustrate the forgetting on knowledge during IL. The full results and further discussion are provided in Appendix \ref{sec:appendix_visualization_one_concept}.

\section{Related Work}

We categorize existing studies on understanding the incremental learning ability of LLMs into three parts: (1) \emph{Understanding Forgetting}, (2) \emph{Understanding Memorization}, and (3) \emph{Applications in NLP}. Due to space limitations, the detailed discussion is provided in the Appendix \ref{sec:appendix_related_work}.

\noindent\textbf{Understanding Forgetting.}\quad
Earlier studies, such as \citet{french1999catastrophic}, assess catastrophic forgetting by measuring performance degradation on old tasks. Recently, studies \cite{taocan,zheng2023learn} use probing techniques to measure forgetting in incremental learning. \citet{zheng2023learn} uses probing techniques to show that LLMs have superior performance on evaluated datasets even before IL. Our work is inspired by \citet{zheng2023learn} and proposes a novel dataset to minimize the influence of data leakage issues.

\section{Conclusion}

In this paper, we introduce Concept-1K, a novel dataset designed to evaluate the IL capabilities of LLMs. Our comprehensive experiments reveal that LLMs still suffer from catastrophic forgetting and that LoRA, despite fine-tuning fewer parameters, limits the ability to learn and generalize new knowledge. We demonstrate that data replay is the most effective method for mitigating forgetting and highlight the significant roles of model scale, buffer size, and pretraining. These findings provide valuable insights into the strengths and limitations of LLMs in IL scenarios, offering a robust benchmark for future research.

\section*{Limitations}
There are two limitations of this research:
(1) The knowledge of Concept-1K is defined in the form of triplets, which can not cover the knowledge in a broad sense.
(2) Apart from the experiments of in-context learning, other experiments are conducted on LLMs with less than 13B parameters.
The conclusion of these experiments may not hold when finetuning SOTA LLMs such as GPT4 with more than 100B parameters.

\section*{Ethical Considerations}
The ethical considerations of our research are carefully addressed to ensure compliance with relevant standards and transparency. To this end, we provide the following clarifications:

1. \textbf{Dataset Collection}: 
Our research employs GPT4 to construct Concept-1K and filter out offensive or harmful instances. 
The use of GPT4 was consistent with their intended use.
The dataset Concept-1K is publicly available for academic and research purposes. 

2. \textbf{Reproducibility}: 
We provide a detailed setting of our experiments.
The source code, data, and scripts will all be publicly available. 
Our findings are in alignment with observed empirical outcomes.

% \section*{Acknowledgements}

\bibliography{anthology,custom}

\clearpage

\appendix

\renewcommand*\contentsname{Appendix}
\clearpage
\addtocontents{toc}{\protect\setcounter{tocdepth}{2}}
\begin{center}
    \tableofcontents
\end{center}

\section{Related Work}
\label{sec:appendix_related_work}

We categorize existing studies on understanding the incremental learning ability of LLMs into three parts: (1) \emph{Understanding Forgetting}, (2) \emph{Understanding Memorization}, and (3) \emph{Applications in NLP}. 

\subsection{Understanding Forgetting}
Earlier studies, such as \citet{french1999catastrophic,kirkpatrick2017overcoming}, assess catastrophic forgetting by measuring performance degradation on old tasks. Recently, studies \cite{davari2022probing,wu2021pretrained,chenforgetting,taocan,zheng2023learn} use probing techniques to measure forgetting in incremental learning. For example, \citet{davari2022probing} uses linear probing to reveal representation drift due to parameter updates. \citet{wu2021pretrained} conducts layer-wise probing on BERT, showing catastrophic forgetting in the top and middle layers. \citet{chenforgetting} reveals a correlation between retaining past information and new task learning efficiency through linear probing on k-shot samples. \citet{taocan} illustrates BERT's resilience to catastrophic forgetting even without buffer data. More recently, \citet{zheng2023learn} reveal that most previous work in NLP overlooks the data leakage issue and uses probing techniques to show that LLMs have superior performance on evaluated datasets even before incremental training. Additionally, they reveal that LLMs have strong anti-forgetting ability even under the sequential fine-tuning setting. Our work is inspired by \citet{zheng2023learn} and proposes a novel dataset to minimize the influence of data leakage issues, allowing for correct conclusions about the incremental learning ability of LLMs.

\subsection{Understanding Memorization}

Fewer studies \cite{carlini2021extracting, tirumala2022memorization, biderman2024emergent, boix2023transformers} explore memorization within LLMs. For instance, \citet{carlini2021extracting} discover that GPT-2 can memorize a small proportion of private information during pretraining, raising privacy concerns. \citet{tirumala2022memorization} show that larger LLMs memorize faster and have higher ``forgetting baselines''. \citet{biderman2024emergent} reveal the difficulty in predicting which training samples will be memorized by large language models. \citet{boix2023transformers} find that transformers incrementally learn new knowledge, with trained and initial weights progressively increasing in rank. These studies explore memorization from both the perspective of sentences and model weights.

In contrast, this paper studies the problem of memorization and forgetting at a more fine-grained level: the concept level. This approach addresses an underexplored research problem in the incremental learning community.

\subsection{Applications in NLP}
Many works \cite{sun2020lamol,chuang-etal-2020-lifelong,liu-huang-2023-teamwork,qiu2024incremental,zhang2023continual,shao2023class,zhang-etal-2023-continual} focus on incremental learning for various NLP tasks, assuming catastrophic forgetting in pre-trained language models and designing techniques to mitigate it. These tasks include text classification \cite{sun2020lamol,chuang-etal-2020-lifelong}, relation extraction \cite{liu-huang-2023-teamwork}, named entity recognition \cite{qiu2024incremental,zheng-etal-2022-distilling,zhang2023continual}, intent classification \cite{shao2023class}, and machine translation \cite{zhang-etal-2023-continual}. 

We refer readers to the survey \cite{zheng2024lifelong} for more applications of incremental learning in NLP tasks. The proposed Concept-1K differs substantially from the aforementioned NLP tasks. It is more challenging and provides better explainability at the concept level.

\section{Data Leakage in IL of LLMs}
\label{sec:appendix_data_leakage}

The data leakage issue in NLP is often implicit. Unlike computer vision, where pretraining involves explicit category information, NLP pretraining is self-supervised and lacks clear categorical distinctions that can be easily compared between the pretraining corpus and downstream datasets. This makes it challenging to detect and address data leakage. Therefore, we urge future studies to exercise greater caution regarding data leakage in the IL of LLMs.

\subsection{Data Leakage in Classification Tasks}

The issue of data leakage in classification tasks for IL with LLMs has recently been highlighted by \cite{zheng2023learn}. Their extensive study revisits over 20 IL methods across four key classification tasks: Text Classification, Intent Classification, Relation Extraction, and Named Entity Recognition. One core finding of \cite{zheng2023learn} is that LLMs, such as BERT and GPT-like models, exhibit high probing performance even before they are incrementally trained on specific downstream tasks. This high initial performance suggests that these models already possess substantial knowledge relevant to the classification tasks due to their extensive pre-training on diverse corpora. Consequently, when these LLMs are evaluated under IL settings, the incremental learning of new tasks may, in fact, be leveraging pre-existing knowledge rather than genuinely incremental learning.

This phenomenon leads to misleading conclusions about the effectiveness of various IL methods \cite{sun2020lamol,chuang-etal-2020-lifelong,liu-huang-2023-teamwork,qiu2024incremental,zhang2023continual,shao2023class,zhang-etal-2023-continual}. The high probing performance before task-specific training indicates that the models are not learning incrementally as assumed but rather recalling previously acquired knowledge. Therefore, many IL studies in the context of classification tasks suffer from data leakage, as the benchmark tasks are not truly novel to the LLMs. Addressing this issue requires carefully designed benchmarks that ensure the novelty and exclusivity of the knowledge being tested, a challenge we aim to tackle with our Concept-1K dataset.

\subsection{Data Leakage in Generation Tasks}

Data leakage poses a significant challenge in IL for generation tasks, which are often more specific and diverse compared to classification tasks. For example, in the IL setting described by \citet{scialom-etal-2022-fine}, the task sequence includes Text Simplification, Headline Generation with Constraints, Haiku Generation, Covid QA, Inquisitive Question Generation, Empathetic Dialogue Generation, Explanation Generation, and Twitter Stylometry. Despite the specificity and diversity of these tasks, data leakage remains a concern because LLMs are trained on extensive corpora from the internet and often undergo supervised finetuning on dialogue data \cite{OpenAI2023GPT4TR}. This pre-training on vast amounts of internet data means that LLMs might already possess significant knowledge relevant to these generation tasks.

Moreover, the data leakage issue in generation tasks is implicit and easy to overlook. Unlike classification tasks, where techniques like probing \cite{zheng2023learn} can measure LLM performance before training, generation tasks lack such straightforward methods to assess initial model capabilities. This makes it challenging to ascertain how much new knowledge is genuinely being learned during IL versus what the model is simply recalling from its pre-trained knowledge base. 

Pioneering work investigating the IL ability of LLMs found that the T0 model \cite{sanh2021multitask} barely forgets when learning new tasks. This suggests that the T0 model may have already acquired the ability to solve multiple generation tasks with appropriate input prompts before explicit training on them. Detailed results are presented in Figure 2 of \citet{sanh2021multitask}.

Another study by \citet{zhang2023citb} defines generation tasks using different instructions, a paradigm they call continual instruction tuning. They train the T5 model \cite{raffel2020exploring} sequentially on 19 tasks, achieving 35.7\% performance, while jointly training on these tasks yields 42.1\%. The close gap between upper bound and lower bound performance indicates minimal forgetting and suggests potential data leakage. Detailed results and experimental settings are provided in Tables 1 and 2 of \citet{zhang2023citb}.

Although recent studies \cite{modulesapt,guo2024q,huang2024mitigating,yang2024moral,peng2024scalable,ren2024analyzing} claim that forgetting is serious in continual instruction tuning, all of them utilize parameter-efficient finetuning techniques such as LoRA \cite{hu2021lora} or prompt tuning \cite{lester-etal-2021-power}. As shown in our experiments in Section \ref{sec:exp_lora}, the IL ability of LoRA and full finetuning differ substantially. LoRA significantly limits the ability to learn new concepts compared to full finetuning, leading to limited new knowledge acquisition and faster forgetting on training samples. Therefore, their IL settings may not accurately reflect the true IL ability of LLMs.

\section{Comparison with Existing Incremental Learning Setting}
\label{sec:appendix_comparison_scenario}

There are three popular IL settings which are widely adopted in the literature of computer vision: \emph{class-incremental learning} (CIL), \emph{task-incremental learning} (TIL), and \emph{continual pretraining} (CPT).
However, none of them are appropriate to evaluate the IL ability of LLMs.

\subsection{Class-Incremental Learning}
CIL is designed for classification tasks such as text classification, and its goal is to learn new classes incrementally.
On the one hand, SOTA LLMs with billion-level parameters are overqualified for the above classification tasks with only dozens of categories.
On the other hand, the pretraining corpus is likely to contain the knowledge required for the classification tasks (data leakage issue).
As shown empirically by \cite{zheng2023learn}, sequential training frozen LLMs with expanding classifiers yields comparable or even superior performance with SOTA IL methods.

\subsection{Task-Incremental Learning}
TIL aims to learn new tasks incrementally \cite{sun2020lamol,qin2022lfpt5}.
Apart from the data leakage issue, the diversity of tasks and orders across research makes it difficult to readily and fairly compare IL algorithms.

\subsection{Continual Pretraining}
The last scenario CPT aims at continual pretraining models on the corpus from different domains.
However, the evaluation relies on the performance of downstream tasks, where we can hardly identify what knowledge is learned or forgotten.

\subsection{Summary}
In this paper, we consider a novel IL scenario called \emph{Instance-level Incremental Learning} (IIL). 
Unlike the IL scenario mentioned above, IIL regards each concept as an instance and is more practical and challenging for existing LLMs.
Specifically, we are motivated by the human learning process and expect LLMs to learn new concepts incrementally without forgetting.
For example, in Figure \ref{fig:illustraion_iil}, humans can learn new concepts that are constantly emerging, such as ``Metaverse'' and ``Quantum Computing''.
After learning more concepts such as ``Web3.0'' and ``Non-Fungible Token'', humans will not immediately forget the previously learned concepts such as ``Metaverse''.

\section{Introduction of Domains in Concept-1K}
\label{sec:appendix_domain_introduction}

The domains in Concept-1K are introduced as follows:

\subsection{Technology Domain}
This domain focuses on both cutting-edge and widely applied technologies. Cutting-edge technologies include artificial intelligence, blockchain, quantum computing, etc., while widely applied technologies encompass cloud computing, the Internet of Things, and more.
    
\subsection{Economic Domain} This domain highlights economic trends and emerging economic concepts. Economic trends cover topics such as digital currency and globalization, whereas emerging concepts include quantitative computing, electronic wallets, peer-to-peer (P2P) networks, and others.
    
\subsection{Education Domain} This domain emphasizes emerging educational technologies and concepts. Technologies such as remote learning and online courses, along with concepts like bilingual education, social education, and lifelong learning, are included.
    
\subsection{Environmental Domain} This domain centers on global environmental issues and green technologies. Topics include climate change, environmental protection, and sustainable energy, as well as technologies like green roofs, shared bicycles, and solar panels.
    
\subsection{Cultural Domain} This domain focuses on diversity and inclusion, and digital media and arts. Diversity and inclusion cover multiculturalism, gender equality, and social inclusion, while digital media and arts include digital art, social media trends, and online communities.
    
\subsection{Health and Medical Domain} This domain is dedicated to emerging medical technologies and public safety. It covers CRISPR gene editing technology, the application of artificial intelligence in medical diagnosis, telemedicine services, wearable health monitoring devices, and concepts related to vaccine development, disease monitoring and prevention strategies, and promoting public health awareness.

\section{Concept Selection Criterion}
\label{sec:appendix_concept_selection_criterion}

In constructing Concept-1K, we carefully selected concepts based on the following criteria to ensure the relevance, novelty, and richness of the learning material:

\subsection{Length criterion} The concept words should not exceed three words in length. This encourages the model to focus on significant and concise terms within the domain, facilitating efficient learning and ensuring that the concepts provide a rich source of information. Shorter concepts are easier to manage and help avoid potential confusion that may arise from overly complex or verbose terms.

\subsection{Timeliness criterion} The chosen concept words should preferably be those that emerged after January 2022. This criterion ensures that the general pre-trained models have not yet encountered and learned these concepts and the associated knowledge. By selecting recent concepts, we aim to test the true incremental learning capabilities of LLMs, avoiding biases introduced by prior knowledge.

\subsection{Trend criterion} We focus on concepts that are currently receiving widespread attention in academia, industry, and the media. This ensures that the selected concepts are not only relevant and contemporary but also significant and impactful in their respective fields. By choosing trending concepts, we can better gauge the models' ability to learn and adapt to the latest advancements and discussions in various domains.

\begin{table*}[htbp]
  \centering
  \caption{The statistics of the LLMs used in this paper. $\dagger$: Non-embedding parameters according to \citet{biderman2023pythia}.}
  \resizebox{\linewidth}{!}{
        \begin{tabular}{llllll}
    \toprule
    \textbf{Model Class} & \textbf{Pretrained Weights} & \textbf{Parameters} & \textbf{Layers} & \textbf{Hidden Dim} & \textbf{Link} \\
    \midrule
    \multirow{6}[2]{*}{GPT-NeoX} & Pythia-70m & 19M$^{\dagger}$   & 6     & 512   & \href{https://huggingface.co/EleutherAI/pythia-70m-deduped}{Link} \\
          & Pythia-160m & 85M$^{\dagger}$   & 12    & 768   & \href{https://huggingface.co/EleutherAI/pythia-160m-deduped}{Link} \\
          & Pythia-410m & 302M$^{\dagger}$  & 24    & 1024  & \href{https://huggingface.co/EleutherAI/pythia-410m-deduped}{Link} \\
          & Pythia-1b & 805M$^{\dagger}$  & 16    & 2048  & \href{https://huggingface.co/EleutherAI/pythia-1b-deduped}{Link} \\
          & Pythia-1.4b & 1.21B$^{\dagger}$ & 24    & 2048  & \href{https://huggingface.co/EleutherAI/pythia-1.4b-deduped}{Link} \\
          & Pythia-2.8b & 2.52B$^{\dagger}$ & 32    & 2560  & \href{https://huggingface.co/EleutherAI/pythia-2.8b-deduped}{Link} \\
    \midrule
    \multirow{3}[2]{*}{LLaMa} & llama-7b-hf & 7B    & 32    & 4096  & \href{https://github.com/facebookresearch/llama/tree/llama_v1}{Link} \\
          & \multicolumn{1}{p{10em}}{vicuna-7b-v1.1} & 7B    & 32    & 4096  & \textcolor[rgb]{ .02,  .388,  .757}{\href{https://huggingface.co/lmsys/vicuna-7b-v1.1}{Link}} \\
          & llama-2-13b-hf & 13B   & 40    & 5120  & \href{https://github.com/facebookresearch/llama}{Link} \\
    \bottomrule
    \end{tabular}%
    }
  \label{tab:backbone_statistics}%
\end{table*}%

\section{Experimental Settings}
\label{sec:appendix_experimental_settings}

\subsection{Backbones}
We use the Pythia suite \cite{biderman2023pythia} and other popular open-source models, including LLaMa and Vicuna, for our experiments. Pythia is based on GPT-NeoX \cite{black-etal-2022-gpt} and includes 8 model sizes and 154 pre-training checkpoints, facilitating research in interpretability and learning dynamics. The statistics of the 9 LLMs used in this paper are summarized in Table \ref{tab:backbone_statistics}. We download the pre-trained weights from Huggingface \citep{wolf2019huggingface}.

\subsection{Implementation Details}
We sort the concepts alphabetically and shuffle the order using random seed 1. The maximum input and output lengths are set to 32 and 10, respectively. The batch size is 32, and the learning rate for the LLMs is $1\times 10^{-5}$. We use the AdamW optimizer \cite{loshchilov2018fixing}. For LLMs with more than 1B parameters, we use A800 GPUs, while smaller LLMs run on RTX3090 GPUs. Each experiment is repeated three times, and we report the average and standard deviations. Additionally, we search for the best hyper-parameters for each baseline method.

For the experiment in Section \ref{sec:exp_icl}, we use ``gpt-3.5-turbo'' and ``gpt-4-turbo'' for GPT-3.5 and GPT-4, respectively. Given the high cost of evaluating the entire Concept-1K dataset, we randomly sample 500 instances for both GPT-3.5 and GPT-4. The outputs and targets of GPT-3.5 and GPT-4 on these 500 instances are provided in the supplementary material.

\subsection{Input Prompt}
We use the following input prompt for training and testing Concept-1K:
\begin{align*}
    &\text{Question: \{Question\}}& \\
    &\text{Short Answer: \{Answer\}}&,
\end{align*}
where \{Question\} and \{Answer\} represent the question and the target output, respectively.
For the experiment in Section \ref{sec:exp_icl}, the prompts are provided in Tables \ref{tab:prompt_1_shot} and \ref{tab:prompt_5_shot}.

\begin{table*}[htbp]
\small
  \centering
  \caption{Prompt for In-Context Learning with 1 shot demonstration. \{Question i\} and \{Answer i\} represent the question and the target output of the $i$-th demonstration training sample respectively. \{Test Question\} represents the test question.}
  \resizebox{\linewidth}{!}{
    \begin{tabular}{l}
    \toprule
    \multicolumn{1}{p{33.875em}}{I will provide some knowledge as follows:}  \\
    Question: \{Question 1\} \\
    Short Answer: \{Answer 1\}  \\
    Please answer the following question according to the above knowledge:  \\
    Question: \{Test Question\} \\
    Short Answer:  \\
    \bottomrule
    \end{tabular}%
    }
  \label{tab:prompt_1_shot}%
\end{table*}%

\begin{table*}[htbp]
\small
  \centering
  \caption{Prompt for In-Context Learning with 5 shot demonstrations. \{Question i\} and \{Answer i\} represent the question and the target output of the $i$-th demonstration training sample respectively. \{Test Question\} represents the test question.}
  \resizebox{\linewidth}{!}{
    \begin{tabular}{l}
    \toprule
    \multicolumn{1}{p{33.875em}}{I will provide some knowledge as follows:}  \\
    Question: \{Question 1\} \\
    Short Answer: \{Answer 1\}  \\
    Question: \{Question 2\} \\
    Short Answer: \{Answer 2\} \\
    Question: \{Question 3\} \\
    Short Answer: {Answer 3}  \\
    Question: \{Question 4\} \\
    Short Answer: \{Answer 4\}  \\
    Question: \{Question 5\} \\
    short Answer: \{Answer 5\} \\
    Please answer the following question according to the above knowledge:  \\
    Question: \{Test Question\} \\
    Short Answer:  \\
    \bottomrule
    \end{tabular}%
    }
  \label{tab:prompt_5_shot}%
\end{table*}%

\section{Introduction of Baseline Methods}
\label{sec:appendix_baselines}

The introduction of the baseline methods is as follows:
\begin{itemize}
    \item \textbf{SEQ}: Sequential fine-tuning (SEQ) is considered the lower bound of incremental learning.
    \item \textbf{REPLAY}: Experience replay stores representative old samples and jointly optimizes both old and new samples when learning new tasks. This is a practical and popular technique in incremental learning.
    \item \textbf{EWC} \cite{kirkpatrick2017overcoming}: Elastic Weight Consolidation (EWC) is a regularization-based method where the weight of each parameter is determined by the diagonal of the Fisher information matrix. The regularization loss weight is searched within \{5$\times10^3$, 1$\times10^4$, 5$\times10^4$, 1$\times10^5$, 1$\times10^6$, 1$\times10^7$\}.
    \item \textbf{LAMOL} \cite{sun2020lamol}: LAMOL trains LLMs with question-answering and generative objectives, generating pseudo-samples before learning each new task for data replay. The generation loss weight is $\lambda=0.25$, and the proportion of pseudo-samples is $\gamma=0.20$. There are two variants: LAMOL\_t and LAMOL\_g, differing by whether a task-specific token is used for generation.
    \item \textbf{L2KD} \cite{chuang-etal-2020-lifelong}: L2KD adds a knowledge distillation target based on LAMOL, with the teacher model trained from scratch. We implemented the word-level variant as it performs best on text classification tasks.
    \item \textbf{LAMOL\_KD} \cite{zheng2023learn}: LAMOL\_KD utilizes knowledge distillation based on LAMOL\_t. Unlike L2KD, the teacher model in LAMOL\_KD is trained on all previous tasks. New data are used to learn the LAMOL objectives, and pseudo data are used for word-level knowledge distillation as a regularization term.
    \item \textbf{PCLL} \cite{zhao-etal-2022-prompt}: PCLL combines the concepts of variational autoencoders and word-level knowledge distillation with the objectives of LAMOL.
    \item \textbf{LFPT5} \cite{qin2022lfpt5}: LFPT5 learns only soft prompts for each new task. The training objective is the same as LAMOL. The number of soft prompt tokens is 10.
    \item \textbf{LoRA} \cite{hu2021lora}: LoRA trains a small proportion of parameters of LLMs. We set the rank $r=8$ and the scaling parameter $\alpha=16$. We use the LoRA implementation from the PEFT library \cite{peft}.
\end{itemize}

Some IL methods are not compared as they are not applicable in the IIL scenario. For example, Progressive Prompt \cite{razdaibiedina2023progressive} requires task IDs during both training and inference stages. VAG \cite{shao-etal-2023-class} requires storing the vocabulary of class labels and does not apply to generation tasks without class labels. Additionally, prompt-based IL methods such as L2P \cite{wang2022learning} are not suitable for generation tasks.

\begin{figure*}[htbp]
    \centering
    \subfloat[After Task 1]{
        \includegraphics[width=0.33\linewidth]{fig/visualization/memorize_0.pdf}
        \label{fig:concept_memorize_full_task_1}
    }
    \subfloat[After Task 2]{
        \includegraphics[width=0.33\linewidth]{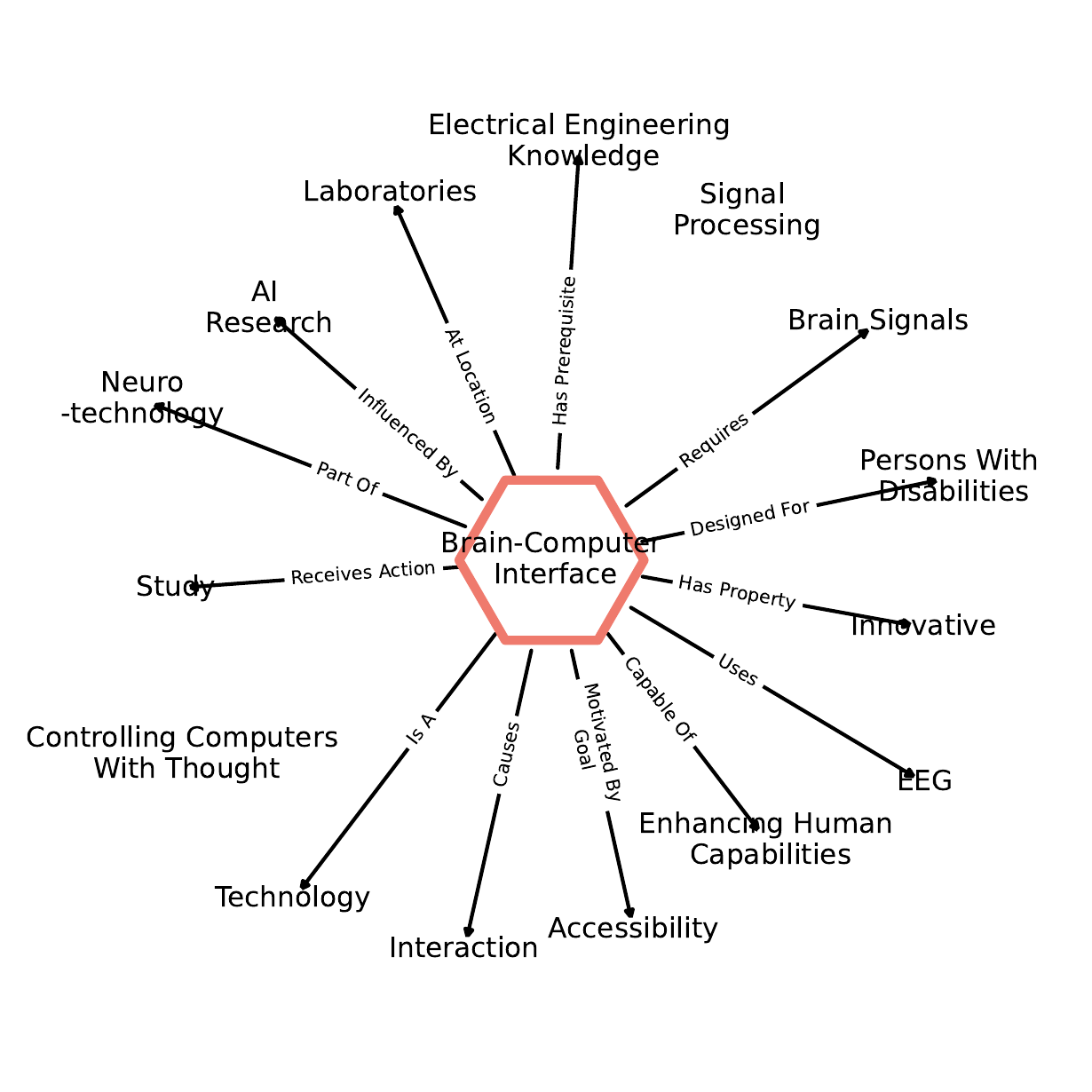}
        \label{fig:concept_memorize_full_task_2}
    }
    \subfloat[After Task 4]{
        \includegraphics[width=0.33\linewidth]{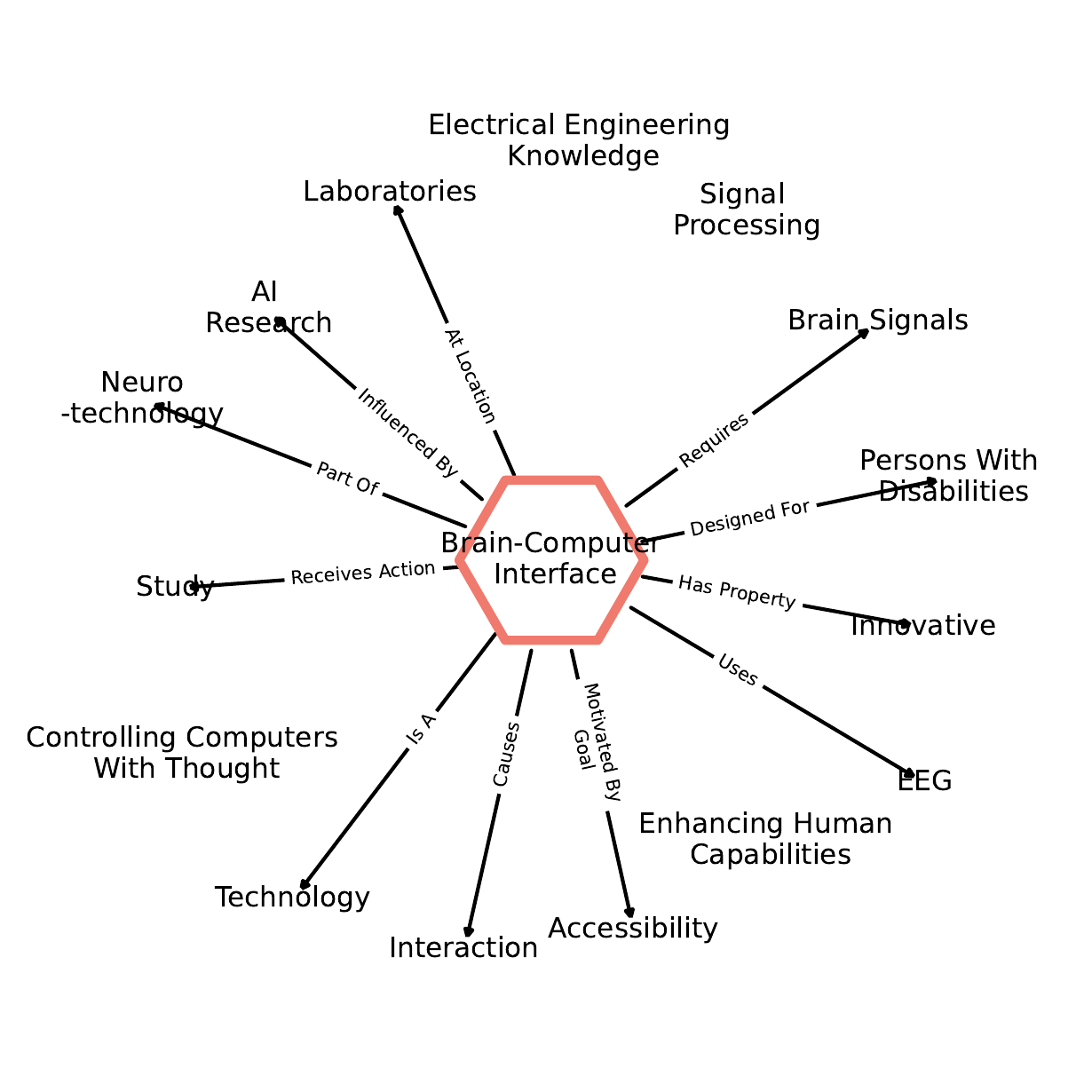}
    }
    
    \subfloat[After Task 6]{
        \includegraphics[width=0.33\linewidth]{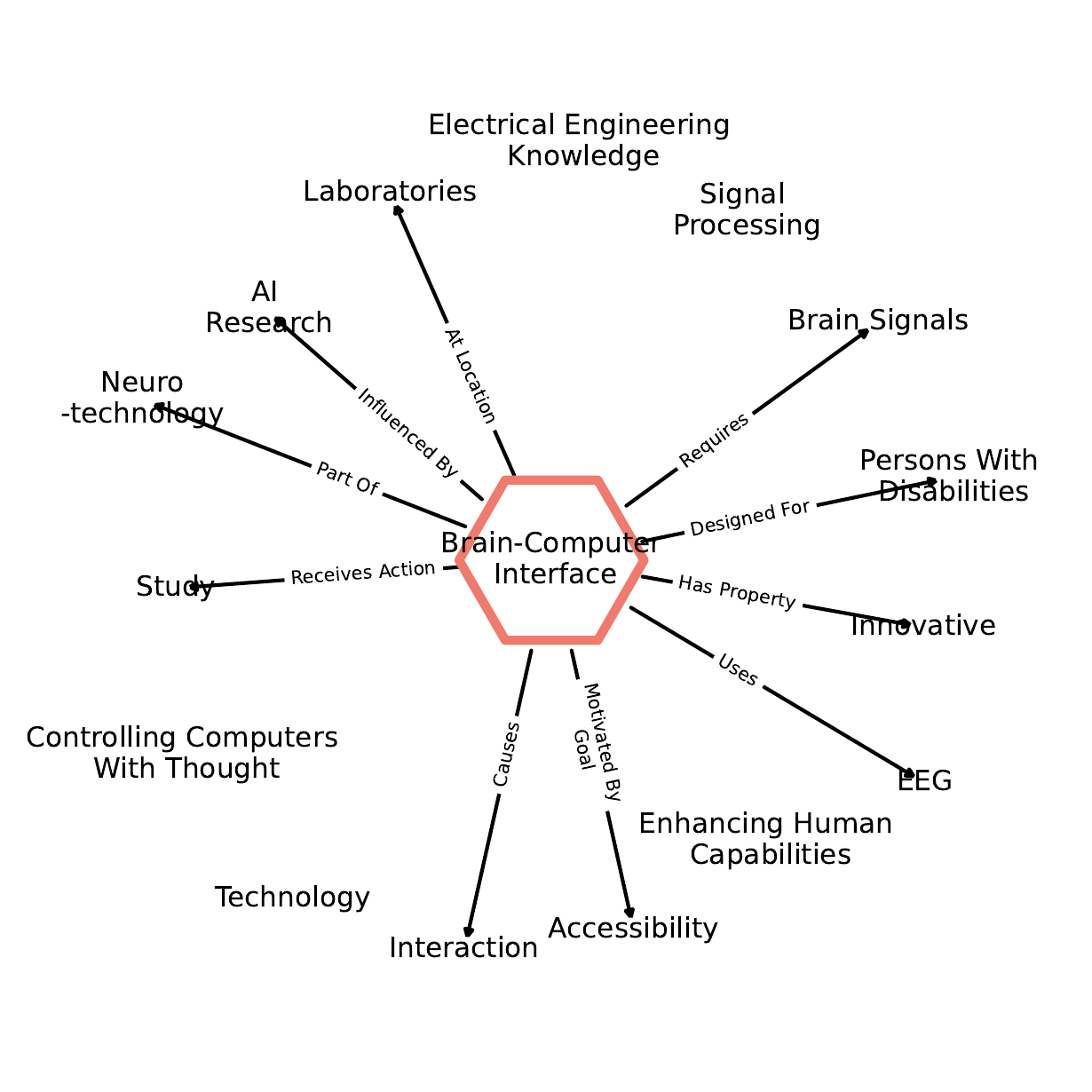}
    }
    \subfloat[After Task 8]{
        \includegraphics[width=0.33\linewidth]{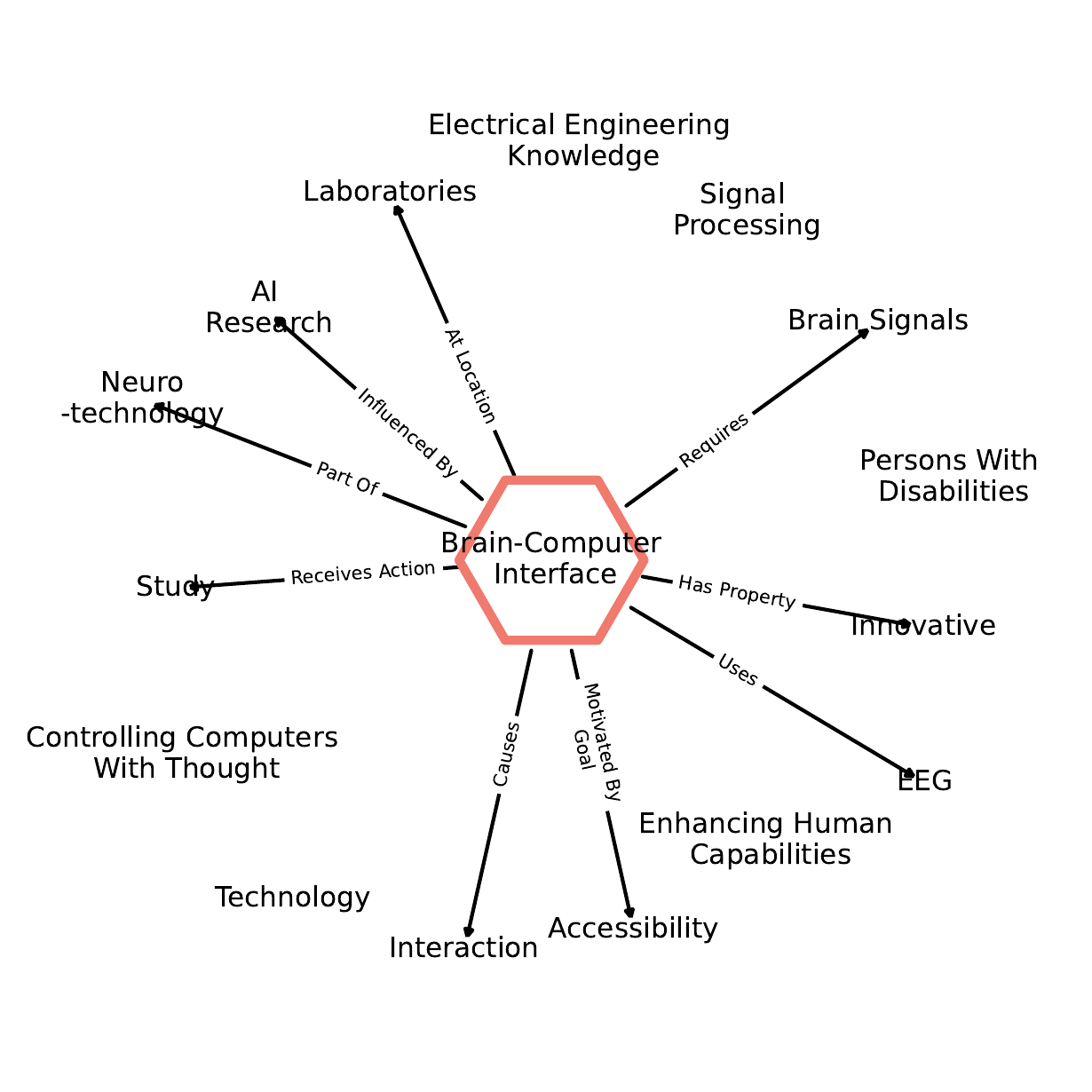}
    }
    \subfloat[After Task 10]{
        \includegraphics[width=0.33\linewidth]{fig/visualization/memorize_5.pdf}
    }

    \caption{The visualization of the \emph{memorized} knowledge related to the ``Brain-Computer Interface'' in IL. The center node represents a concept, while the linked and unlinked edges indicate whether the corresponding \emph{training} samples are answered correctly. }
    \label{fig:concept_memorize_full}
\end{figure*}

\begin{figure*}[htbp]
    \centering
    
    \subfloat[After Task 1]{
        \includegraphics[width=0.33\linewidth]{fig/visualization/generalize_0.pdf}
    }
    \subfloat[After Task 2]{
        \includegraphics[width=0.33\linewidth]{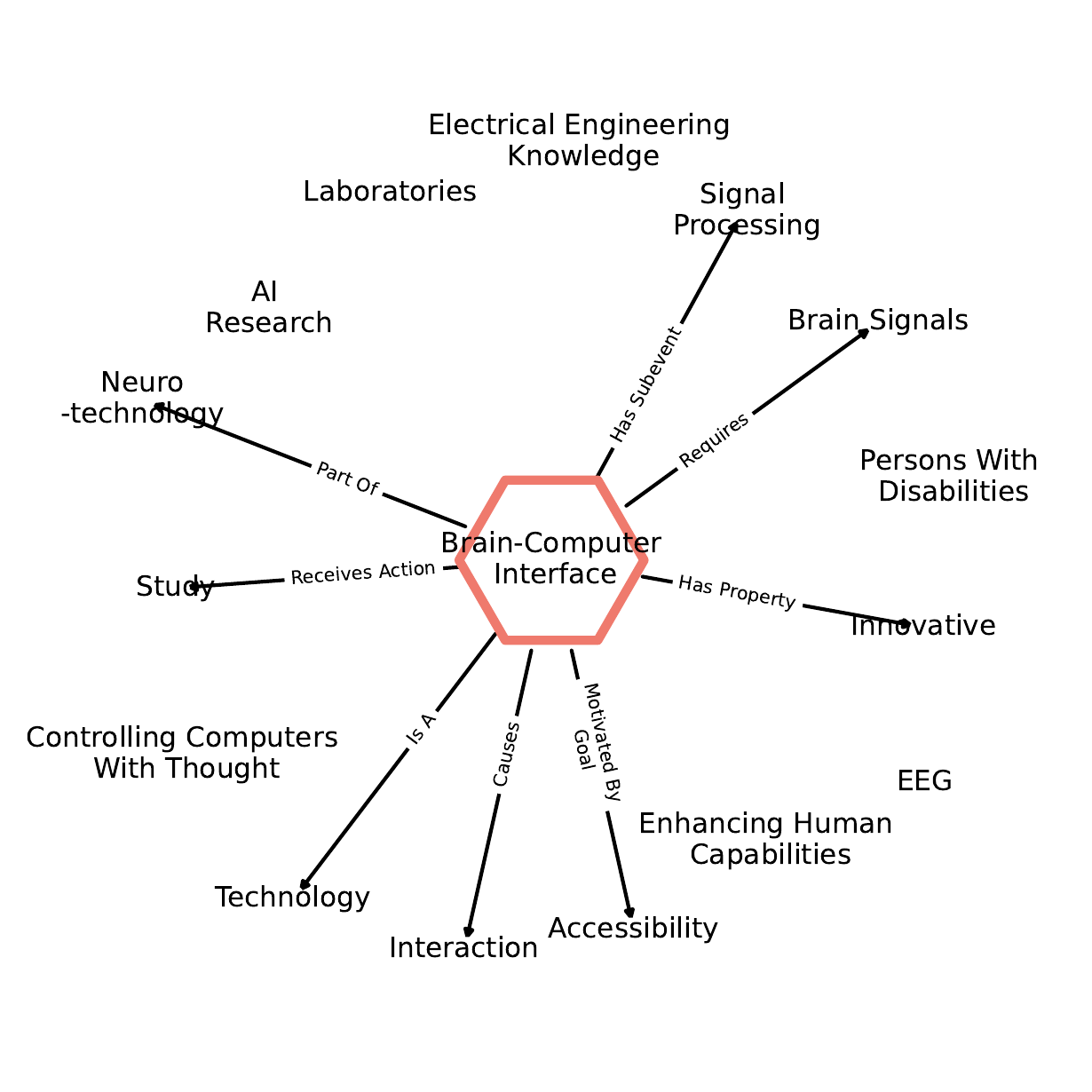}
    }
    \subfloat[After Task 4]{
        \includegraphics[width=0.33\linewidth]{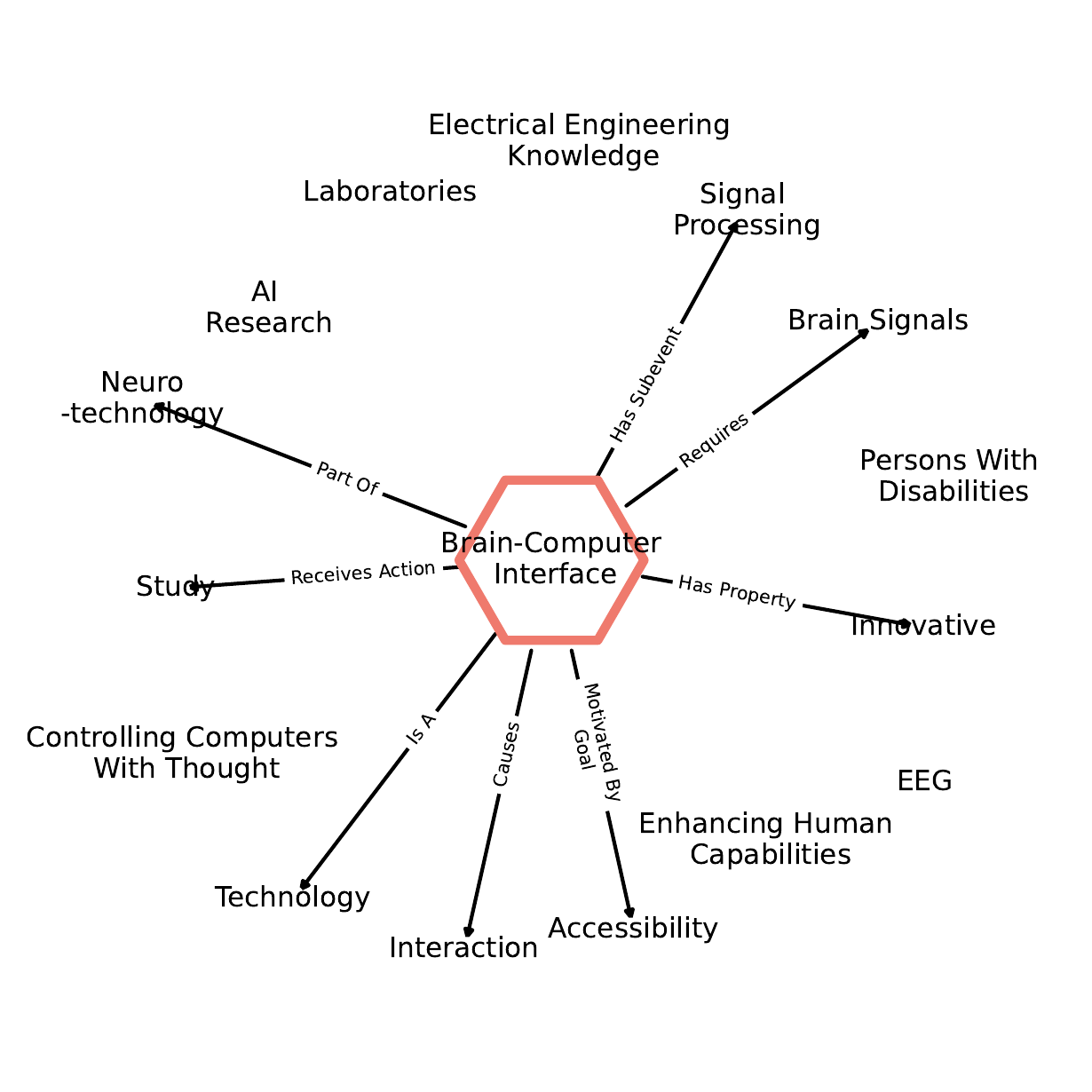}
    }
    
    \subfloat[After Task 6]{
        \includegraphics[width=0.33\linewidth]{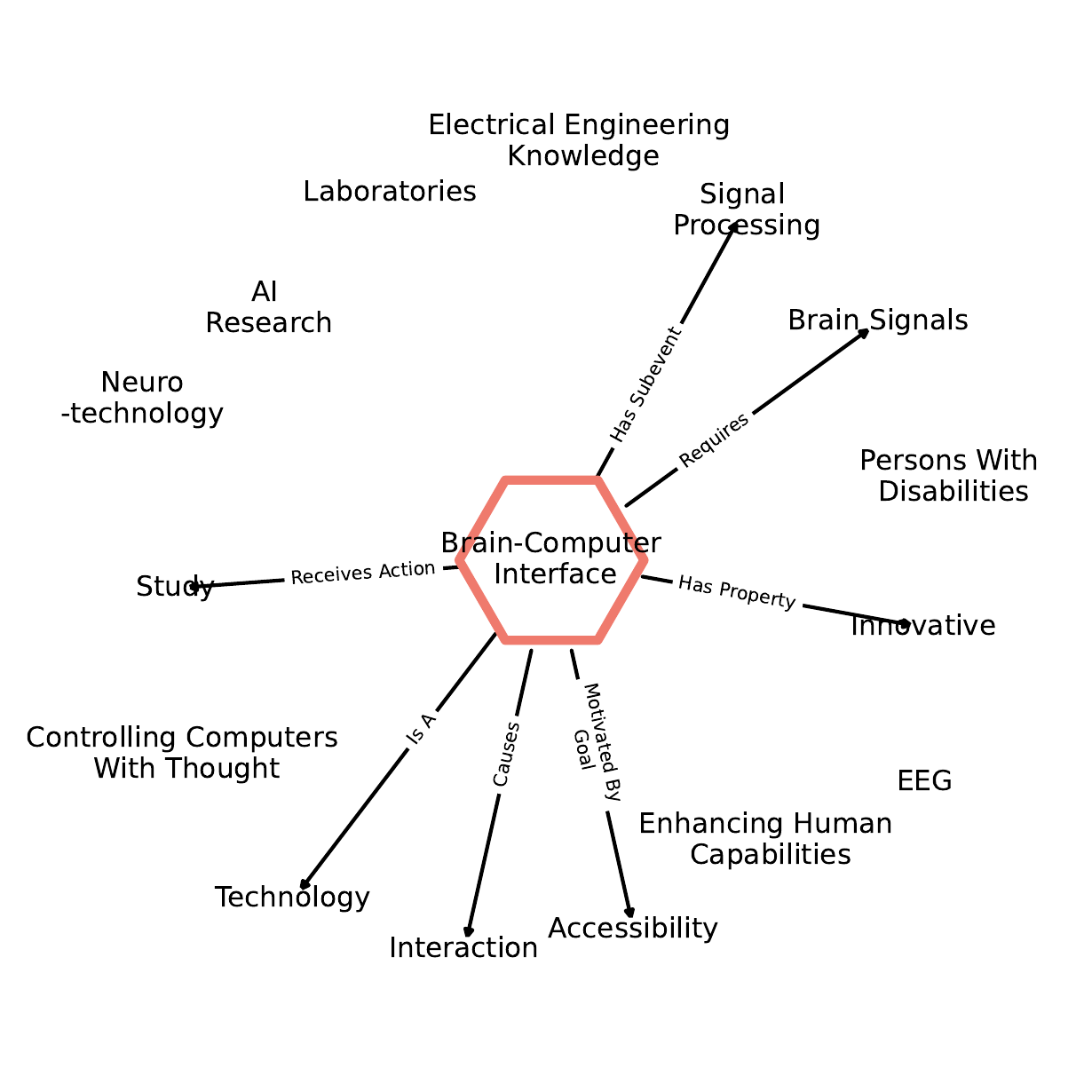}
    }
    \subfloat[After Task 8]{
        \includegraphics[width=0.33\linewidth]{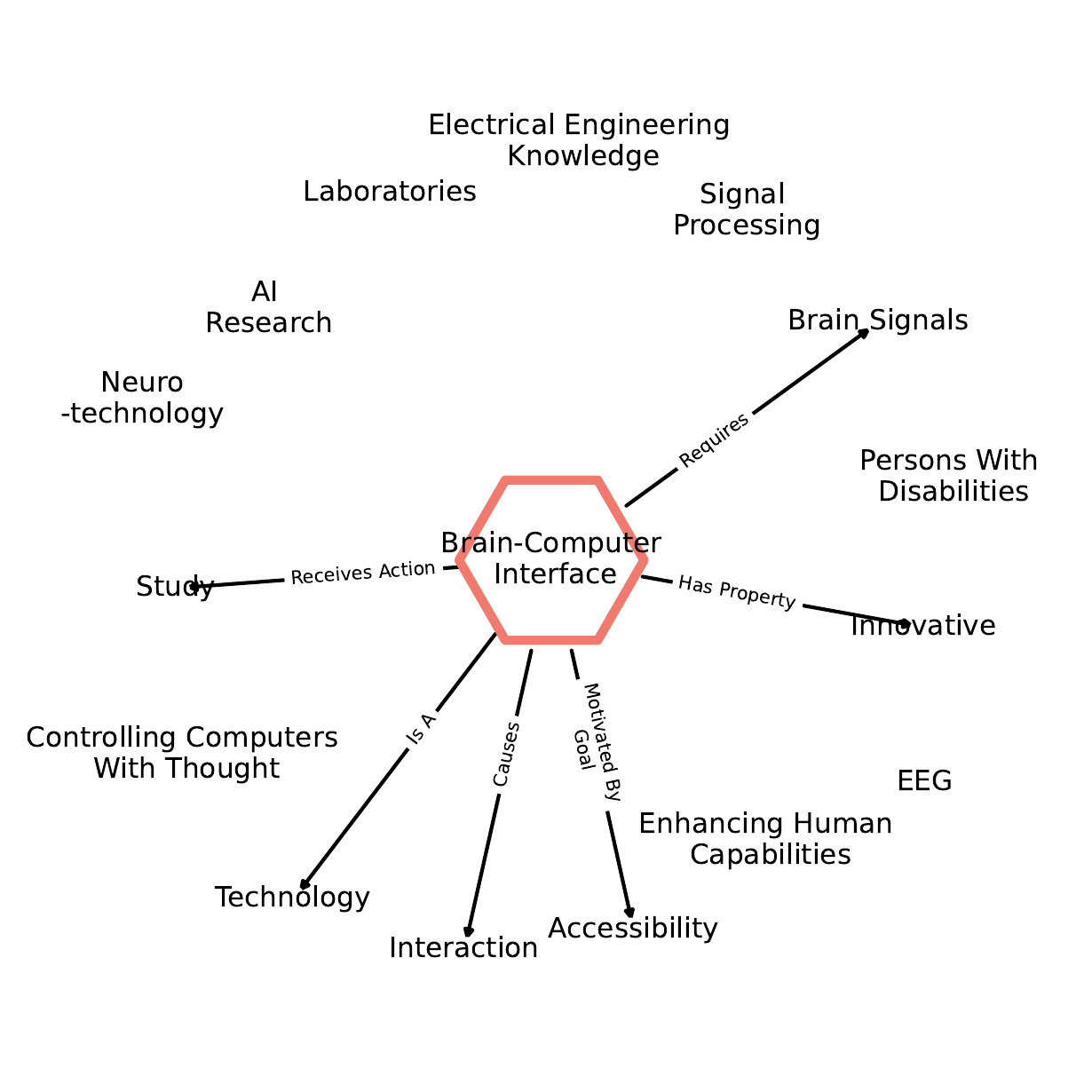}
    }
    \subfloat[After Task 10]{
        \includegraphics[width=0.33\linewidth]{fig/visualization/generalize_5.pdf}
    }
    
    \caption{The visualization of the \emph{generalized} knowledge related to the ``Brain-Computer Interface'' in IL. The center node represents a concept, while the linked and unlinked edges indicate whether the corresponding \emph{test} samples are answered correctly. }
    \label{fig:concept_generalize_full}
\end{figure*}

\section{Visualization of One Concept}
\label{sec:appendix_visualization_one_concept}

We visualize the memorized and generalized knowledge of the concept ``Brain-Computer Interface'' in Figures \ref{fig:concept_memorize_full} and \ref{fig:concept_generalize_full}. In each graph, the center node represents the concept ``Brain-Computer Interface''. The linked and unlinked edges indicate whether the corresponding \emph{training} or \emph{test} samples are answered correctly.

For example, in Figure \ref{fig:concept_memorize_full_task_1}, the edge between ``Brain-Computer Interface'' and ``Signal Processing'' signifies that the LLM correctly outputs the target answer ``Signal Processing'' when the question is the training sample ``What subevent occurs in a Brain-Computer Interface?''. In Figure \ref{fig:concept_memorize_full_task_2}, the edge between ``Brain-Computer Interface'' and ``Signal Processing'' is missing, indicating that the LLM fails to provide the correct target answer.

Figure \ref{fig:concept_memorize_full} shows that even when LLMs can memorize all the knowledge, they tend to first forget more complex knowledge, such as ``(Brain-Computer Interface, UsedFor, Controlling Computers With Thought)''. Conversely, some common knowledge, such as ``(Brain-Computer Interface, Requires, Brain Signals)'', ``(Brain-Computer Interface, Has Property, Innovative)'', and ``(Brain-Computer Interface, Motivated By Goal, Accessibility)'', remains robust and is not forgotten after learning 10 tasks. This indicates that certain knowledge is easier to memorize, generalize, and retain.

Further exploration at the concept-level knowledge in IL is left for future work. We also encourage future studies to utilize the provided Concept-1K dataset for a fine-grained analysis of the memorization and generalization dynamics in IL.

\section{Additional Experimental Results}
\label{sec:appendix_additional_experimental_results}

\begin{table*}[htbp]
  \centering
  \caption{The \textbf{memorization accuracy} when different \textbf{model scales} and \textbf{buffer size} are selected. The pretraining step is 143000 (final version). Figure \ref{fig:buffer_scale_pretraining_exp} (a) summarises this figure's content.}
    \begin{tabular}{lllllll}
        \toprule
          & \multicolumn{1}{c}{\textbf{70M}} & \multicolumn{1}{c}{\textbf{160M}} & \multicolumn{1}{c}{\textbf{410M}} & \multicolumn{1}{c}{\textbf{1B}} & \multicolumn{1}{c}{\textbf{1.4B}} & \multicolumn{1}{c}{\textbf{2.8B}} \\
    \midrule
    \textbf{Buffer Size=0} & 6.34\tiny{±0.86} & 33.53\tiny{±1.39} & 58.28\tiny{±0.61} & 65.97\tiny{±0.31} & 64.95\tiny{±0.55} & 68.03\tiny{±0.46} \\
    \textbf{Buffer Size=2000} & 22.87\tiny{±1.28} & 56.12\tiny{±0.81} & 77.31\tiny{±0.23} & 81.57\tiny{±0.32} & 80.81\tiny{±0.62} & 82.26\tiny{±0.66} \\
    \textbf{Buffer Size=5000} & 42.13\tiny{±1.88} & 73.32\tiny{±1.00} & 90.05\tiny{±0.64} & 91.93\tiny{±0.49} & 91.49\tiny{±0.32} & 92.03\tiny{±0.35} \\
    \textbf{Buffer Size=10000} & 58.50\tiny{±0.50} & 84.91\tiny{±1.64} & 97.81\tiny{±0.12} & 97.21\tiny{±0.73} & 97.35\tiny{±0.84} & 96.74\tiny{±0.43} \\
    \textbf{Buffer Size=All} & 62.37\tiny{±1.82} & 90.04\tiny{±0.64} & 99.01\tiny{±0.14} & 99.10\tiny{±0.57} & 98.86\tiny{±0.55} & 98.72\tiny{±0.34} \\
    \bottomrule
    \end{tabular}%
  \label{tab:scale_buffer_train}%
\end{table*}%

\begin{table*}[htbp]
  \centering
  \caption{The \textbf{generalization accuracy} when different  \textbf{model scales} and \textbf{buffer size} are selected. The pretraining step is 143000 (final version). Figure \ref{fig:buffer_scale_pretraining_exp} (e) summarises this figure's content.}
    \begin{tabular}{lllllll}
        \toprule
          & \multicolumn{1}{c}{\textbf{70M}} & \multicolumn{1}{c}{\textbf{160M}} & \multicolumn{1}{c}{\textbf{410M}} & \multicolumn{1}{c}{\textbf{1B}} & \multicolumn{1}{c}{\textbf{1.4B}} & \multicolumn{1}{c}{\textbf{2.8B}} \\
    \midrule
    \textbf{Buffer Size=0} & 2.52\tiny{±0.12} & 6.57\tiny{±0.28} & 17.69\tiny{±0.35} & 22.65\tiny{±0.39} & 25.57\tiny{±0.36} & 29.56\tiny{±0.35} \\
    \textbf{Buffer Size=2000} & 4.04\tiny{±0.07} & 8.49\tiny{±0.23} & 22.49\tiny{±0.32} & 27.29\tiny{±0.55} & 30.80\tiny{±0.28} & 34.21\tiny{±0.40} \\
    \textbf{Buffer Size=5000} & 5.14\tiny{±0.09} & 10.08\tiny{±0.40} & 25.19\tiny{±0.46} & 29.34\tiny{±0.38} & 33.50\tiny{±0.46} & 37.04\tiny{±0.36} \\
    \textbf{Buffer Size=10000} & 5.88\tiny{±0.10} & 10.36\tiny{±0.52} & 26.20\tiny{±0.35} & 30.81\tiny{±0.24} & 34.85\tiny{±0.57} & 37.72\tiny{±0.52} \\
    \textbf{Buffer Size=All} & 5.80\tiny{±0.08} & 10.80\tiny{±0.46} & 25.70\tiny{±0.28} & 30.99\tiny{±0.42} & 35.03\tiny{±0.27} & 37.97\tiny{±0.37} \\
    \bottomrule
    \end{tabular}%
  \label{tab:scale_buffer_test}%
\end{table*}%

% Table generated by Excel2LaTeX from sheet 'Size-Pretrain-buf0'
\begin{table*}[htbp]
  \centering
  \caption{The \textbf{memorization accuracy} when different \textbf{pretraining steps} and \textbf{model scales} are selected. The buffer size is 0. Figure \ref{fig:buffer_scale_pretraining_exp} (b) summarizes the content of this figure.}
    \begin{tabular}{lllllll}
        \toprule
          & \multicolumn{1}{c}{\textbf{Step=0}} & \multicolumn{1}{c}{\textbf{Step=16}} & \multicolumn{1}{c}{\textbf{Step=128}} & \multicolumn{1}{c}{\textbf{Step=1000}} & \multicolumn{1}{c}{\textbf{Step=10000}} & \multicolumn{1}{c}{\textbf{Step=143000}} \\
    \midrule
    \textbf{Pythia-70M} & 2.92\tiny{±0.06} & 4.89\tiny{±0.16} & 4.58\tiny{±0.10} & 24.69\tiny{±0.04} & 36.08\tiny{±0.63} & 6.34\tiny{±0.86} \\
    \textbf{Pythia-160M} & 26.33\tiny{±0.31} & 29.34\tiny{±0.25} & 29.11\tiny{±0.11} & 48.78\tiny{±0.09} & 60.12\tiny{±0.34} & 33.53\tiny{±1.39} \\
    \textbf{Pythia-410M} & 29.58\tiny{±1.10} & 31.06\tiny{±0.24} & 30.78\tiny{±0.22} & 49.84\tiny{±0.06} & 64.31\tiny{±0.38} & 58.28\tiny{±0.61} \\
    \textbf{Pythia-1B} & 33.18\tiny{±0.59} & 33.71\tiny{±0.17} & 33.91\tiny{±0.49} & 53.14\tiny{±0.59} & 66.66\tiny{±0.59} & 65.97\tiny{±0.31} \\
    \textbf{Pythia-1.4B} & 35.61\tiny{±0.59} & 36.31\tiny{±0.51} & 36.15\tiny{±0.32} & 55.61\tiny{±0.59} & 70.66\tiny{±0.51} & 64.95\tiny{±0.55} \\
    \textbf{Pythia-2.8B} & 37.76\tiny{±0.29} & 37.73\tiny{±0.38} & 37.65\tiny{±0.39} & 57.09\tiny{±0.34} & 72.92\tiny{±0.59} & 68.03\tiny{±0.46} \\
    \bottomrule
    \end{tabular}%
  \label{tab:pretrain_scale_buf0_train}%
\end{table*}%

\begin{table*}[htbp]
  \centering
  \caption{The \textbf{generalization accuracy} when different \textbf{pretraining steps} and \textbf{model scales} are selected. The buffer size is 0. Figure \ref{fig:buffer_scale_pretraining_exp} (f) summarises this figure's content.}
    \begin{tabular}{lllllll}
        \toprule
          & \multicolumn{1}{c}{\textbf{Step=0}} & \multicolumn{1}{c}{\textbf{Step=16}} & \multicolumn{1}{c}{\textbf{Step=128}} & \multicolumn{1}{c}{\textbf{Step=1000}} & \multicolumn{1}{c}{\textbf{Step=10000}} & \multicolumn{1}{c}{\textbf{Step=143000}} \\
    \midrule
    \textbf{Pythia-70M} & 0.83\tiny{±0.02} & 1.10\tiny{±0.01} & 1.08\tiny{±0.04} & 2.94\tiny{±0.04} & 3.95\tiny{±0.03} & 2.52\tiny{±0.12} \\
    \textbf{Pythia-160M} & 1.64\tiny{±0.04} & 1.81\tiny{±0.02} & 1.82\tiny{±0.06} & 3.79\tiny{±0.05} & 6.96\tiny{±0.04} & 6.57\tiny{±0.28} \\
    \textbf{Pythia-410M} & 2.34\tiny{±0.07} & 2.35\tiny{±0.04} & 2.21\tiny{±0.04} & 4.18\tiny{±0.10} & 12.91\tiny{±0.14} & 17.69\tiny{±0.35} \\
    \textbf{Pythia-1B} & 2.78\tiny{±0.14} & 2.85\tiny{±0.24} & 2.80\tiny{±0.24} & 6.03\tiny{±0.24} & 17.70\tiny{±0.24} & 22.65\tiny{±0.39} \\
    \textbf{Pythia-1.4B} & 2.71\tiny{±0.37} & 2.32\tiny{±0.24} & 2.40\tiny{±0.53} & 5.68\tiny{±0.24} & 20.64\tiny{±0.24} & 25.57\tiny{±0.36} \\
    \textbf{Pythia-2.8B} & 2.48\tiny{±0.23} & 1.79\tiny{±0.29} & 2.09\tiny{±0.32} & 5.35\tiny{±0.49} & 24.79\tiny{±0.24} & 29.56\tiny{±0.35} \\
    \bottomrule
    \end{tabular}%
  \label{tab:pretrain_scale_buf0_test}%
\end{table*}%

% Table generated by Excel2LaTeX from sheet 'Size-Pretrain-buf2000'
\begin{table*}[htbp]
  \centering
  \caption{The \textbf{memorization accuracy} when different \textbf{pretraining steps} and \textbf{model scales} are selected. The buffer size is 2000. Figure \ref{fig:buffer_scale_pretraining_exp} (c) summarises this figure's content.}
    \begin{tabular}{lllllll}
    \toprule
          & \multicolumn{1}{c}{\textbf{Step=0}} & \multicolumn{1}{c}{\textbf{Step=16}} & \multicolumn{1}{c}{\textbf{Step=128}} & \multicolumn{1}{c}{\textbf{Step=1000}} & \multicolumn{1}{c}{\textbf{Step=10000}} & \multicolumn{1}{c}{\textbf{Step=143000}} \\
    \midrule
    \textbf{Pythia-70M} & 22.15\tiny{±0.13} & 24.91\tiny{±0.20} & 23.54\tiny{±0.15} & 55.06\tiny{±0.16} & 64.89\tiny{±0.23} & 22.87\tiny{±1.28} \\
    \textbf{Pythia-160M} & 59.42\tiny{±0.13} & 59.62\tiny{±0.77} & 59.30\tiny{±0.18} & 78.84\tiny{±0.44} & 82.30\tiny{±0.18} & 56.12\tiny{±0.81} \\
    \textbf{Pythia-410M} & 60.43\tiny{±0.59} & 61.09\tiny{±0.66} & 61.24\tiny{±0.27} & 79.77\tiny{±0.15} & 83.24\tiny{±0.12} & 77.31\tiny{±0.23} \\
    \textbf{Pythia-1B} & 63.11\tiny{±0.15} & 62.05\tiny{±0.40} & 63.51\tiny{±0.31} & 80.11\tiny{±0.50} & 83.11\tiny{±0.38} & 81.57\tiny{±0.32} \\
    \textbf{Pythia-1.4B} & 63.77\tiny{±0.15} & 64.14\tiny{±0.23} & 65.40\tiny{±0.32} & 82.38\tiny{±0.34} & 84.19\tiny{±0.22} & 80.81\tiny{±0.62} \\
    \textbf{Pythia-2.8B} & 65.10\tiny{±0.45} & 65.84\tiny{±0.57} & 66.04\tiny{±0.53} & 82.63\tiny{±0.18} & 84.80\tiny{±0.44} & 82.26\tiny{±0.66} \\
    \bottomrule
    \end{tabular}%
  \label{tab:pretrain_scale_buf2000_train}%
\end{table*}%

\begin{table*}[htbp]
  \centering
  \caption{The \textbf{generalization accuracy} when different \textbf{pretraining steps} and \textbf{model scales} are selected. The buffer size is 2000. Figure \ref{fig:buffer_scale_pretraining_exp} (g) summarises this figure's content.}
    \begin{tabular}{lllllll}
    \toprule
          & \multicolumn{1}{c}{\textbf{Step=0}} & \multicolumn{1}{c}{\textbf{Step=16}} & \multicolumn{1}{c}{\textbf{Step=128}} & \multicolumn{1}{c}{\textbf{Step=1000}} & \multicolumn{1}{c}{\textbf{Step=10000}} & \multicolumn{1}{c}{\textbf{Step=143000}} \\
    \midrule
    \textbf{Pythia-70M} & 2.07\tiny{±0.08} & 2.46\tiny{±0.06} & 2.51\tiny{±0.13} & 3.84\tiny{±0.06} & 4.57\tiny{±0.07} & 4.04\tiny{±0.07} \\
    \textbf{Pythia-160M} & 3.17\tiny{±0.08} & 3.55\tiny{±0.14} & 3.52\tiny{±0.06} & 5.26\tiny{±0.14} & 8.05\tiny{±0.08} & 8.49\tiny{±0.23} \\
    \textbf{Pythia-410M} & 3.96\tiny{±0.06} & 4.97\tiny{±0.09} & 5.09\tiny{±0.04} & 6.25\tiny{±0.13} & 15.64\tiny{±0.06} & 22.49\tiny{±0.32} \\
    \textbf{Pythia-1B} & 5.51\tiny{±0.17} & 5.81\tiny{±0.43} & 6.37\tiny{±0.56} & 8.85\tiny{±0.55} & 20.01\tiny{±0.51} & 27.29\tiny{±0.55} \\
    \textbf{Pythia-1.4B} & 6.76\tiny{±0.21} & 6.98\tiny{±0.16} & 7.55\tiny{±0.55} & 10.21\tiny{±0.42} & 23.67\tiny{±0.18} & 30.80\tiny{±0.28} \\
    \textbf{Pythia-2.8B} & 8.07\tiny{±0.28} & 8.57\tiny{±0.54} & 9.32\tiny{±0.40} & 12.61\tiny{±0.25} & 27.50\tiny{±0.23} & 34.21\tiny{±0.40} \\
    \bottomrule
    \end{tabular}%
  \label{tab:pretrain_scale_buf2000_test}%
\end{table*}%

% Table generated by Excel2LaTeX from sheet 'Size-Pretrain-buf20000'
\begin{table*}[htbp]
  \centering
  \caption{The \textbf{memorization accuracy} when different \textbf{pretraining steps} and \textbf{model scales} are selected. The buffer size is unlimited (all old samples are stored). Figure \ref{fig:buffer_scale_pretraining_exp} (d) summarizes the content of this figure.}
    \begin{tabular}{lllllll}
    \toprule
          & \multicolumn{1}{c}{\textbf{Step=0}} & \multicolumn{1}{c}{\textbf{Step=16}} & \multicolumn{1}{c}{\textbf{Step=128}} & \multicolumn{1}{c}{\textbf{Step=1000}} & \multicolumn{1}{c}{\textbf{Step=10000}} & \multicolumn{1}{c}{\textbf{Step=143000}} \\
    \midrule
    \textbf{Pythia-70M} & 65.77\tiny{±0.56} & 67.75\tiny{±0.19} & 66.36\tiny{±0.09} & 87.61\tiny{±0.14} & 92.59\tiny{±0.11} & 62.37\tiny{±1.82} \\
    \textbf{Pythia-160M} & 86.54\tiny{±0.15} & 88.75\tiny{±0.28} & 88.64\tiny{±0.07} & 98.95\tiny{±0.03} & 99.62\tiny{±0.05} & 90.04\tiny{±0.64} \\
    \textbf{Pythia-410M} & 87.46\tiny{±0.19} & 89.07\tiny{±0.27} & 90.03\tiny{±0.16} & 99.30\tiny{±0.22} & 99.59\tiny{±0.02} & 99.01\tiny{±0.14} \\
    \textbf{Pythia-1B} & 88.77\tiny{±0.02} & 89.75\tiny{±0.19} & 90.36\tiny{±0.09} & 99.75\tiny{±0.19} & 99.36\tiny{±0.09} & 99.10\tiny{±0.57} \\
    \textbf{Pythia-1.4B} & 89.65\tiny{±0.02} & 90.34\tiny{±0.03} & 90.76\tiny{±0.54} & 99.85\tiny{±0.21} & 99.76\tiny{±0.02} & 98.86\tiny{±0.55} \\
    \textbf{Pythia-2.8B} & 90.31\tiny{±0.36} & 90.89\tiny{±0.53} & 91.43\tiny{±0.44} & 99.95\tiny{±0.56} & 99.86\tiny{±0.02} & 98.72\tiny{±0.34} \\
    \bottomrule
    \end{tabular}%
  \label{tab:pretrain_scale_buf20000_train}%
\end{table*}%

\begin{table*}[htbp]
  \centering
  \caption{The \textbf{generalization accuracy} when different \textbf{pretraining steps} and \textbf{model scales} are selected. The buffer size is unlimited (all old samples are stored). Figure \ref{fig:buffer_scale_pretraining_exp} (h) summarises this figure's content.}
    \begin{tabular}{lllllll}
    \toprule
          & \multicolumn{1}{c}{\textbf{Step=0}} & \multicolumn{1}{c}{\textbf{Step=16}} & \multicolumn{1}{c}{\textbf{Step=128}} & \multicolumn{1}{c}{\textbf{Step=1000}} & \multicolumn{1}{c}{\textbf{Step=10000}} & \multicolumn{1}{c}{\textbf{Step=143000}} \\
    \midrule
    \textbf{Pythia-70M} & 3.18\tiny{±0.12} & 3.10\tiny{±0.02} & 3.08\tiny{±0.04} & 4.37\tiny{±0.06} & 5.40\tiny{±0.03} & 5.80\tiny{±0.08} \\
    \textbf{Pythia-160M} & 3.79\tiny{±0.25} & 3.95\tiny{±0.10} & 3.93\tiny{±0.04} & 6.09\tiny{±0.06} & 9.08\tiny{±0.06} & 10.80\tiny{±0.46} \\
    \textbf{Pythia-410M} & 4.48\tiny{±0.03} & 5.12\tiny{±0.18} & 5.24\tiny{±0.16} & 7.29\tiny{±0.16} & 18.14\tiny{±0.11} & 25.70\tiny{±0.28} \\
    \textbf{Pythia-1B} & 6.12\tiny{±0.56} & 6.12\tiny{±0.41} & 6.75\tiny{±0.20} & 9.74\tiny{±0.26} & 25.07\tiny{±0.38} & 30.99\tiny{±0.42} \\
    \textbf{Pythia-1.4B} & 6.96\tiny{±0.36} & 6.87\tiny{±0.55} & 7.45\tiny{±0.50} & 11.07\tiny{±0.40} & 27.93\tiny{±0.56} & 35.03\tiny{±0.27} \\
    \textbf{Pythia-2.8B} & 8.47\tiny{±0.20} & 8.81\tiny{±0.31} & 9.82\tiny{±0.45} & 14.73\tiny{±0.15} & 30.65\tiny{±0.21} & 37.97\tiny{±0.37} \\
    \bottomrule
    \end{tabular}%
  \label{tab:pretrain_scale_buf20000_test}%
\end{table*}%

\begin{figure*}[htbp]
    \centering
    \subfloat[Average Accuracy on Training Set]{
        \includegraphics[width=0.49\linewidth]{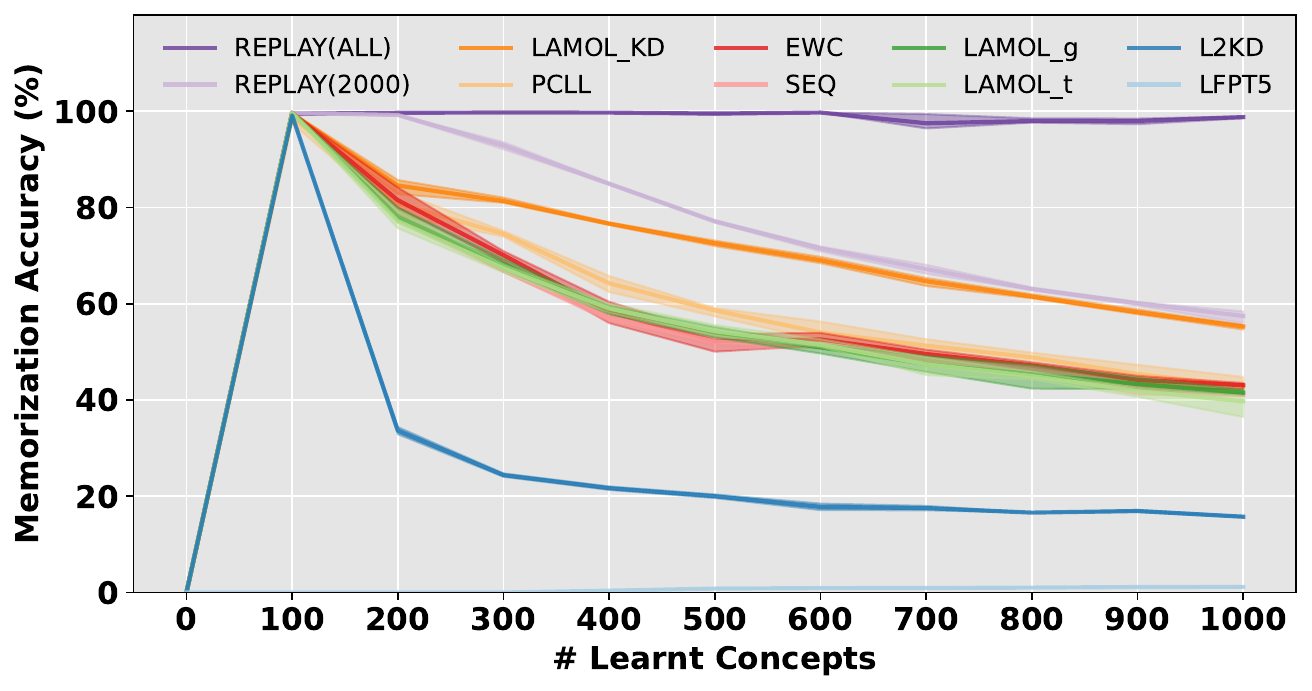}
    }
    \subfloat[Average Accuracy on Test Set]{
        \includegraphics[width=0.49\linewidth]{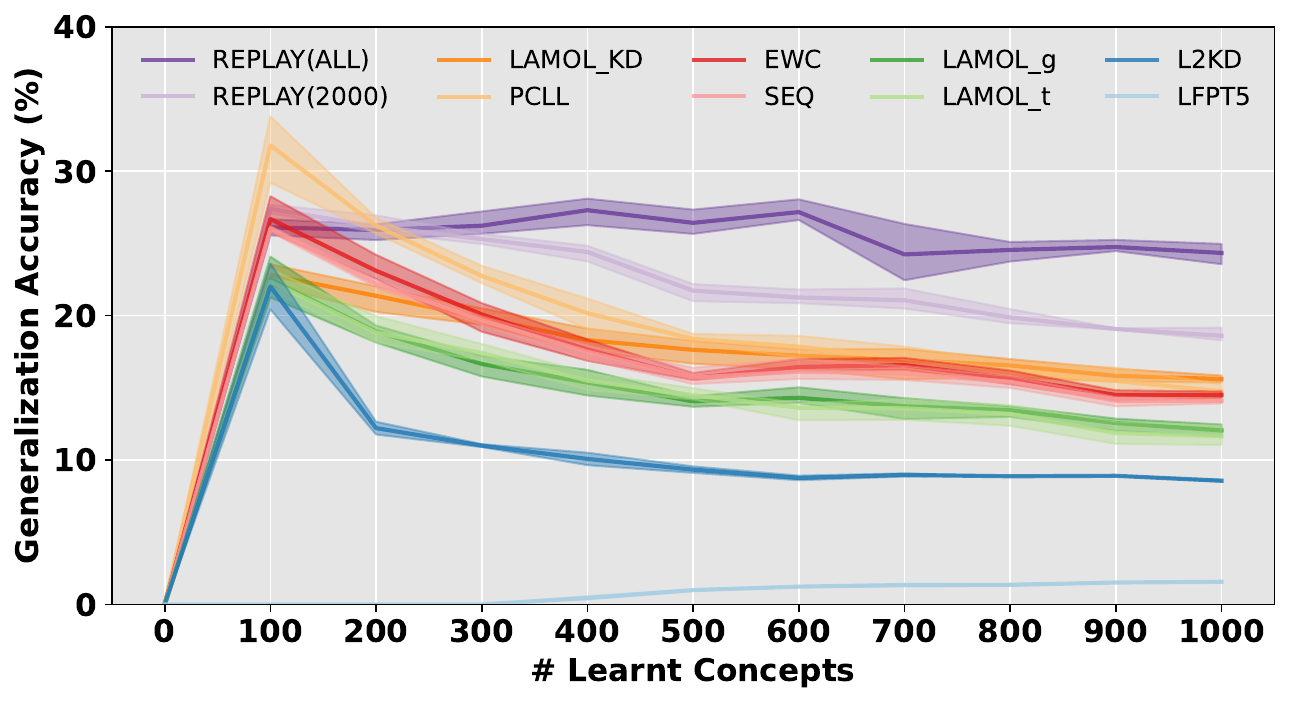}
    }
    
    \subfloat[Memorization Accuracy]{
        \includegraphics[width=0.49\linewidth]{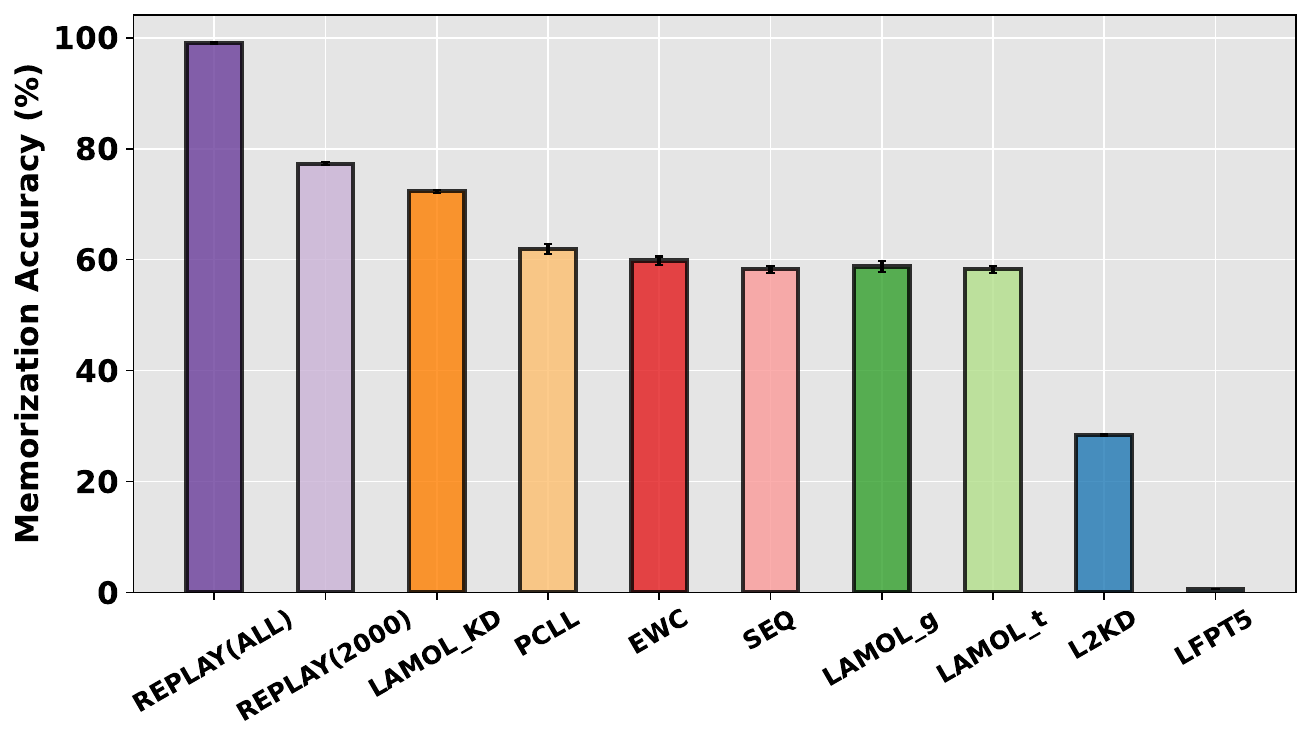}
    }
    \subfloat[Generalization Accuracy]{
        \includegraphics[width=0.49\linewidth]{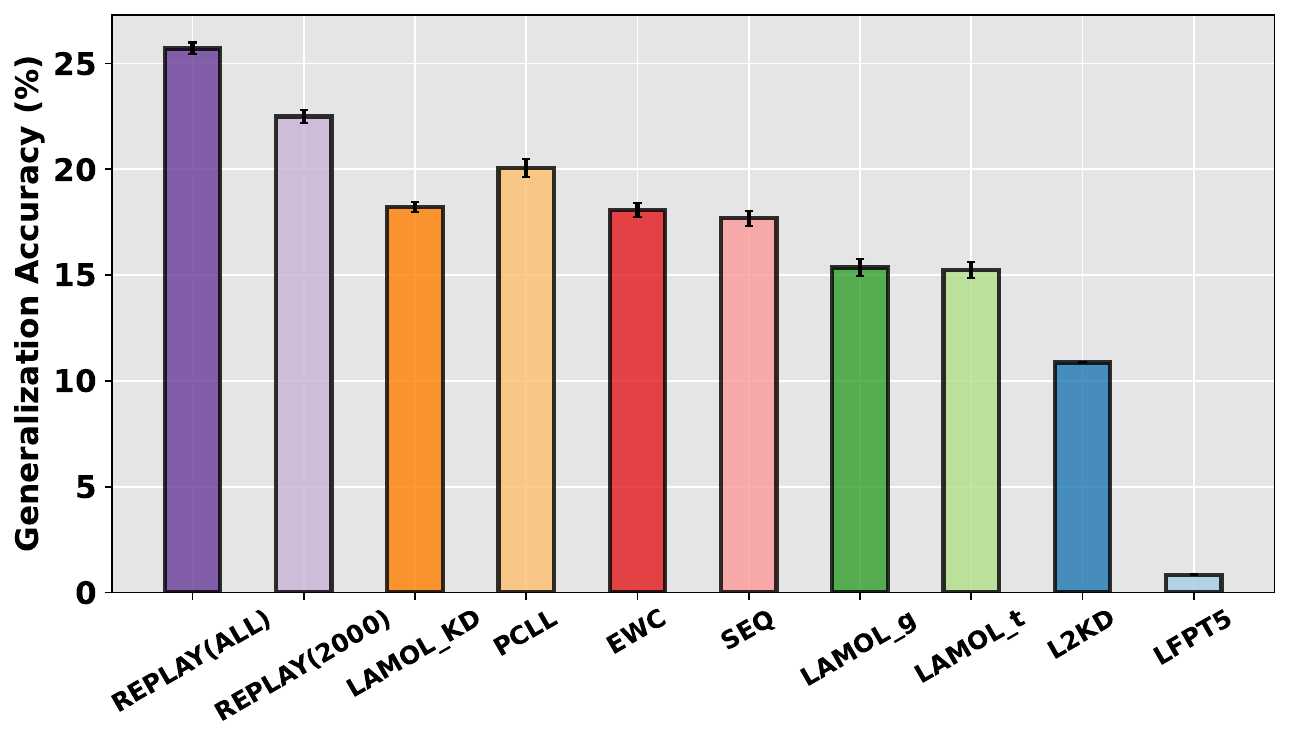}
    }

    \subfloat[Memorization Forgetting]{
        \includegraphics[width=0.49\linewidth]{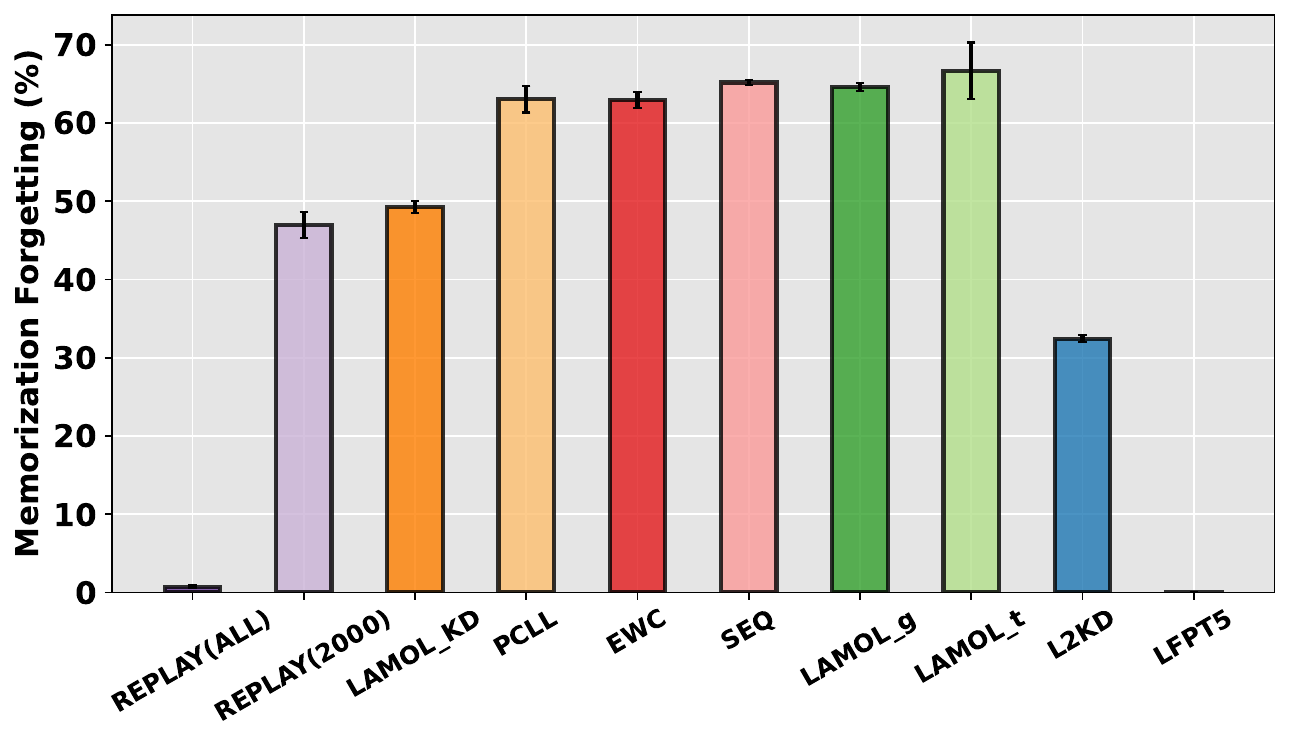}
    }
    \subfloat[Generalization Forgetting]{
        \includegraphics[width=0.49\linewidth]{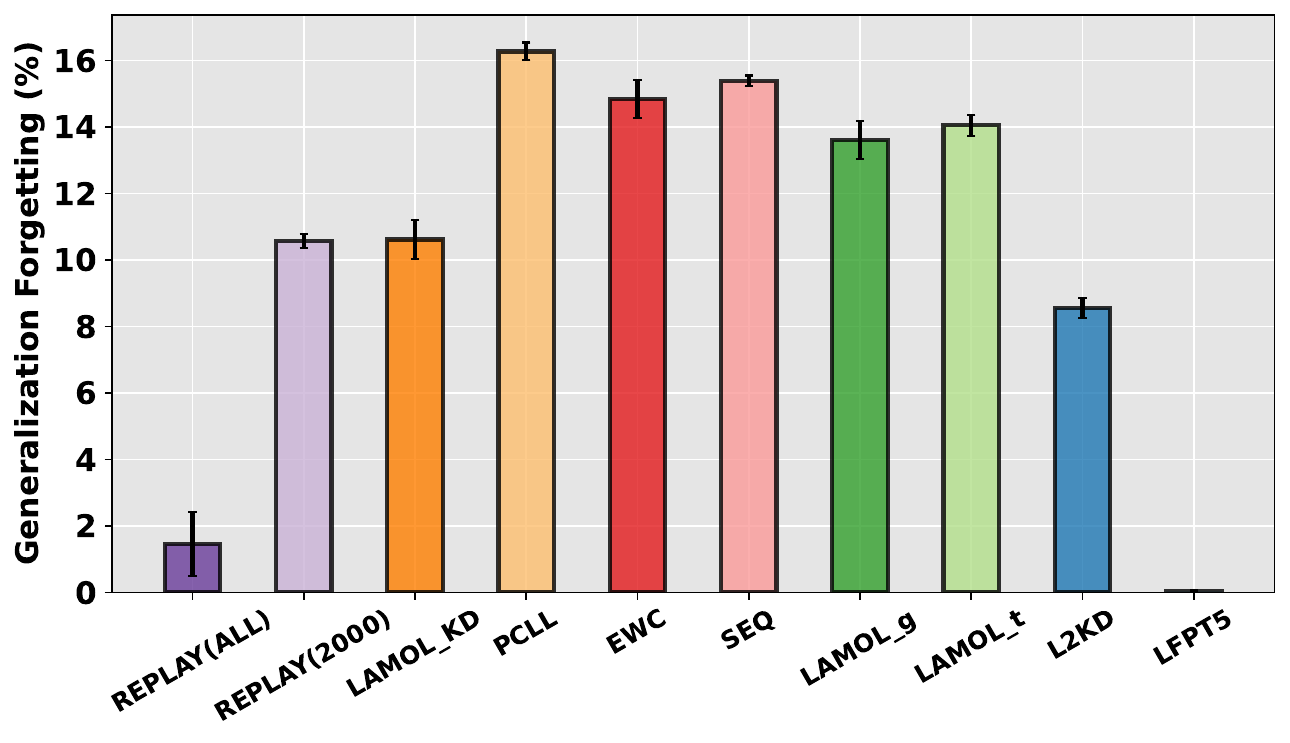}
    }

    \subfloat[Training Loss]{
        \includegraphics[width=0.49\linewidth]{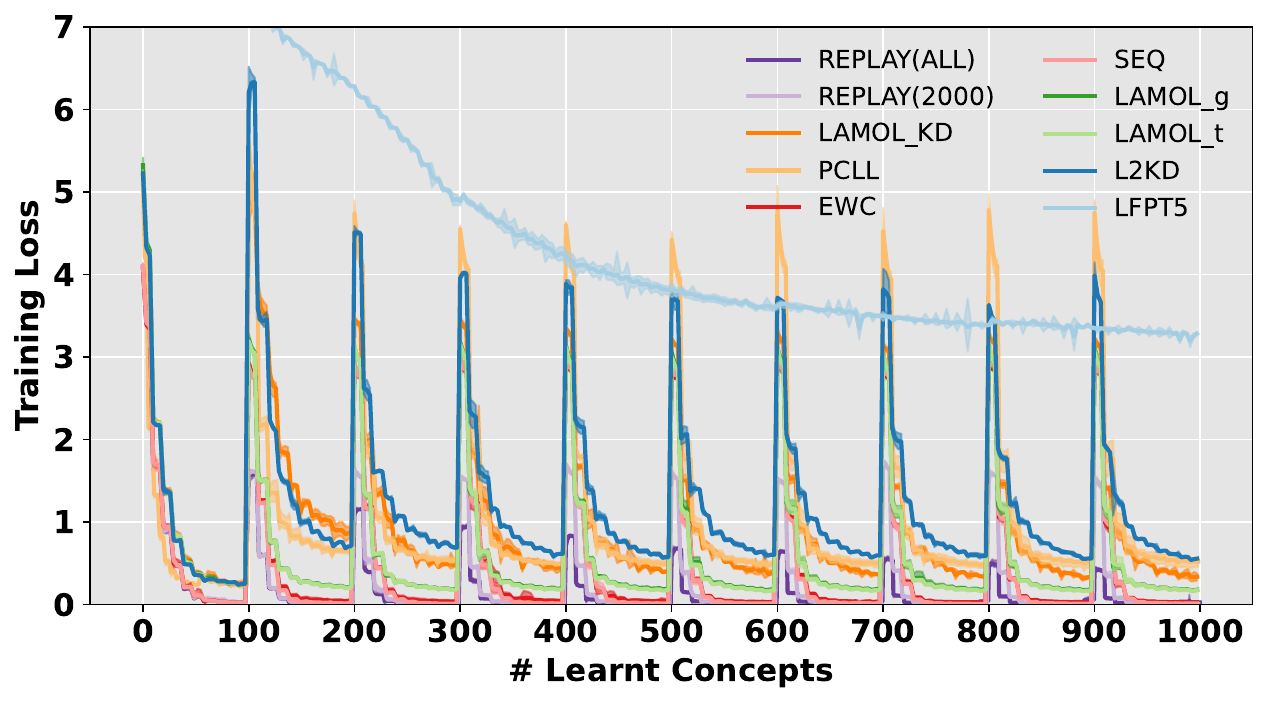}
    }
    \caption{The detailed result of SOTA methods on Concept-1K.}
    \label{fig:sota_detailed}
\end{figure*}

\begin{table*}[htbp]
  \centering
  \caption{Examples of Concept-1K. Each triplet corresponds to a training instance and a test instance.}
  \resizebox{0.97\linewidth}{!}{
    \begin{tabular}{cllll}
    \toprule
    \multicolumn{1}{l}{\textbf{Domain}} & \textbf{Concept} & \textbf{Triplet} & \textbf{Training and Test Input} & \textbf{Target Output} \\
    \midrule
    \multirow{20}[20]{*}{\textbf{Science and Technology}} & \multirow{10}[10]{*}{\textbf{Brain-Computer Interface}} & \multicolumn{1}{l}{\multirow{2}[2]{*}{\textbf{(Brain-Computer Interface, UsedFor, ControllingComputersWithThought)}}} & What is the main use of a Brain-Computer Interface? & \multirow{2}[2]{*}{controlling computers with thought} \\
          &       &       & How can individuals utilize a Brain-Computer Interface? &  \\
\cmidrule{3-5}          &       & \multirow{2}[2]{*}{\textbf{(Brain-Computer Interface, Uses, EEG)}} & What does a Brain-Computer Interface use for its functioning? & \multirow{2}[2]{*}{EEG} \\
          &       &       & What technology is utilized by a Brain-Computer Interface? &  \\
\cmidrule{3-5}          &       & \multirow{2}[2]{*}{\textbf{(Brain-Computer Interface, DesignedFor, PersonsWithDisabilities)}} & Who is the Brain-Computer Interface designed for? & \multirow{2}[2]{*}{persons with disabilities} \\
          &       &       & What group benefits from the Brain-Computer Interface's design? &  \\
\cmidrule{3-5}          &       & \multirow{2}[2]{*}{\textbf{(Brain-Computer Interface, PartOf, Neurotechnology)}} & What field is Brain-Computer Interface a part of? & \multirow{2}[2]{*}{neurotechnology} \\
          &       &       & To which broader technology does Brain-Computer Interface belong? &  \\
\cmidrule{3-5}          &       & \multirow{2}[2]{*}{\textbf{(Brain-Computer Interface, Requires, BrainSignals)}} & What does a Brain-Computer Interface require to function? & \multirow{2}[2]{*}{brain signals} \\
          &       &       & What is necessary for the operation of a Brain-Computer Interface? &  \\
\cmidrule{2-5}          & \multirow{10}[10]{*}{\textbf{IPTV}} & \multicolumn{1}{l}{\multirow{2}[2]{*}{\textbf{(IPTV, IsA, TelevisionService)}}} & What type of service is IPTV classified as? & \multirow{2}[2]{*}{television service} \\
          &       &       & How is IPTV categorized in terms of service? &  \\
\cmidrule{3-5}          &       & \multirow{2}[2]{*}{\textbf{(IPTV, CompetesWith, CableTelevision)}} & Who does IPTV compete with? & \multirow{2}[2]{*}{cable television} \\
          &       &       & What is IPTV's competition in the television market? &  \\
\cmidrule{3-5}          &       & \multirow{2}[2]{*}{\textbf{(IPTV, Supports, HighDefinition)}} & What type of video quality is supported by IPTV? & \multirow{2}[2]{*}{high definition} \\
          &       &       & What does IPTV support? &  \\
\cmidrule{3-5}          &       & \multirow{2}[2]{*}{\textbf{(IPTV, Allows, TimeShifting)}} & What does IPTV allow? & \multirow{2}[2]{*}{time shifting} \\
          &       &       & What feature does IPTV provide concerning watching schedules? &  \\
\cmidrule{3-5}          &       & \multirow{2}[2]{*}{\textbf{(IPTV, Needs, SetTopBox)}} & What does IPTV need to operate? & \multirow{2}[2]{*}{set top box} \\
          &       &       & What equipment is necessary for IPTV functionality? &  \\
    \midrule
    \multirow{20}[20]{*}{\textbf{Education}} & \multirow{10}[10]{*}{\textbf{MOOCs}} & \multicolumn{1}{l}{\multirow{2}[2]{*}{\textbf{(MOOCs, PartOf, OnlineLearning)}}} & What is MOOCs a part of? & \multirow{2}[2]{*}{online learning} \\
          &       &       & Under what broader learning category do MOOCs fall? &  \\
\cmidrule{3-5}          &       & \multirow{2}[2]{*}{\textbf{(MOOCs, AffiliatedWith, Universities)}} & With whom are MOOCs affiliated? & \multirow{2}[2]{*}{universities} \\
          &       &       & What types of institutions are MOOCs commonly associated with? &  \\
\cmidrule{3-5}          &       & \multirow{2}[2]{*}{\textbf{(MOOCs, Offers, FreeCourses)}} & What do MOOCs offer to learners? & \multirow{2}[2]{*}{free courses} \\
          &       &       & What can learners benefit from MOOCs at no cost? &  \\
\cmidrule{3-5}          &       & \multirow{2}[2]{*}{\textbf{(MOOCs, AimedAt, SelfLearners)}} & Who are MOOCs primarily aimed at? & \multirow{2}[2]{*}{self learners} \\
          &       &       & What target audience do MOOCs cater to? &  \\
\cmidrule{3-5}          &       & \multirow{2}[2]{*}{\textbf{(MOOCs, Employs, AutomatedGrading)}} & What grading method do MOOCs employ? & \multirow{2}[2]{*}{automated grading} \\
          &       &       & How are assignments typically graded in MOOCs? &  \\
\cmidrule{2-5}          & \multirow{10}[10]{*}{\textbf{STEM Education}} & \multicolumn{1}{l}{\multirow{2}[2]{*}{\textbf{(STEM Education, FocusOn, ScienceTechnologyEngineeringMathematics)}}} & What does STEM Education focus on? & \multirow{2}[2]{*}{science technology engineering mathematics} \\
          &       &       & What are the main areas of study in STEM Education? &  \\
\cmidrule{3-5}          &       & \multirow{2}[2]{*}{\textbf{(STEM Education, Encourages, CriticalThinking)}} & What does STEM Education encourage in students? & \multirow{2}[2]{*}{critical thinking} \\
          &       &       & What skill is STEM Education known to foster among learners? &  \\
\cmidrule{3-5}          &       & \multirow{2}[2]{*}{\textbf{(STEM Education, BenefitsFrom, InterdisciplinaryApproach)}} & What does STEM Education benefit from? & \multirow{2}[2]{*}{interdisciplinary approach} \\
          &       &       & Which approach enhances the effectiveness of STEM Education? &  \\
\cmidrule{3-5}          &       & \multirow{2}[2]{*}{\textbf{(STEM Education, Enhances, TechnologicalLiteracy)}} & What does STEM Education enhance among students? & \multirow{2}[2]{*}{technological literacy} \\
          &       &       & What aspect of literacy is improved through STEM Education? &  \\
\cmidrule{3-5}          &       & \multirow{2}[2]{*}{\textbf{(STEM Education, SupportedBy, GovernmentPolicies)}} & Who supports STEM Education through policies? & \multirow{2}[2]{*}{government policies} \\
          &       &       & What supports the implementation of STEM Education? &  \\
    \midrule
    \multirow{20}[20]{*}{\textbf{Economy}} & \multirow{10}[10]{*}{} & \multicolumn{1}{l}{\multirow{2}[2]{*}{\textbf{(Ethereum 2.0, Offers, Scalability)}}} & What does Ethereum 2.0 offer to its users? & \multirow{2}[2]{*}{scalability} \\
          &       &       & What is a key feature of Ethereum 2.0? &  \\
\cmidrule{3-5}          &       & \multirow{2}[2]{*}{\textbf{(Ethereum 2.0, Implements, ProofOfStake)}} & What consensus mechanism does Ethereum 2.0 implement? & \multirow{2}[2]{*}{proof of stake} \\
          &       &       & Which algorithm is used by Ethereum 2.0 for transaction validation? &  \\
\cmidrule{3-5}          &       & \multirow{2}[2]{*}{\textbf{(Ethereum 2.0, AimsTo, ReduceEnergyConsumption)}} & What is Ethereum 2.0's goal regarding energy use? & \multirow{2}[2]{*}{reduce energy consumption} \\
          &       &       & How does Ethereum 2.0 plan to impact the environment? &  \\
\cmidrule{3-5}          &       & \multirow{2}[2]{*}{\textbf{(Ethereum 2.0, Uses, BeaconChain)}} & What does Ethereum 2.0 use to manage its consensus process? & \multirow{2}[2]{*}{beacon chain} \\
          &       &       & What is a key component of Ethereum 2.0’s infrastructure? &  \\
\cmidrule{3-5}          &       & \multirow{2}[2]{*}{\textbf{(Ethereum 2.0, PlansFor, eWASM)}} & What does Ethereum 2.0 plan for? & \multirow{2}[2]{*}{eWASM} \\
          &       &       & What technology is Ethereum 2.0 planning to incorporate? &  \\
\cmidrule{2-5}          & \multirow{10}[10]{*}{\textbf{Bitcoin}} & \multicolumn{1}{l}{\multirow{2}[2]{*}{\textbf{(Bitcoin, CreatedBy, SatoshiNakamoto)}}} & Who created Bitcoin? & \multirow{2}[2]{*}{satoshi nakamoto} \\
          &       &       & By whom was Bitcoin developed? &  \\
\cmidrule{3-5}          &       & \multirow{2}[2]{*}{\textbf{(Bitcoin, Uses, BlockchainTechnology)}} & What technology does Bitcoin utilize? & \multirow{2}[2]{*}{blockchain technology} \\
          &       &       & What is the basis of Bitcoin's operation? &  \\
\cmidrule{3-5}          &       & \multirow{2}[2]{*}{\textbf{(Bitcoin, CapableOf, PeerToPeerTransactions)}} & What type of transactions is Bitcoin capable of? & \multirow{2}[2]{*}{peer-to-peer transactions} \\
          &       &       & What kind of financial transactions can Bitcoin perform? &  \\
\cmidrule{3-5}          &       & \multirow{2}[2]{*}{\textbf{(Bitcoin, HasProperty, Decentralized)}} & What property does Bitcoin have? & \multirow{2}[2]{*}{decentralized} \\
          &       &       & How is Bitcoin characterized in terms of its structure? &  \\
\cmidrule{3-5}          &       & \multirow{2}[2]{*}{\textbf{(Bitcoin, Causes, Speculation)}} & What does Bitcoin cause in the market? & \multirow{2}[2]{*}{speculation} \\
          &       &       & What outcome is often associated with Bitcoin's presence in the market? &  \\
    \midrule
    \multirow{20}[20]{*}{\textbf{Culture}} & \multirow{10}[10]{*}{\textbf{Vlogging}} & \multicolumn{1}{l}{\multirow{2}[2]{*}{\textbf{(Vlogging, HasPrerequisite, ContentPlanning)}}} & What is a prerequisite for Vlogging? & \multirow{2}[2]{*}{content planning} \\
          &       &       & Before starting to Vlog, what must one prepare? &  \\
\cmidrule{3-5}          &       & \multirow{2}[2]{*}{\textbf{(Vlogging, CausesDesire, ToBeFamous)}} & What desire does Vlogging cause? & \multirow{2}[2]{*}{to be famous} \\
          &       &       & Why do many people start Vlogging? &  \\
\cmidrule{3-5}          &       & \multirow{2}[2]{*}{\textbf{(Vlogging, Entails, VideoEditing)}} & What does Vlogging entail? & \multirow{2}[2]{*}{video editing} \\
          &       &       & Besides shooting content, what is an essential skill for Vlogging? &  \\
\cmidrule{3-5}          &       & \multirow{2}[2]{*}{\textbf{(Vlogging, Causes, ViewerEngagement)}} & What does vlogging increase? & \multirow{2}[2]{*}{viewer engagement} \\
          &       &       & What is a result of vlogging? &  \\
\cmidrule{3-5}          &       & \multirow{2}[2]{*}{\textbf{(Vlogging, HasLastSubevent, Uploading)}} & What is the final stage in the process of Vlogging? & \multirow{2}[2]{*}{uploading} \\
          &       &       & What activity typically concludes the vlogging process? &  \\
\cmidrule{2-5}          & \multirow{10}[10]{*}{\textbf{Fan Fiction}} & \multicolumn{1}{l}{\multirow{2}[2]{*}{\textbf{(Fan Fiction, IsA, LiteraryGenre)}}} & What is Fan Fiction classified as? & \multirow{2}[2]{*}{literary genre} \\
          &       &       & Under which category does Fan Fiction fall? &  \\
\cmidrule{3-5}          &       & \multirow{2}[2]{*}{\textbf{(Fan Fiction, InspiredBy, ExistingWorks)}} & What inspires Fan Fiction? & \multirow{2}[2]{*}{existing works} \\
          &       &       & What is the source material for Fan Fiction? &  \\
\cmidrule{3-5}          &       & \multirow{2}[2]{*}{\textbf{(Fan Fiction, IsFor, Entertainment)}} & What is the purpose of Fan Fiction? & \multirow{2}[2]{*}{entertainment} \\
          &       &       & Why do people engage with Fan Fiction? &  \\
\cmidrule{3-5}          &       & \multirow{2}[2]{*}{\textbf{(Fan Fiction, MayViolate, CopyrightLaw)}} & What law may fan fiction violate? & \multirow{2}[2]{*}{copyright law} \\
          &       &       & What legal issue does fan fiction potentially encounter? &  \\
\cmidrule{3-5}          &       & \multirow{2}[2]{*}{\textbf{(Fan Fiction, Addresses, UnexploredPlotlines)}} & What does Fan Fiction specifically address? & \multirow{2}[2]{*}{unexplored plotlines} \\
          &       &       & What kind of plotlines does Fan Fiction focus on? &  \\
    \midrule
    \multirow{20}[20]{*}{\textbf{Health and Medical}} & \multirow{10}[10]{*}{\textbf{Nanomedicine}} & \multicolumn{1}{l}{\multirow{2}[2]{*}{\textbf{(Nanomedicine, UsedFor, Targeted Drug Delivery)}}} & What is Nanomedicine used for? & \multirow{2}[2]{*}{targeted drug delivery} \\
          &       &       & What application of Nanomedicine allows for precise medication delivery? &  \\
\cmidrule{3-5}          &       & \multirow{2}[2]{*}{\textbf{(Nanomedicine, MotivatedByGoal, Improve Treatment Efficacy)}} & What goal motivates the field of nanomedicine? & \multirow{2}[2]{*}{improve treatment efficacy} \\
          &       &       & Why is nanomedicine being developed and researched? &  \\
\cmidrule{3-5}          &       & \multirow{2}[2]{*}{\textbf{(Nanomedicine, CapableOf, Crossing Blood-Brain Barrier)}} & What is nanomedicine capable of in terms of biological barriers? & \multirow{2}[2]{*}{crossing blood-brain barrier} \\
          &       &       & How does nanomedicine interact with the blood-brain barrier? &  \\
\cmidrule{3-5}          &       & \multirow{2}[2]{*}{\textbf{(Nanomedicine, CausesDesire, Minimally Invasive Treatment)}} & What does nanomedicine cause a desire for? & \multirow{2}[2]{*}{minimally invasive treatment} \\
          &       &       & What kind of treatment does nanomedicine make more desirable? &  \\
\cmidrule{3-5}          &       & \multirow{2}[2]{*}{\textbf{(Nanomedicine, Requires, Nanoparticles)}} & What does Nanomedicine require to function? & \multirow{2}[2]{*}{nanoparticles} \\
          &       &       & What are the key components essential for Nanomedicine? &  \\
\cmidrule{2-5}          & \multirow{10}[10]{*}{\textbf{Nutritional Genomics}} & \multicolumn{1}{l}{\multirow{2}[2]{*}{\textbf{(Nutritional Genomics, Studies, Gene-Nutrient Interaction)}}} & What does Nutritional Genomics primarily study? & \multirow{2}[2]{*}{gene-nutrient interaction} \\
          &       &       & What is the focus of research in Nutritional Genomics? &  \\
\cmidrule{3-5}          &       & \multirow{2}[2]{*}{\textbf{(Nutritional Genomics, AimsTo, Personalize Nutrition)}} & What is the goal of Nutritional Genomics? & \multirow{2}[2]{*}{personalize nutrition} \\
          &       &       & What does Nutritional Genomics strive to achieve in its application? &  \\
\cmidrule{3-5}          &       & \multirow{2}[2]{*}{\textbf{(Nutritional Genomics, HelpsIn, Disease Prevention)}} & How does Nutritional Genomics contribute to health? & \multirow{2}[2]{*}{disease prevention} \\
          &       &       & In what way does Nutritional Genomics play a role in healthcare? &  \\
\cmidrule{3-5}          &       & \multirow{2}[2]{*}{\textbf{(Nutritional Genomics, ContributesTo, Healthier Diet Planning)}} & What does Nutritional Genomics contribute to? & \multirow{2}[2]{*}{healthier diet planning} \\
          &       &       & What aspect of health is Nutritional Genomics associated with improving? &  \\
\cmidrule{3-5}          &       & \multirow{2}[2]{*}{\textbf{(Nutritional Genomics, Affects, Nutrient Metabolism)}} & What does Nutritional Genomics affect? & \multirow{2}[2]{*}{nutrient metabolism} \\
          &       &       & What aspect of health does Nutritional Genomics have an impact on? &  \\
    \bottomrule
    \end{tabular}%
    }
  \label{tab:examples_concept_1k_appendix}%
\end{table*}%

\begin{figure*}[htbp]
    \centering
    \includegraphics[width=0.99\linewidth]{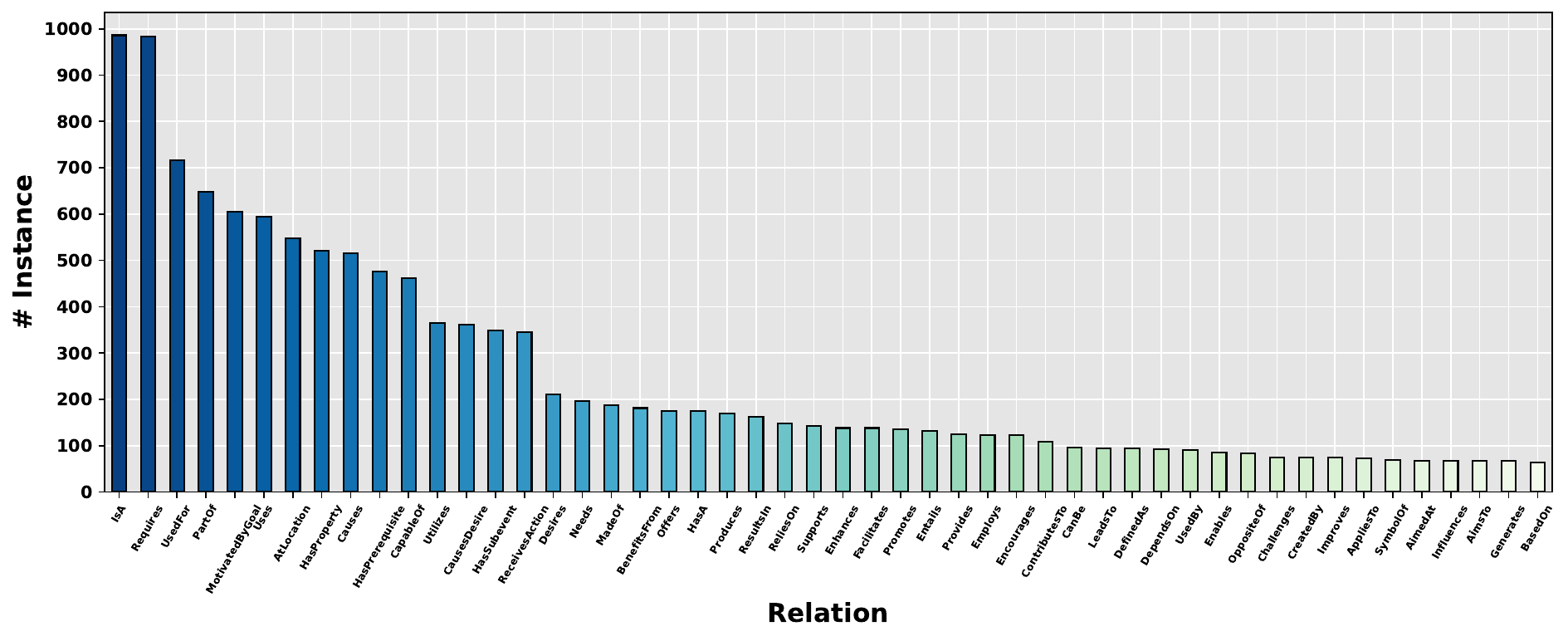}
    \caption{The histogram of the relations in Concept-1K. The top-50 relations are shown.}
    \label{fig:histogram_relation}
\end{figure*}

\begin{figure*}[htbp]
    \centering
    \includegraphics[width=0.99\linewidth]{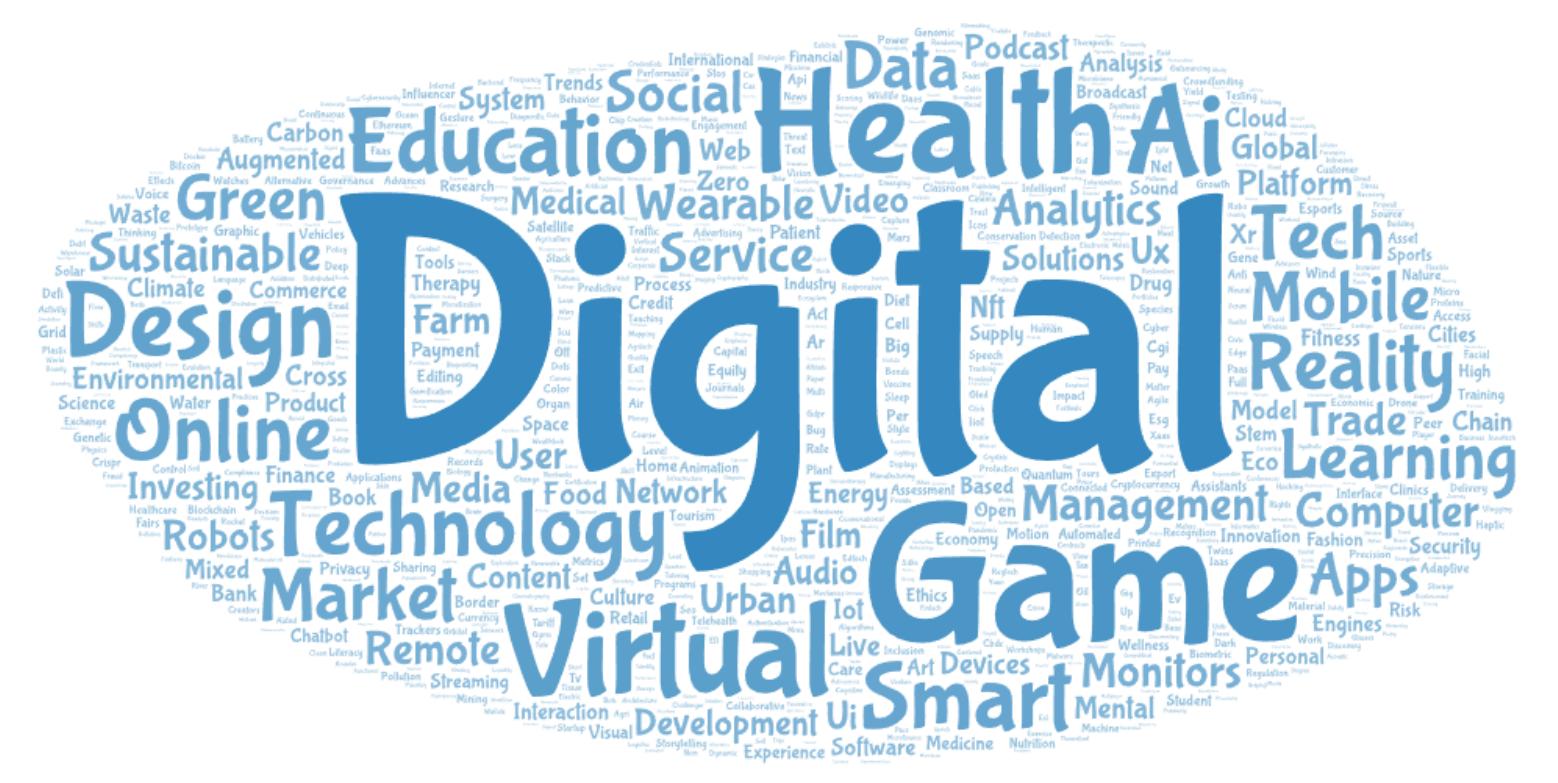}
    \caption{The word cloud diagram of Concept-1K.}
    \label{fig:word_cloud_diagram}
\end{figure*}

% Table generated by Excel2LaTeX from sheet 'Concepts List'
\begin{table*}[htbp]
  \tiny
  \centering
  \caption{The concept list of Concept-1K. The concepts are sorted in alphabetically order.}
  % \resizebox{0.75\linewidth}{!}{
    \begin{tabular}{p{3.5cm}p{3.5cm}p{3.5cm}p{3.5cm}}
    \toprule
    360-Degree Videos & Bioprinting Organs & Critical Thinking & Documentary Collections \\
    3D Audio & Biotechnology & Cross-border e-Commerce & Documentary Films \\
    3D Bioprinting & Bitcoin & Cross-border E-commerce & Drone Technology \\
    3D Modeling & Bitcoin Halving & Cross-Border Healthcare & Drought Resilience \\
    3D Printed Drugs & Blended Learning & Cross-Border Payments & Drug Delivery Systems \\
    3D Printing & Blockchain Health Records & Cross-Cultural Exchange & Dynamic Game Content \\
    3D Rendering & Blockchain Supply Chain & Cross-Platform Gaming & E-Book Popularity \\
    3D Rendering Engines & Blockchain Technology & Crowdfunding & E-Books in Education \\
    3D Scanning & Brain-Computer Interface & Crowdfunding Arts & Eco-Cities \\
    5G Networks & Branding & Crowdfunding Innovations & Eco-Friendly \\
    5G Technology & Brexit Impact & Crowdfunding Platforms & Eco-Friendly Medical Products \\
    Academic Journals Online & Broadcast Graphics & Crowdsourced Projects & E-commerce \\
    Accessibility in Design & Broadcast Technology & Cryptocurrency & e-Commerce Innovation \\
    Acoustic Modeling & Bug Bounty Programs & Cryptocurrency Adoption & E-Commerce Technology \\
    Activity Trackers & Business Model Innovation & Cryptocurrency Exchanges & Economic Sanctions \\
    Adaptive Biotechnologies & Business Process Outsourcing & Cryptocurrency Investing & Ecosystem Services \\
    Adaptive Game AI & Cable TV Technology & Cryptocurrency Mining & Ecotourism \\
    Adaptive Learning & California Consumer Privacy Act & Cryptography & Eco-Tourism \\
    Addiction Recovery Apps & Cancer Immunotherapy & Cultural Competence & Edge Computing \\
    Additive Manufacturing & Car Sharing & Cultured Meat & Edtech Evolution \\
    Adult Education & Carbon Capture & Currency Fluctuations & EdTech Innovation \\
    Affiliate Marketing & Carbon Credits & Currency Hedging & Educational Apps \\
    Agile Methodology & Carbon Footprint & Customer Experience & Educational Data Mining \\
    Agile UX & Carbon Neutral & Customer Relationship Management & Educational Games \\
    Agri-Robots & Carbon Tax & Cyber Activism & Educational Podcasts \\
    AgriTech & CBDC  & Cyber Forensics & Educational Vlogging \\
    Agroforestry & Central Bank Policies & Cybernetic Implants & Edutainment \\
    Agtech Innovation & CGI   & Cybersecurity & E-Governance \\
    AI Diagnostics & CGI Animation & Cybersecurity Investments & E-Health \\
    AI Drug Discovery & Challenger Banks & DAOs  & E-Learning \\
    AI Epidemiology & Chatbot Design & Dark Matter Research & Electric Vehicles \\
    AI Ethics & Chatbot Technology & Data Privacy Regulations & Electronic Health Records \\
    AI Governance & Chatbots & Data Science & Electronic Skin Patches \\
    AI in Education & Cinema Podcasts & Data Security & Email Marketing \\
    AI in Radiology & Cinematography & Decentralized Finance & E-Mentoring \\
    AI Pathology & Circular Economy & Deep Learning & Emerging Markets \\
    Air Pollution Sensors & Citizen Journalism & DeFi Platforms & Emerging Markets Growth \\
    Algorithmic Trading & Civic Tech & Deflation Risks & Encryption Algorithms \\
    Alternative Credentialing & Classroom Technologies & Deforestation & Endangered Species \\
    Alternative Proteins & Clean Technology & Dementia Care Technology & Energy Efficiency \\
    Animation Advances & Climate Adaptation & Design Sprints & Environmental Advocacy \\
    Animation in UI & Climate Change & Design Thinking & Environmental Economics \\
    Animation Software & Climate Finance & DevOps Practices & Environmental Education \\
    Animation Technology & Climate Legislation & Diet Tracking Apps & Environmental Health Surveillance \\
    Antibiotic Stewardship & Climate Mitigation & Digital Advertising & Environmental Monitoring \\
    Anti-Money Laundering Solutions & Climate Tech & Digital Art & Environmental Policy \\
    Antitrust Regulations & Clinical Decision Support & Digital Asset Management & Environmental Protection \\
    Antiviral Therapies & Cloud Computing & Digital Badges & Environmental Sustainability \\
    API Economy & Cloud Gaming & Digital Broadcasting & E-Paper Technology \\
    Apprenticeships & Cloud Storage & Digital Cinema & E-Portfolios \\
    Aquaponics & Cloud Video Editing & Digital Cinematography & ESG Investing \\
    AR Contact Lenses & Cloud-Based PACS & Digital Collaboration & ESG Reporting \\
    Artificial Intelligence & Coastal Erosion & Digital Comics & ESL/EFL Teaching \\
    Astrophysics & Coding Education & Digital Content Creation & eSports \\
    Attendance Tracking & Cognitive Behavioral Apps & Digital Creators & eSports Growth \\
    Audience Measurement & Cognitive Walkthrough & Digital Currencies & eSports Platforms \\
    Audio Books & Collaborative Editing & Digital Curriculum & eSports Tournaments \\
    Audio Compression & Collaborative Learning & Digital Education & Ethereum \\
    Audio Editing Software & Collaborative Platforms & Digital Festivals & Ethereum 2.0 \\
    Audio Recognition & Color Grading & Digital Health & Ethical Clothing \\
    Audio Synthesis & Color Theory & Digital Illustration & Ethical Hacking \\
    Augmented Reality & Commodity Markets & Digital Journalism & EV Battery Technology \\
    Augmented Reality Content & Community Learning & Digital Libraries & Exercise Apps \\
    Augmented Reality Displays & Competency-Based Education & Digital Literacy & Exit Strategies \\
    Augmented Reality Experiences & Composting & Digital Marketing & Exoplanet Discovery \\
    Augmented Reality Games & Computational Thinking & Digital Mental Health & Experiential Learning \\
    Augmented Reality Gaming & Computer Hardware & Digital Nomadism & Export Controls \\
    Augmented Reality Learning & Computer Vision & Digital Patient Engagement & Export Credit Insurance \\
    Augmented Reality Retail & Computer-aided Design & Digital Pharmacy & Export Import \\
    Augmented Reality Surgery & Connected Devices & Digital Portfolios & FaaS \\
    Automated Dispensing & Conservation Efforts & Digital Psychiatry & Facial Recognition \\
    Autonomous Vehicles & Contactless Payments & Digital Rights & Facial Recognition Tech \\
    BaaS  & Containerization & Digital Rights Management & Fact-Checking \\
    Backend Development & Content Delivery Networks & Digital Runways & Fan Fiction \\
    Battery Technologies & Content Distribution Networks & Digital Sculpture & Farm Management Software \\
    Beekeeping & Content Management Systems & Digital Signal Processing & Fashion Apps \\
    Behavioral Health Integration & Content Marketing & Digital Storytelling & Fashion Blogging \\
    Big Data & Continuous Assessment & Digital Textbooks & Feedback Culture \\
    Big Data Analytics & Continuous Glucose Monitors & Digital Therapeutics & Fermented Foods \\
    Bike Sharing & Continuous Patient Monitoring & Digital Transformation & Field Trips \\
    Bilingual Education & Conversational AI & Digital Twins & Film Festivals \\
    Biodiversity & Conversational Design & Digital Twins Healthcare & Film Production Tech \\
    Bioelectronic Medicine & Corporate Restructuring & Digital Wallets & Film Reviews \\
    Bioenergy & Corporate Sustainability & Digital Yuan & Film Scoring \\
    Biohacking & Counseling Services & Direct Listings & Film Workshops \\
    Biomaterials & Course Management Systems & Disease Prediction AI & Financial Crime Compliance \\
    Biomedical Engineering & COVID-19 Pandemic & Distance Learning & Financial Inclusion \\
    Biometric Authentication & Credit Scoring & Distributed Computing & Financial Stability \\
    Biometric Monitoring & CRISPR Cas-9 & Diversity Training & Financial Technology \\
    Biometric Systems & CRISPR Technology & Docker Technology & Fintech Evolution \\
    \bottomrule
    \end{tabular}%
    % }
  \label{tab:concept_list_1}%
\end{table*}%

% Table generated by Excel2LaTeX from sheet 'Concepts List'
\begin{table*}[htbp]
   \tiny
  \centering
  \caption{(Continual) The concept list of Concept-1K. The concepts are sorted in alphabetically order.}
  % \resizebox{0.75\linewidth}{!}{
    \begin{tabular}{p{3.5cm}p{3.5cm}p{3.5cm}p{3.5cm}}
    \toprule
    Firewall Technologies & Health Analytics & Live Streaming & Network Monitoring \\
    Fiscal Stimulus & Health Coaching Bots & Live Video Technology & Network Security \\
    Fitness Technology & Health Data Analytics & Livestream Shopping & Net-Zero Targets \\
    Fitness Trackers & Health Data Exchange & Localization in Gaming & Neural Networks \\
    Fitness Wearables & Health Data Privacy & Logistics Technology & Neurofeedback \\
    Flexible Displays & Health Equity & Longevity Medicine & Neuromodulation \\
    Flipped Classroom & Health Equity Solutions & Low-Carbon Technology & News Podcasts \\
    Flood Management & Health Gamification & Machine Learning & NFT Art \\
    Food Safety Technologies & Health Informatics & Machine Learning Teaching & NFT Marketplaces \\
    Food Security & Health IoT & Malware Analysis & NFTs \\
    Food Technology & Health Literacy Platforms & Manufacturing Reshoring & Non-Invasive Diagnostics \\
    Food Traceability & Health Social Networks & Market Regulation Changes & Nutraceuticals \\
    Foreign Exchange & Healthtech Advances & Market Volatility Analysis & Nutrigenomics \\
    Forex Market Trends & Herbicides & Marketing Automation & Nutritional Genomics \\
    Fraud Detection & Heuristic Evaluation & Marketing Strategies & Nutritional Tech \\
    Freemium Gaming Models & High-Frequency Trading & Mars Missions & Ocean Acidification \\
    Frontend Development & High-frequency Trading & Mastery Learning & Ocean Economy \\
    Full-stack Development & High-performance Computing & Material Science & Offshoring \\
    Functional Foods & Holistic Health Approaches & Media Analytics & Oil Price Dynamics \\
    Game Accessibility Features & Home Workout Solutions & Media Encoding & OLED Technology \\
    Game AI & Human-Centered Design & Media Literacy & Omni-channel Retailing \\
    Game Development Engines & Human-Computer Interaction & Media Monitoring & Online Assessments \\
    Game Development Tools & Humanoid Robots & Media Storage Solutions & Online Book Clubs \\
    Game Engines & Hybrid Vehicles & Media Transcoding & Online Certifications \\
    Game Level Design & Hydroelectric Power & Medical Chatbots & Online Communities \\
    Game Monetization Models & Hydroponics & Medical Devices & Online Conferences \\
    Game Optimization & IaaS  & Medical Drones & Online Courses \\
    Game Sound Design & ICOs  & Medical Imaging AI & Online Courseware \\
    Game Streaming & Identity Management & Medical Tricorders & Online Education \\
    Game Voice Acting & IIoT  & Medical Wearables & Online Fairs \\
    Game World Building & Immersive Storytelling & Mental Health Apps & Online Gaming Infrastructure \\
    Gamification in Education & Immune System Mapping & Mental Health Awareness & Online Learning Platforms \\
    Gamification Techniques & Immunogenomics & Mental Health Chatbots & Online Museums \\
    Gaming Consoles & Immunotherapy Advances & Mental Health Monitoring & Online Newsrooms \\
    Gaming Conventions & Impact Investing & Mental Health Technologies & Online Petitions \\
    Gaming Headsets & Inclusive Education & Mental Wellness Apps & Online Retail \\
    Gaming Influencers & Inclusive Game Design & Mergers and Acquisitions & Online Styling \\
    Gaming Monetization & Independent Filmmakers & Metamaterials & Online Tutoring \\
    Gaming Peripherals & Industrial Automation & Metaverse & Online Workshops \\
    Gaming Platforms & Industrial Design & Microbiome Analysis & Open Access Publishing \\
    Gaming SDKs & Industrial Robots & Micro-Credentials & Open Banking \\
    GDPR Compliance & Industry 4.0 & Microfinance & Open Educational Resources \\
    Gender Lens Investing & Inflation Trends & Microfinance Growth & Open Source Software \\
    Gene Editing & Influencer Culture & Microgravity Research & Opioid Alternatives \\
    Gene Therapy & Influencer Marketing & Micro-interactions & Orbital Mechanics \\
    Genetic Counseling & Infographics & Microloans & Organic Farming \\
    Genetic Risk Assessment & Information Architecture & Microprocessors & Organ-on-a-Chip \\
    Genetic Testing & In-game Advertising & Microservices & OTT Services \\
    Genomic Sequencing & In-Game Advertising & Mindfulness Apps & Outcome-Based Education \\
    Genomics & In-Game Purchases & Mixed Reality & Outdoor Education \\
    Geopolitical Risks & Insurtech & Mixed Reality Applications & Outsourcing \\
    Geopolitical Tensions & Insurtech Trends & Mixed Reality Classrooms & P2P \\
    Geothermal Energy & Integrated Care Models & Mixed Reality Gaming & PaaS \\
    Gestural Interfaces & Intelligent Traffic Systems & Mobile Applications & Pandemic Preparedness \\
    Gesture Control & Interaction Design & Mobile Banking & Patient Empowerment \\
    Gig Economy & Interactive Exhibits & Mobile Commerce & Payment Gateways \\
    Gigabit Society & Interactive Game Design & Mobile Game Development & Payment Processing \\
    Global Citizenship & Interactive Storytelling & Mobile Gaming & Pay-per-click Advertising \\
    Global Economic Recovery & Interactive Whiteboards & Mobile Gaming Hardware & Pay-Per-View Services \\
    Global Education & Interest Rate Forecasts & Mobile Health & Peer Tutoring \\
    Global Health Security & Interest Rate Hikes & Mobile Health Applications & Peer-to-Peer Lending \\
    Global Sourcing & International Baccalaureate & Mobile Health Clinics & Penetration Testing \\
    Global Supply Chains & International Marketing & Mobile Learning & Performance Analytics \\
    Global Trade Tensions & International Payments & Mobile Medical Units & Permaculture \\
    Global Warming & International Students & Mobile Payments & Persona Development \\
    Government Bonds & International Trade & Mobility as a Service & Personal Assistants \\
    Graphene Applications & Internships & Mockups & Personal Health Analytics \\
    Graphic Design Software & Intrusion Detection & MOOCs & Personal Health Records \\
    GraphQL & Invasive Species & Motion Capture Technology & Personalized Learning \\
    Green Bonds & Inventory Management & Motion Control Gaming & Personalized Medicine \\
    Green Building Materials & IoT   & Motion Graphics & Personalized Nutrition \\
    Green Corridors & IoT Security & Motion Graphics Design & Pesticides \\
    Green Finance & IPOs  & Movie Streaming & Pharmacogenetics \\
    Green Infrastructure & IPTV  & mRNA Vaccines & Pharmacogenomics \\
    Green Living & Know Your Customer & Multi-camera Setup & Photodynamic Therapy \\
    Green Roofs & Knowledge Process Outsourcing & Multi-factor Authentication & Photonic Crystals \\
    Green Schools & Kubernetes & Multilingual Education & Physics Engines \\
    Green Spaces & Landscape Ecology & Multimedia Production & Planetary Science \\
    Green Technology & Language Learning Apps & Multiplayer Game Servers & Plant-Based Diets \\
    Greenhouse Gases & Lean UX & Music Recommendation Systems & Plant-based Proteins \\
    Grid Computing & Learning Analytics & Nanomedicine & Plastic Alternatives \\
    Groundwater Recharge & Learning Management Systems & Nanotech Drug Delivery & Plastic Pollution \\
    Group Projects & Learning Platforms & Native Species & Player Behavior Analysis \\
    Gut Microbiome & LEED Certification & Natural Capital & Player Engagement Metrics \\
    Habitat Destruction & Letters of Credit & Natural Language Processing & Podcast Popularity \\
    Hacking & Lifelong Learning & Nature Connectivity & Podcasting \\
    Handheld Gaming Devices & Liquidity Mining & Nearshoring & Podcasting Tech \\
    Haptic Feedback Devices & Literary Podcasts & Neobanks & Podcasts in Education \\
    Haptic Technology & Live Broadcasting & Neobanks Emergence & Poetry Slams \\
    Hashtag Campaigns & Live Game Streaming & Net Zero & Pollinator Conservation \\
    \bottomrule
    \end{tabular}%
    % }
  \label{tab:concept_list_2}%
\end{table*}%

% Table generated by Excel2LaTeX from sheet 'Concepts List'
\begin{table*}[htbp]
  \tiny
  \centering
  \caption{(Continual) The concept list of Concept-1K. The concepts are sorted in alphabetically order.}
  % \resizebox{0.75\linewidth}{!}{
    \begin{tabular}{p{3.5cm}p{3.5cm}p{3.5cm}p{3.5cm}}
    \toprule
    Pollution Prevention & Smart Manufacturing & Toxic Chemicals & \multicolumn{1}{l}{Wearable Computers} \\
    Population Health Management & Smart Materials & Trade Agreements & \multicolumn{1}{l}{Wearable Fitness Trackers} \\
    Post-production Workflow & Smart Meters & Trade Finance & \multicolumn{1}{l}{Wearable Gaming Devices} \\
    Precious Metals Investments & Smart Prosthetics & Trade Wars & \multicolumn{1}{l}{Wearable Health} \\
    Precision Agriculture & Smart Speakers & Traffic Management & \multicolumn{1}{l}{Wearable Health Devices} \\
    Precision Farming & Smart Sutures & Typography & \multicolumn{1}{l}{Wearable Medical Technology} \\
    Precision Medicine & Smart Transportation & UI Design Patterns & \multicolumn{1}{l}{Wearable Sports Technology} \\
    Predictive Analytics & Smart Waste Management & Unicorn Startups & \multicolumn{1}{l}{Wearable Tech} \\
    Predictive Healthcare Analytics & Smart Watches & Upcycling & \multicolumn{1}{l}{Web Analytics} \\
    Printmaking & Social Commerce & Urban Analytics & \multicolumn{1}{l}{Web Development} \\
    Privacy Protection & Social Gaming & Urban Farming & \multicolumn{1}{l}{Web Novels} \\
    Private Equity Trends & Social Impact Bonds & Urban Farming Solutions & \multicolumn{1}{l}{Web3} \\
    Probiotics & Social Learning & Urban Forestry & \multicolumn{1}{l}{Webcasting} \\
    Procedural Generation & Social Media Marketing & Urban Greening & \multicolumn{1}{l}{Webinars} \\
    Product Design & Social Media Trends & Urban Mobility & \multicolumn{1}{l}{Wellness Apps} \\
    Professional Development & Social Movements & Urban Sustainability & \multicolumn{1}{l}{Wellness Economy} \\
    Protectionism & Social News & Urban Wildlife & \multicolumn{1}{l}{Wellness Programs} \\
    Prototype Design & Socially Responsible Investing & Usability Testing & \multicolumn{1}{l}{Wetland Restoration} \\
    Public Transportation & Software Development & User Experience Design & \multicolumn{1}{l}{Wildlife Conservation} \\
    Quantitative Trading & Soil Contamination & User Experience in Gaming & \multicolumn{1}{l}{Wildlife Tourism} \\
    Quantum Computing & Solar Energy Finance & User Flow Diagrams & \multicolumn{1}{l}{Wind Energy} \\
    Quantum Cryptography & Solar Power & User Interface Design & \multicolumn{1}{l}{Wind Energy Investments} \\
    Quantum Dots & Sound Analysis & User Journey Mapping & \multicolumn{1}{l}{Wireframing} \\
    Quantum Mechanics & Sound Engineering & User Research & \multicolumn{1}{l}{Wireless Networking} \\
    Racial Equity Investing & Sound Mixing & UX Metrics & \multicolumn{1}{l}{Work from Home} \\
    Radiomics & Sovereign Debt Issues & UX Writing & \multicolumn{1}{l}{XaaS} \\
    Rainwater Harvesting & Space Exploration & UX/UI Design & \multicolumn{1}{l}{XR} \\
    Ransomware & Space Telescopes & Vaccine Technology & \multicolumn{1}{l}{XR Gaming Experiences} \\
    Recycling & Space Tourism & Venture Capital Shifts & \multicolumn{1}{l}{Yield Farming} \\
    Regenerative Medicine & Spatial Audio & Vertical Farming & \multicolumn{1}{l}{YouTube Creators} \\
    Regtech & Speech Synthesis & Video Content Creation & \multicolumn{1}{l}{Zero Trust Security} \\
    Regtech Solutions & Speech-to-Text & Video Marketing & \multicolumn{1}{l}{Zero Waste} \\
    Reinforcement Learning & Sports Data Analysis & Video on Demand & \multicolumn{1}{l}{Zero Waste Lifestyle} \\
    Remote Health Consultations & Sports Tech & Video Streaming &  \\
    Remote ICU Monitoring & Stagflation Concerns & Viral Challenges &  \\
    Remote Learning & Stakeholder Capitalism & Virtual Assets &  \\
    Remote Monitoring Tools & Startup Ecosystems & Virtual Assistants &  \\
    Remote Patient Monitoring & STEAM Education & Virtual Book Tours &  \\
    Remote Performances & Stem Cell Research & Virtual Care &  \\
    Remote Therapy Sessions & Stem Cell Therapy & Virtual Classrooms &  \\
    Remote Video Production & STEM Education & Virtual Clinical Trials &  \\
    Remote Work Technologies & Stormwater Management & Virtual Concerts &  \\
    Remote Workshops & Storyboarding & Virtual Conferences &  \\
    Renewable Energy & STOs  & Virtual Desktops &  \\
    Responsible Travel & Streaming Audio & Virtual Events &  \\
    Responsive Web Design & Streaming Services & Virtual Exhibitions &  \\
    RESTful Services & Stress Management Tools & Virtual Fashion &  \\
    River Restoration & Student Engagement & Virtual Fitting Rooms &  \\
    Robo-advisors & Student Engagement Tools & Virtual Galleries &  \\
    Robotic Surgery & Student Wellbeing & Virtual Goods &  \\
    Robotics & Study Abroad Programs & Virtual Gyms &  \\
    Robotics in Education & Style Influencers & Virtual Health Assistants &  \\
    Rocket Science & Subscription Gaming Services & Virtual Health Fairs &  \\
    SaaS  & Supercomputing & Virtual Keyboards &  \\
    Sales Automation & Supply Chain Finance & Virtual Protests &  \\
    Satellite Broadcasting & Supply Chain Innovation & Virtual Prototyping &  \\
    Satellite Internet & Supply Chain Optimization & Virtual Reality &  \\
    Satellite Technology & Sustainable Agriculture & Virtual Reality Arcades &  \\
    Scale-up Companies & Sustainable Cities & Virtual Reality Education &  \\
    Screencasting & Sustainable Development & Virtual Reality Experiences &  \\
    Screenwriting & Sustainable Development Goals & Virtual Reality Filmmaking &  \\
    Scrum Framework & Sustainable Energy & Virtual Reality Gaming &  \\
    Sea Level Rise & Sustainable Fashion & Virtual Reality Sports &  \\
    Self-Publishing & Sustainable Fishing & Virtual Reality Theater &  \\
    Semantic Analysis & Sustainable Healthcare & Virtual Reality Therapy &  \\
    Semiconductor Tech & Sustainable Investing & Virtual Reality Training &  \\
    SEO   & Sustainable Tech & Virtual Set Design &  \\
    Serious Games & Sustainable Transport & Virtual Shopping &  \\
    Serverless Architecture & Synthetic Biology & Virtual Teaching Assistants &  \\
    Service Design & Tariff Implications & Virtual Tours &  \\
    Service Robots & Tariff Negotiations & Visual Effects &  \\
    Sharing Economy & Teacher Training & Visual Effects Software &  \\
    Shipping Solutions & Telehealth Expansion & Visual Hierarchy &  \\
    Short Films & Telehealth Licensing & Visual Storytelling &  \\
    Simulation Games & Telehealth Solutions & Vlogging &  \\
    Skill Sharing & Telemedicine & Vocational Training &  \\
    Sleep Tech Devices & Teleophthalmology & Voice Recognition &  \\
    Sleep Technology & Telepsychiatry & Voice User Interface &  \\
    Slow Fashion & Tele-rehabilitation & Vulnerability Assessment &  \\
    Smart Beds & Telestroke Services & Walkability &  \\
    Smart Cities & Television Production & Warehouse Automation &  \\
    Smart Cities Technologies & Text-to-Speech & Waste Management &  \\
    Smart Contracts & Theoretical Physics & Waste Reduction &  \\
    Smart Diagnostic Wearables & Therapeutic Games & Water Conservation &  \\
    Smart Glasses & Therapeutic Robots & Water Quality Monitoring &  \\
    Smart Grids & Threat Intelligence & Water Scarcity Solutions &  \\
    Smart Health Watches & TikTok Dance & Watershed Management &  \\
    Smart Home & Tissue Engineering & Wealth Gap &  \\
    Smart Inhalers & Tokenization & Wealthtech &  \\
    Smart Lighting & Tokenization of Assets & Wealthtech Advancements &  \\
    \bottomrule
    \end{tabular}%
    % }
  \label{tab:concept_list_3}%
\end{table*}%

\end{document}